\pdfoutput=1
\documentclass[10pt,twocolumn,letterpaper]{article}

\usepackage[pagenumbers]{cvpr}
\usepackage{amsmath,amssymb,booktabs,graphicx,multirow}
\usepackage{float}
\usepackage{wrapfig}%
\makeatletter
\newcommand{\needspace}[1]{\par\penalty-100\begingroup\setlength{\dimen@}{#1}%
  \dimen@ii\pagegoal\advance\dimen@ii-\pagetotal
  \ifdim\dimen@>\dimen@ii\ifdim\dimen@ii>\z@\vfil\fi\break\fi\endgroup}
\makeatother
\usepackage{xcolor}
\usepackage{xspace}
\usepackage{tikz}
\usetikzlibrary{arrows.meta,positioning,calc,shapes.geometric}

\newcommand{\paper}{Blur2Dynamics\xspace}

\newcommand{\ulabel}[1]{\hbox to 0pt{\hss\scriptsize #1\hss}}

\newcommand{\best}[1]{\textbf{#1}}
\newcommand{\second}[1]{\underline{#1}}
\IfFileExists{colortbl.sty}{\usepackage{colortbl}\newcommand{\oursrow}{\rowcolor[gray]{0.92}}}{\newcommand{\oursrow}{}}
\newcommand{\ours}[1]{\textcolor[rgb]{0.165,0.471,0.839}{\textbf{#1}}}%

\usepackage[T1,OT1]{fontenc}
\usepackage{array}
\newcolumntype{L}[1]{>{\raggedright\arraybackslash}p{#1}}
\usepackage{adjustbox}
\definecolor{oursbg}{RGB}{229,239,250}
\IfFileExists{colortbl.sty}{\newcolumntype{O}{>{\columncolor{oursbg}}c}}{\newcolumntype{O}{c}}

\definecolor{cvprblue}{rgb}{0.21,0.49,0.74}
\usepackage[pagebackref,breaklinks,colorlinks,allcolors=cvprblue]{hyperref}

\def\confName{CVPR}
\def\confYear{2027}

\title{Blur2Dynamics: Adapting Video Encoders to Infer Latent-Video Features from a Single Blurred Image}
\author{Harshana Weligampola \qquad Tharindu Wickramasinghe \qquad Yu Yuan \qquad Stanley H. Chan\\
Purdue University}

\newcommand{\teaserfigure}{%
  \vspace{0.1em}
  \begin{center}
    \includegraphics[width=0.88\textwidth]{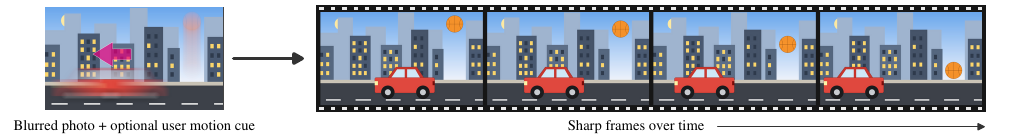}\\[0.15em]
    \includegraphics[width=0.88\textwidth]{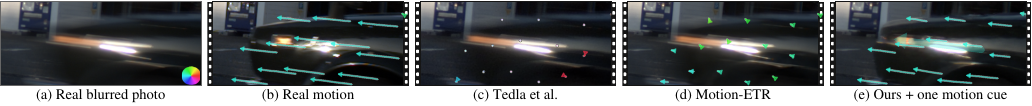}
    \captionsetup{hypcap=false}
    \captionof{figure}{Top, an illustration of the task: from one blurred photo, optionally with an arrow cue (magenta) drawn by the user,
      recover the sharp frames of the exposure. Bottom, a real blurred photo (RBI~\cite{zhong2023blur}) (a): prior
      methods~\cite{tedla2025generating,zhang2022exposure} stay blurred and barely move the car (c, d); ours with one cue is sharp
      and follows the real motion (b, e). Paths: CoTracker3~\cite{karaev2025cotracker3}, colored by direction.}
    \label{fig:teaser}
  \end{center}
  \vspace{0.6em}}
\makeatletter
\apptocmd{\@maketitle}{\teaserfigure}{}{%
  \PackageError{Blur2Dynamics}{Could not append teaser to title block}{}}
\makeatother

\begin{document}
\addtocontents{toc}{\protect\setcounter{tocdepth}{-1}}
\maketitle

\begin{abstract}\setlength{\baselineskip}{12.8pt}%
Blur-to-video generation is a challenging inverse problem where the goal is to
take a single blurred image and generate the underlying clean video. However,
since motion blur is the result of temporal integration and there are
infinitely many pathways that can lead to the same blurred image, the problem
is intrinsically ill-posed, and many existing methods either produce incorrect
dynamics or require dense motion trajectories as side information. In this
paper, we address the problem by deriving the underlying probabilistic
graphical model and a set of approximations that lead to a practical
implementation. The proposed method, \paper, uses a pretrained V-JEPA encoder
to predict the features that a frozen V-JEPA teacher extracts from the sharp
video that produced the blur, and a conditional video generator to reconstruct
the sharp video from the blur and these features. Because these features
carry motion information the renderer follows, user-drawn arrows (arrow cues)
resolve the ambiguous temporal ordering. Without cues, \paper has the
lowest optical-flow error, scored up to time reversal, of tested methods on most
held-out benchmarks, with competitive fidelity. With simulated
cues, cued objects follow in 95.8\% of synthetic multi-object commands and 58 of 60 commands on real clips.
\par\vspace*{-12pt}\end{abstract}

\section{Introduction}
\label{sec:intro}

A motion-blurred photograph is a temporal measurement formed by integrating an
evolving scene during exposure. Recovering the underlying video rather than a
single sharp frame enables temporal inspection, high-frame-rate reconstruction,
and motion understanding. However, this inverse problem is inherently
ambiguous: different motion sequences can produce the same blur
(\cref{fig:teaser}). A car moving left and a falling ball leave two streaks, and under a time-symmetric
exposure each could equally have moved the other way: $n$ movers admit up to $2^n$ videos, since the blur fixes each
motion's axis and extent but not its sign~\cite{jin2018learning}. The challenge is therefore to reconstruct plausible dynamics
alongside visually convincing frames.

Existing blur-to-video methods reconstruct frames
directly~\cite{jin2018learning,tedla2025generating} or estimate motion through
task-specific
representations~\cite{purohit2019bringing,zhang2022exposure,zhong2022animation},
such as exposure trajectories. These approaches can produce plausible frames
whose motion the blur does not support under either temporal order and, beyond
coarse regional direction guidance~\cite{zhong2022animation}, offer little means to
choose among the motions the blur admits. Motion-controlled video
generators~\cite{wang2024motionctrl,yin2023dragnuwa,wu2024draganything,tora2024,wanmove2025}
offer additional control but typically require prescribed trajectories or dense
motion fields.
\textbf{The missing capability is to infer an editable, motion-bearing
representation of the latent video directly from the blur.}

To address this challenge, we study the probabilistic relationship between an
evolving sharp video and its blurred observation (\cref{sec:model}).
Introducing learned, motion-bearing video features, the \emph{dynamics
features}, as an intermediate variable decomposes the inverse problem into two
connected tasks: inferring these features from the blur and reconstructing the
sharp sequence conditioned on them. A pretrained video encoder yields them from a
video but not from a blur, the only test-time input. This
motivates \paper, which adapts a pretrained self-supervised video encoder to
infer features consistent with the blur. A frozen
V-JEPA~\cite{bardes2024revisiting,murlabadia2026vjepa21} teacher
extracts features from sharp training sequences, while a LoRA-adapted encoder
learns to predict them from temporally repeated copies of the blurred image; we call this cross-observation adaptation.

The inferred features also support motion control: optional \emph{arrow cues} (cues for short), drawn
by users on moving objects, set which way each moved.
Without cues, a learned default favors the valid order on one synthetic
settling-versus-reversal ambiguity. A conditional generator
then renders the sharp video from the blur and the features.

Without cues, \paper has the lowest optical-flow error, scored up to time
reversal, among the tested published methods on most held-out benchmarks, with competitive picture fidelity. With simulated
cues, both cued motion groups follow
their arrow cues in 96.7\% of commands on held-out real clips, against 93.3\%, 58.3\% and 35.0\% for
trajectory-controlled generators given each arrow cue as a point track; reversing one region against the rest of a real
clip succeeds in 93.9\% of trials. Linear probes show that the inferred features encode the blur's unsigned motion axis, and
interventions on sharp-video features show that the renderer follows the motion they encode.

Our contributions are:
\begin{enumerate}
    \item \textbf{Dynamics from a single blur.} Motivated by a probabilistic
    model of blur formation with dynamics features as the intermediate variable,
    cross-observation adaptation of a pretrained video encoder (V-JEPA) predicts,
    from repeated copies of one blurred image, the dense features its frozen copy
    extracts from the sharp video.

    \item \textbf{Editable dynamics.} Arrow cues choose each object's
    direction inside feature inference: an arrow cue sets only the sign of
    an object's motion, while the blur supplies its path and extent, so a cue
    cannot contradict the observation. One encoder takes zero, one or several arrow cues,
    trained on arrow cues simulated from tracked motion, with a learned default without them.

    \item \textbf{The renderer follows the features.} Swapping, reversing, or making
    the sharp-video features physically impossible changes the generated motion accordingly in
    an adapted Wan~2.2~\cite{wan2025wan} renderer, and an oracle-feature sweep shows
    that the remaining error at large blur comes from inferring the features, not
    from rendering them.
\end{enumerate}

\newpage

\section{Related Work}
\label{sec:related}

\noindent\textbf{Blur-to-video reconstruction.}
Several methods predict the sharp sequence directly from the
blur~\cite{zhang2020every,pham2023hypercut,pang2025image,shin2026pickmoment};
Tedla et al.~\cite{tedla2025generating} adapt video diffusion for event
reconstruction and continuation. BiT++, the larger variant of BiT~\cite{zhong2023blur}, whose RBI data we test on,
reads three blurred frames and is trained on that camera, as
blurry-video interpolation~\cite{shen2020blurry,jin2019learning,zhang2020video,oh2022demfi,shang2023joint}
reads several; event cameras~\cite{pan2019bringing,lin2020learning,zhang2022unifying} add a
sensor that records the motion. We start from one blur, and unlike diffusion
deblurring~\cite{whang2022deblurring}, sample the whole exposure rather than one frame. Others pass through an explicit motion or
appearance variable: Motion-ETR~\cite{zhang2022exposure} recovers geometric
exposure trajectories, others estimate motion
fields~\cite{argaw2021optical,argaw2021restoration} or a fast-moving object's
appearance and trajectory~\cite{kotera2019intra,rozumnyi2021shape}, and Jin et
al.~\cite{jin2018learning} first recover the sharp middle frame, then the other
frames outward, with an order-invariant loss for the temporal ambiguity.
The closest precedent is Purohit et al.~\cite{purohit2019bringing}, who train a
blur encoder to regress the motion embedding of a recurrent video autoencoder,
which then decodes from a deblurred middle frame.
DeFMO~\cite{rozumnyi2021defmo} renders sub-frames from a latent embedding of a
blurred object. Unlike Purohit et al., we adapt a pretrained video encoder with
LoRA, regress the dense tokens its frozen copy extracts from the sharp video,
render with a diffusion model, and let per-object cues act inside inference.

\noindent\textbf{Video representations and prediction from one image.}
VideoMAE~\cite{tong2022videomae,wang2023videomaev2} and the JEPA
family~\cite{assran2023ijepa,bardes2024revisiting,murlabadia2026vjepa21}
learn dense spatiotemporal features of an evolving video, which can encode
the arrow of time~\cite{pickup2014seeing,misra2016shuffle,wei2018learning,ghodrati2018video,alrasheed2026latent}, and still-image models
predict future representations, flow, or 3D-CNN
features~\cite{vondrick2016anticipating,gao2018im2flow,parmar2019hallucinet}.
None of these infers a sharp video's features from its blur, which is what our
adaptation learns.

\noindent\textbf{Motion-conditioned generation.}
Zhong et al.~\cite{zhong2022animation} (\emph{Animation from Blur}) condition on
the blurred input itself and resolve blur ambiguity with a dense per-pixel map of
quantized directions, the closest precedent for our optional arrow cues. General
motion-controlled
generators~\cite{wang2024motionctrl,yin2023dragnuwa,wu2024draganything,shi2024motioni2v,niu2024mofa,tora2024,geng2025motion,burgert2025go,wang2025ati,wanmove2025},
built on video diffusion models~\cite{ho2022video,blattmann2023align,yang2025cogvideox,wan2025wan},
instead receive externally specified motion (camera poses, object trajectories,
point tracks or optical flow), sometimes as a few drawn
drags~\cite{yin2023dragnuwa,wu2024draganything}, with an image or text prompt rather
than a blurred observation; keyframe interpolation even runs an image-to-video model
backward in time~\cite{wang2025inbetweening}, but between sharp frames. What sets our arrow cues apart is thus not sparsity: they
carry only a sign and act inside the inference of motion that must stay consistent
with the blur, whereas Zhong et al.'s guide and these generators act at rendering
time.

\noindent\textbf{Motion--content decomposition and feature-aligned generation.}
Video generators have long separated motion from content, as latent
codes~\cite{tulyakov2018mocogan,denton2017unsupervised} or as motion paths and
control signals of diffusion models~\cite{zhao2024motiondirector,wang2023videocomposer},
and recent work aligns video diffusion with pretrained video encoders' features, in
training~\cite{zhang2025videorepa,xiong2026physalign} or at inference with V-JEPA~2
as a reward that steers sampling~\cite{yuan2026inference}. We claim neither the decomposition nor a
feature-conditioned renderer; what is new is inferring the encoder's dense feature
field from a blurred observation, with sparse cues acting inside that inference.
Closest in ingredients, RVD~\cite{yuan2026rvd} reads a compact dynamics token from
a frozen V-JEPA~2's features of a sharp video and renders it with Wan. A blur shows
motion only as its time integral, so the frozen encoder no longer provides these
features and we adapt it.

\section{Blur-to-Dynamics Adaptation}
\label{sec:method}

We first model how a blurred photograph forms (\cref{sec:model}), then learn the
factor this model leaves missing
(\cref{sec:cross_observation_adaptation,sec:ambiguity_training,sec:sparse_cues,sec:narrow_default})
and render the video from it (\cref{sec:feature_renderer}).

\subsection{From blur formation to a learned inverse}
\label{sec:model}

\begin{figure}[t]
  \centering
  \includegraphics[width=\columnwidth]{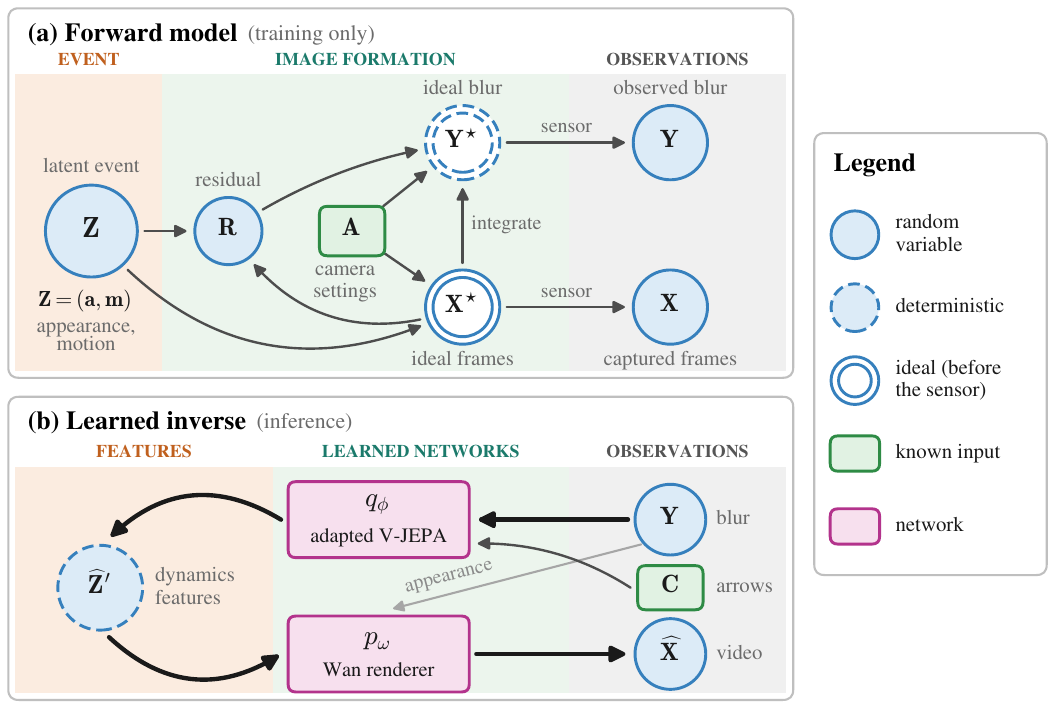}
  \caption{\textbf{Forward model and learned inverse.}
  (a) Forward model, used only in training: the latent event $\mathbf{Z}=(\mathbf{a},\mathbf{m})$
  and camera settings $\mathbf{A}$ give the ideal sharp sequence $\mathbf{X}^\star$, which with $\mathbf{A}$
  and residual $\mathbf{R}$ produces the observed blur $\mathbf{Y}$; a frozen video encoder
  defines $\mathbf{Z}'$ from captured training frames $\mathbf{X}$. (b) Learned inverse (inference): adapted
  V-JEPA, $q_\phi$, maps repeated copies of $\mathbf{Y}$, optionally oriented by arrows
  $\mathbf{C}$, to dynamics features $\widehat{\mathbf{Z}}'$; the Wan renderer, $p_\omega$,
  reconstructs $\widehat{\mathbf{X}}$ from them and the blur. Shaded columns pair the event with the
  features and image formation with the learned networks.}
  \label{fig:background_pgm}
\end{figure}

\noindent\textbf{Blur formation.}
A camera exposure adds up the light that reaches each pixel while the scene
moves, so a ball crossing the frame leaves a streak. Let
$\mathbf{X}^\star=\{\mathbf{x}_t^\star\}_{t=1}^{T}$ be ideal sharp frames of the
event. Since short frames need not cover the entire exposure, the ideal long
exposure is
\begin{equation}
    \underbrace{\mathbf{Y}^\star}_{\text{ideal blur}}=
    \underbrace{\sum_{t=1}^{T}w_t \mathbf{x}_t^\star}_{\text{weighted sharp frames}}
    +\underbrace{\mathbf{R}}_{\text{residual}},
    \label{eq:exposure_gap}
\end{equation}
where $w_t$ are exposure weights and $\mathbf{R}$ is the light accumulated between
the sampled frames. Sensor processing turns these ideal signals into the observed
blur $\mathbf{Y}$ and, during training, into captured sharp frames $\mathbf{X}$.
Real blur--video pairs are hard to collect, so we synthesize $\mathbf{Y}$ from
high-frame-rate clips, as prior work
does~\cite{nah2017deep,jin2018learning,purohit2019bringing}. As in REDS~\cite{nah2019ntire} and Brooks and Barron~\cite{brooks2019learning}, we stand in for
$\mathbf{R}$ with interpolated frames, here from RIFE~\cite{huang2022rife}, and average in
linear light.

\begin{figure*}[t]
  \centering
  \includegraphics[width=\textwidth]{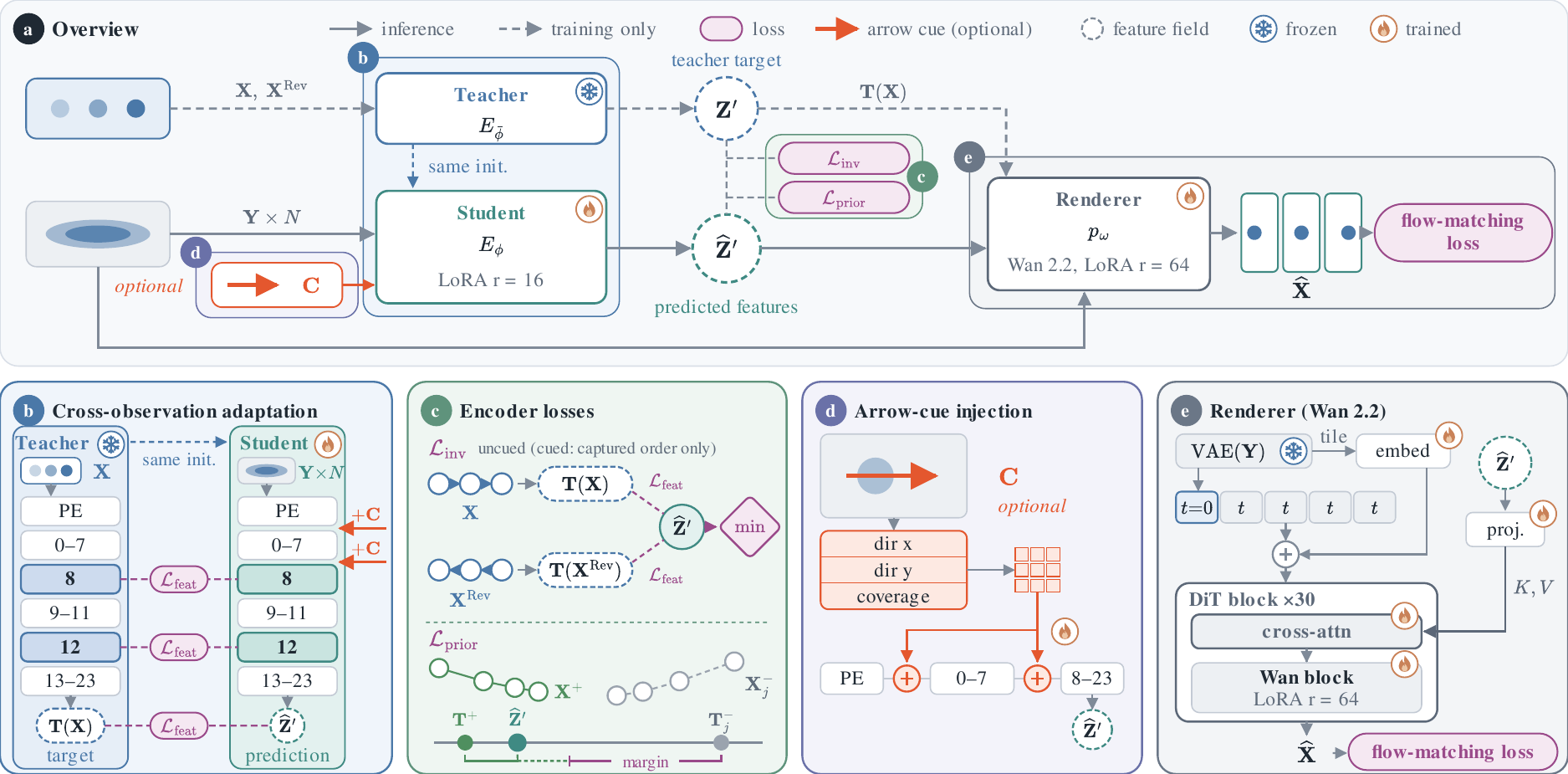}
  \caption{\textbf{Training pipeline.} \textbf{(a)} Overview; tinted regions zoom in below.
  \textbf{(b)} Cross-observation adaptation: teacher and student start from the same V-JEPA
  weights and are matched by $\mathcal L_{\mathrm{feat}}$ at the final tokens and blocks 8/12.
  \textbf{(c)} Encoder losses: uncued, $\mathcal L_{\mathrm{inv}}$ takes the closer of
  $\mathbf{T}(\mathbf{X})$/$\mathbf{T}(\mathbf{X}^{\mathrm{Rev}})$ (cued: $\mathbf{T}(\mathbf{X})$ only); without a cue,
  $\mathcal L_{\mathrm{prior}}$ ranks the valid settling target ahead.
  \textbf{(d)} Arrow cues: a zero-initialized bias on the student only.
  \textbf{(e)} Renderer: the blur and, through cross-attention in every block, the
  features. Dashed: training-only paths.}
  \label{fig:method_training}
\end{figure*}

\noindent\textbf{Why the inverse is ambiguous.}
\Cref{fig:background_pgm}(a) draws this process as a graphical model in which a
latent event $\mathbf{Z}$ produces the sharp frames and the blur. The model
motivates the factorization below and is not fitted to data (derivation in the
supplement). Because integration couples appearance and motion,
different events can explain the same photograph. The simplest case is time
reversal. Under a time-symmetric exposure, the frames $\mathbf{X}^\star$ and their
reversal $\mathbf{X}^{\star,\mathrm{Rev}}{:=}\{\mathbf{x}_{T+1-t}^\star\}_{t=1}^{T}$
give the same blur. Spatially separated motions add local directional
ambiguities~\cite{jin2018learning,zhong2022animation}.

\noindent\textbf{The approximation: dynamics features.}
Rather than estimating $\mathbf{Z}$, we approximate its spatiotemporal
information with $\mathbf{Z}'=E_{\bar\phi}(\mathbf{X})$, the dense spatiotemporal
field that a frozen pretrained video encoder (here, V-JEPA) extracts from
captured sharp training frames. We call $\mathbf{Z}'$ the \emph{dynamics
features}. They may encode appearance and scene structure alongside
motion~\cite{garrido2025intphys}, and we do not assume that they correspond to
$\mathbf{Z}$ or any part of it; their value as motion guidance is tested experimentally. Letting
optional arrow cues $\mathbf{C}$ act through this representation,
\begin{equation}
\begin{split}
    p(\mathbf{X}\mid \mathbf{Y},\mathbf{C})
    &=\int p(\mathbf{X}\mid \mathbf{Y},\mathbf{Z}',\mathbf{C})\,p(\mathbf{Z}'\mid \mathbf{Y},\mathbf{C})\,d\mathbf{Z}' \\
    &\approx\int \underbrace{p_\omega(\mathbf{X}\mid \mathbf{Y},\mathbf{Z}')}_{\text{renderer}}\,
    \underbrace{q_\phi(\mathbf{Z}'\mid \mathbf{Y},\mathbf{C})}_{\text{feature inference}}\,d\mathbf{Z}'.
\end{split}
\label{eq:feature_mediated_model}
\end{equation}
The first line is exact for any well-defined $\mathbf{Z}'$; the second replaces the exact conditionals with learned
models and drops $\mathbf{C}$ from the renderer (\cref{fig:background_pgm}(b)).
The renderer also sees the blur, so $\mathbf{Z}'$ need not determine every video
detail. The cues $\mathbf{C}$, optional arrow cues over objects or their blur traces,
are not part of the forward model but an intervention that selects a temporal
orientation inside $q_\phi$, without prescribing a trajectory or controlling the
renderer.

\noindent\textbf{From the approximation to an implementation.}
At inference only $\mathbf{Y}$ and optional $\mathbf{C}$ are available, so we
replace the posterior by the single estimate of an adapted encoder $E_\phi$
trained mainly with a squared-error loss (\cref{sec:cross_observation_adaptation}),
\begin{equation}
q_\phi(\mathbf{Z}'\mid \mathbf{Y},\mathbf{C})=
\delta\big(\mathbf{Z}'-\underbrace{E_\phi(\operatorname{Repeat}(\mathbf{Y},N),\mathbf{C})}_{\ulabel{adapted encoder on $N$ copies of the blur and the arrow cues}}\big).
\label{eq:deterministic_feature_posterior}
\end{equation}
The posterior can be multi-modal under the ambiguities above; the cues narrow it
toward one temporal interpretation. This point mass collapses
\cref{eq:feature_mediated_model} to one deterministic feature estimate that
conditions the renderer (\cref{sec:feature_renderer}), and keeping the two factors
apart separates feature-inference errors from rendering errors.
\Cref{fig:method_training} shows the training; arrow cues extend, rather than enable,
cue-free inference.

\subsection{Cross-observation adaptation to a static blur}
\label{sec:cross_observation_adaptation}

Let $E_{\bar\phi}$ be a frozen pretrained V-JEPA encoder with stopped-gradient
encoding $\mathbf{T}(\cdot)=\operatorname{stopgrad}(E_{\bar\phi}(\cdot))$.
For a sharp training clip $\mathbf{X}$, the teacher target is
$\mathbf{T}(\mathbf{X})$. We initialize a second copy $E_\phi$ from the same
pretrained weights and adapt its attention and MLP projections with
rank-16 LoRA~\cite{hu2022lora} updates. It receives $N$ identical copies of the
corresponding blurred photograph,
$\widetilde{\mathbf{Y}}=\operatorname{Repeat}(\mathbf{Y},N)$, and predicts
$\widehat{\mathbf{Z}}'=E_\phi(\widetilde{\mathbf{Y}},\mathbf{C})$.

\begin{wrapfigure}{r}{\dimexpr0.44\columnwidth-0.5\columnsep\relax}
\vspace{-12pt}
\hspace*{-0.5\columnsep}\begin{minipage}{\dimexpr\linewidth+0.5\columnsep\relax}
\centering
\includegraphics[width=\linewidth]{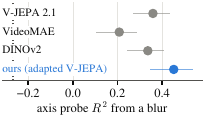}
\vspace{-16pt}
\caption{Motion-axis probes ($R^2$), no renderer, layer chosen in training folds; dotted: no features.}
\label{fig:motiv_probes}
\end{minipage}
\vspace{-12pt}
\end{wrapfigure}

Teacher and student share the same architecture but observe different
measurements of the same event: the teacher sees an evolving sequence and the
student a temporally constant tensor. We call this cross-observation adaptation.
As in cross-modal~\cite{gupta2016cross} or degraded-input~\cite{su2017adapting}
feature distillation~\cite{hinton2015distilling,romero2015fitnets}, learning with
privileged information~\cite{vapnik2009new,lopezpaz2016unifying}, or modality
hallucination~\cite{hoffman2016learning,crasto2019mars}, the student
matches features of another observation. Unlike in that work, our student starts
from the teacher's own weights, and its input has none of the teacher's temporal
content. Only V-JEPA is adapted in this work. Because the $N$ repeated frames
still enter through V-JEPA's tokenization pathway, the student keeps the teacher's
token grid. Matching it therefore turns spatial blur evidence into a temporally
varying field, but it adds no new temporal content. Probes read the motion axis more
easily from the adapted features of a blur than from frozen encoders' features of
the same blur (\cref{fig:motiv_probes}, \cref{sec:res_probes}).

We use the V-JEPA~2.1 ViT-L/384 encoder (1024-dimensional final tokens).
Besides final-token regression, we regularize with layer-normalized
targets from two intermediate blocks:
\begin{equation}
\begin{split}
    \mathcal L_{\mathrm{feat}}(\widehat{\mathbf{Z}}',\mathbf{Z}')
    ={}& \underbrace{\operatorname{MSE}(\widehat{\mathbf{Z}}'_{\mathrm{out}},\mathbf{Z}'_{\mathrm{out}})}_{\text{final tokens}} \\
    &+\underbrace{\frac{\lambda_{\mathrm{mid}}}{|\mathcal B|}
    \sum_{\ell\in\mathcal B}
    \operatorname{MSE}(\operatorname{LN}(\widehat{\mathbf{Z}}'_\ell),
      \operatorname{LN}(\mathbf{Z}'_\ell))}_{\text{intermediate layers}},
\end{split}
\label{eq:feature_distance}
\end{equation}
where $\operatorname{MSE}(\cdot,\cdot)$ is mean squared token distance, $\mathbf{Z}'$ is a
generic target (e.g.\ $\mathbf{T}(\mathbf{X})$), subscripts $\ell$
and $\mathrm{out}$ denote intermediate and final token grids, and $\mathcal B=\{8,12\}$ indexes two
intermediate transformer blocks weighted by $\lambda_{\mathrm{mid}}$,
regularizing the adaptation without assuming any one layer is a
motion-only representation.
We choose these blocks because probing of V-JEPA~2 finds motion direction decodable from about a third of the
encoder's depth, fading toward the output~\cite{joseph2026interpreting}, so final tokens alone under-weight it.

\needspace{8\baselineskip}%
\subsection{Learning under temporal ambiguity}
\label{sec:ambiguity_training}

Under uniform exposure weighting (\cref{sec:model}), a sharp sequence
$\mathbf{X}$ and its temporal reversal $\mathbf{X}^{\mathrm{Rev}}$ give the same
blur, an ambiguity of the observation itself. For an uncued
example we therefore use the lower of the two sequence-level losses against the
teacher's forward and reversed encodings:
\begin{equation}
    \mathcal L_{\mathrm{inv}}
    =\min\{\underbrace{\mathcal L_{\mathrm{feat}}(\widehat{\mathbf{Z}}',\mathbf{T}(\mathbf{X}))}_{\text{captured order}},
    \underbrace{\mathcal L_{\mathrm{feat}}(\widehat{\mathbf{Z}}',\mathbf{T}(\mathbf{X}^{\mathrm{Rev}}))}_{\text{reversed order}}\}.
    \label{eq:inverse-loss}
\end{equation}
Importantly, $\mathbf{T}(\mathbf{X}^{\mathrm{Rev}})$ is obtained by encoding the
reversed video, not by reversing the token array. Unlike Jin et al.'s frame-level
order-invariant loss~\cite{jin2018learning} or HyperCUT's learned
ordering~\cite{pham2023hypercut}, the choice is made between teacher encodings,
once per example rather than per spatial cell. This is the oracle loss of
multiple-choice learning~\cite{guzmanrivera2012multiple,rupprecht2017learning}, with one
prediction and two fixed targets. This handles whole-clip reversal
only. Since $n$ independently moving objects that neither occlude nor interact
admit up to $2^n$ direction combinations, the optional sparse cues
(\cref{sec:sparse_cues}) address the remaining per-object ambiguity. Plain
captured-order matching in place of $\mathcal L_{\mathrm{inv}}$ raises the order-invariant feature error by
only 1.4\%, within seed noise, so $\mathcal L_{\mathrm{inv}}$ is a mild
regularizer.

\subsection{Arrow cues}
\label{sec:sparse_cues}

Optional cues resolve part of the ambiguity in Eq.~\eqref{eq:inverse-loss}.
Each arrow cue, placed over an object or its blur trace, is rasterized as
three channels (horizontal/vertical unit direction weighted by local
coverage, plus coverage itself), resampled to the V-JEPA token grid, and
injected as zero-initialized additive biases after patch embedding and
after block~7 (counting from 0). The cue thus changes
$\widehat{\mathbf{Z}}'$ but has no direct path to the renderer.

Training cues are simulated from the sharp clip's own tracked motion
(supplement), so for a cued example the teacher target is simply
$\mathbf{T}(\mathbf{X})$ (the teacher never receives $\mathbf{C}$), with no
reversal choice. We therefore minimize
$\mathcal L_{\mathrm{feat}}(\widehat{\mathbf{Z}}',\mathbf{T}(\mathbf{X}))$
directly, matching the entire field, including uncued objects, against the
captured order. In effect, an arrow cue sets only the sign of the motion along the blur's
axis in its region, with no speed, path or magnitude, unlike the trajectories of
controllers such as MotionCtrl~\cite{wang2024motionctrl},
DragNUWA~\cite{yin2023dragnuwa} or Tora~\cite{tora2024}. One encoder handles zero,
one or several arrow cues on different objects. Uncued objects receive no explicit directional constraint and infer
their motion from the blur and the shared learned context.

\subsection{Shaping the no-cue default}
\label{sec:narrow_default}

\begin{wrapfigure}{r}{\dimexpr0.44\columnwidth-0.5\columnsep\relax}
\vspace{-12pt}
\hspace*{-0.5\columnsep}\begin{minipage}{\dimexpr\linewidth+0.5\columnsep\relax}
\centering
\includegraphics[width=\linewidth]{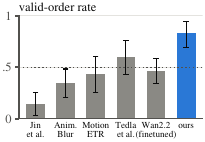}
\vspace{-16pt}
\caption{Blind, 58 fresh drop pairs; ties count as failures; dotted: chance.}
\label{fig:motiv_default}
\end{minipage}
\vspace{-12pt}
\end{wrapfigure}
\Cref{sec:cross_observation_adaptation,sec:ambiguity_training} let the
encoder run without cues but give its features no preference between a valid drop
and its reversal: trained without physical clips, its features prefer the valid order in
only half of the fresh test pairs (\cref{sec:res_default}). We therefore add
synthetic falling-object groups that pair a valid settling sequence, in which a
dropped object falls and comes to rest, with invalid orders that give the same
blur, chiefly the energy-gaining reversal of a passive object. On them a one-sided
ranking loss $\mathcal L_{\mathrm{prior}}$ asks the prediction to be nearer the
valid order's teacher features than each invalid one by a margin. It
acts as a margin regularizer, widening the rendered margin between the orders
without raising the share of correctly ordered fresh pairs (\cref{sec:res_default}).
Cued predictions are exempt, so an arrow cue can override the default. In summary, uncued examples use $\mathcal L_{\mathrm{inv}}$, the settling
subset also uses $\mathcal L_{\mathrm{prior}}$, and cued examples regress directly
on $\mathbf{T}(\mathbf{X})$.

\subsection{Rendering from inferred dynamics features}
\label{sec:feature_renderer}

\begin{figure*}[t]
\centering
\includegraphics[width=\textwidth]{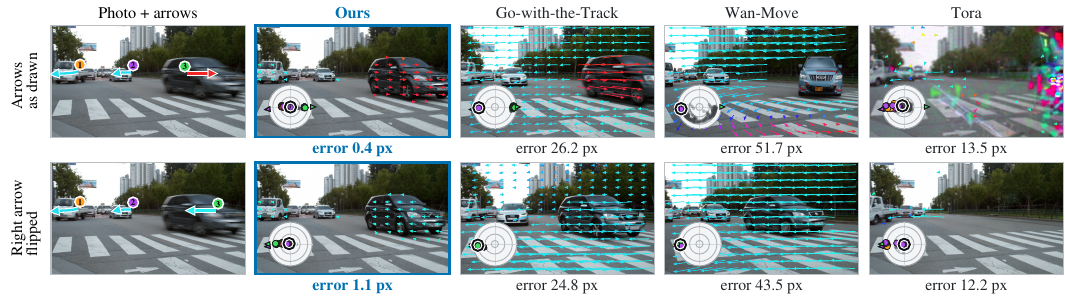}
\caption{Arrow cues set the temporal direction of the parts they mark: arrow cues as drawn (top) and with the right arrow cue flipped
(bottom). \paper reverses only the flipped car; Go-with-the-Track~\cite{gowiththetrack2026} and Wan-Move rewrite the rest
of the scene, and Tora barely reacts. Dials: tracked displacement by direction and log distance, coloured by the nearest
arrow cue; below: track error to the commanded motion. Held-out clip.}
\label{fig:more_qualitative}
\end{figure*}
\noindent We implement the second learned factor,
$\widehat{\mathbf{X}}\sim p_\omega(\mathbf{X}\mid \mathbf{Y},\widehat{\mathbf{Z}}')$,
with Wan~2.2 TI2V-5B, adapted with rank-64 LoRA and
trained with flow matching~\cite{lipman2023flow,liu2023flow} (50 sampling steps). $\mathbf{Y}$ enters through Wan's
image-conditioning slot and a zero-initialized embedding that tiles the blur
latent across frames. During training, a frozen V-JEPA supplies
$\mathbf{T}(\mathbf{X})$ with Gaussian noise and token dropout; at inference it
is replaced by $\widehat{\mathbf{Z}}'$. The full grid of final-layer tokens is read by zero-initialized cross-attention in every DiT block, with a
learned bias toward spatially closer keys. The direct $\mathbf{Y}$
path carries mainly appearance and $\widehat{\mathbf{Z}}'$ mainly the motion.

\section{Experiments}
\label{sec:experiments}

\textbf{Baselines.} The blind baselines are Tedla et al.~\cite{tedla2025generating}, Motion-ETR, Animation from Blur,
Jin et al., deblurring followed by Wan2.2 image-to-video, and a blur-copy control. Outside methods use their authors'
public weights. Animation
from Blur, trained on B-AIST++ only, runs zero-shot elsewhere with its own guide (Table~\ref{tab:main}'s guided block: a dense guide from the true motion). For local control,
Wan-Move~\cite{wanmove2025} and Tora~\cite{tora2024} get the blur as first frame and each arrow cue as one point track as long as the
region's true travel; Animation from Blur gets its guide with each cued region set to the arrow cue's direction. Oracle features (from the true sharp video) only diagnose feature inference.

\textbf{Data and metrics.} We train on blur synthesized from the training recordings of GoPro, Adobe240, X4K1000FPS,
FMO, TbD and Tedla et al.'s clips, plus synthetic scenes. Test sets: GoPro: 22 windows of 11 test
recordings. Adobe240: 154 windows from Tedla et al.'s test recordings, unseen by us and by Tedla et al.
I2-2000FPS~\cite{chennuri2024quiver}: 31 windows of 2000\,fps videos no method trained on, motion-matched to GoPro. RBI~\cite{zhong2023blur}: 115 windows of 5 test videos with real camera blur. GoPro and Adobe240 blur follows our training recipe. Outputs have nine frames at $1280\times736$. EPE is
the end-point error (px) between SEA-RAFT~\cite{wang2024searaft} optical flow of consecutive output frames and of the true frames, averaged over all pixels.

\section{Results}

Our central claim is that a blur determines motion through editable dynamics features inferred from it, which the
renderer follows (Sec.~\ref{sec:res_probes}). We first show the outcome without cues (Sec.~\ref{sec:res_heldout})
and with them (Sec.~\ref{sec:res_control}).

\subsection{Reconstruction without cues}
\label{sec:res_heldout}

\begin{figure*}[t]
\centering
\includegraphics[width=\textwidth]{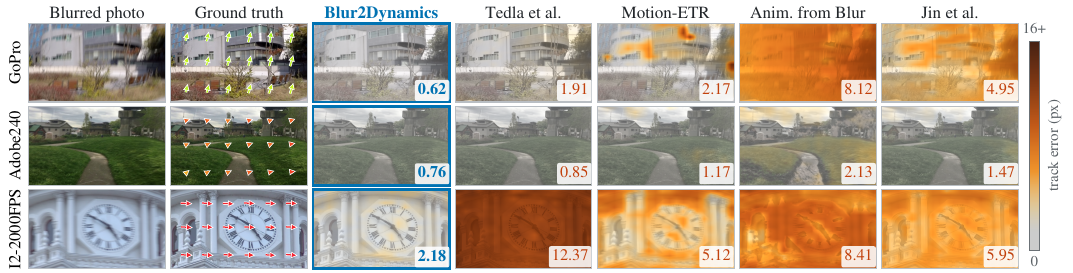}
\caption{Blind reconstruction on held-out clips: each method's middle frame tinted by its per-point track error (CoTracker3~\cite{karaev2025cotracker3}); arrows on
the ground truth show the true motion ($4\times$). Insets: mean track error (px) in the better time order, as in Table~\ref{tab:main}; colour bar: tint scale.}
\label{fig:qualitative}
\end{figure*}

Without cues, \paper recovers motion more faithfully than prior blur-to-video methods on GoPro, Adobe240,
I2-2000FPS and real camera blur (RBI), at competitive image quality; on GoPro its flow error is about half that of the best baseline
(Table~\ref{tab:main}, Fig.~\ref{fig:qualitative}). Since a blur fits a sequence and its reversal, blind EPE scores the
better order: it measures motion up to reversal, not the chosen direction.

\begin{figure}[t]
\centering
\includegraphics{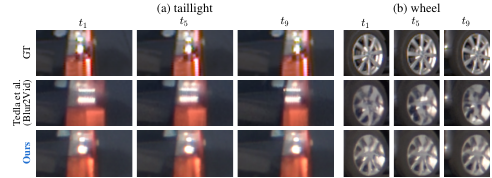}
\caption{Real 18\,ms camera blur (RBI, Table~\ref{tab:main}), frames $t_1$, $t_5$, $t_9$: (a) Tedla et al.'s taillight
stays a streak; (b) its wheel spokes change shape between frames. \paper keeps both shapes.}
\label{fig:rbi}\label{tab:rbi}
\end{figure}

The difference is visible on real camera blur, unlike the synthetic blur we train on
(RBI~\cite{zhong2023blur}, Fig.~\ref{fig:rbi}): Tedla et al.\ leaves parts of moving objects smeared (a car's taillight) and renders
unrealistic motion (wheel spokes that change shape instead of turning), while \paper recovers sharp, moving objects
(supplement). Conditioning on the blur itself matters: deblurring first (Deblur + I2V) removes the streaks that
constrain the motion, and its motion error is close to a blur copy's.

\phantomsection\label{sec:res_default}Without arrow cues, the model chooses the order itself. On new falling-object scenes, where settling is more plausible
than its reversal, \paper keeps the settling order in 82.8\% of pairs, against at most 60.3\% for prior methods
(Fig.~\ref{fig:motiv_default}). The physical training clips set this preference: trained without them, the features
favor neither order (\cref{sec:narrow_default}).%
\begin{figure}[t]
\centering
\includegraphics[width=\columnwidth]{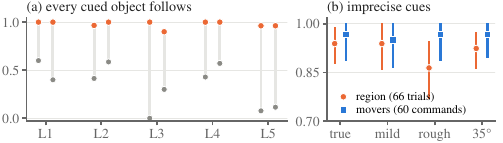}
\caption{Cue control. (a) Synthetic scenes by motion level: share of windows in which every cued object follows, with
cues (orange) and without (grey), for two opposite commands. (b) Real clips with true, perturbed or rotated cues. Bars:
95\% CIs.}
\label{fig:control}
\end{figure}%
\begin{table*}[t]
\centering
\caption{Held-out comparison (one sampler seed). EPE: optical-flow end-point error (px; Sec.~\ref{sec:experiments}). Blind rows take the better time order per clip, guided rows the commanded order. RBI: real camera blur. Bold /
underline: best / second within each ranked block. Rows: Sec.~\ref{sec:res_probes}.}
\label{tab:main}
\renewcommand{\arraystretch}{0.9}
\resizebox{\textwidth}{!}{%
\begin{tabular}{l ccc ccc ccc ccc}
\toprule
& \multicolumn{3}{c}{GoPro} & \multicolumn{3}{c}{Adobe240} & \multicolumn{3}{c}{I2-2000FPS} & \multicolumn{3}{c}{RBI (real blur)} \\
\cmidrule(lr){2-4}\cmidrule(lr){5-7}\cmidrule(lr){8-10}\cmidrule(lr){11-13}
Method & PSNR$\uparrow$ & LPIPS$\downarrow$ & EPE$\downarrow$ & PSNR$\uparrow$ & LPIPS$\downarrow$ & EPE$\downarrow$ & PSNR$\uparrow$ & LPIPS$\downarrow$ & EPE$\downarrow$ & PSNR$\uparrow$ & LPIPS$\downarrow$ & EPE$\downarrow$ \\
\midrule
Motion-ETR & 25.96 & 0.171 & \second{1.00} & 25.87 & 0.160 & 1.55 & 27.26 & 0.230 & 3.58 & 24.06 & 0.250 & 2.20 \\
Animation from Blur & 22.67 & 0.302 & 2.23 & 23.58 & 0.263 & 2.35 & 25.28 & 0.294 & 4.41 & 22.49 & 0.296 & 4.14 \\
Jin et al. & 25.87 & 0.188 & 1.61 & 26.61 & 0.168 & 1.77 & 27.35 & 0.222 & 3.83 & 24.55 & 0.237 & 3.47 \\
Tedla et al. & \second{26.23} & \second{0.129} & 1.10 & \best{26.81} & \second{0.099} & \second{1.29} & \second{27.60} & \second{0.147} & \second{2.22} & \second{26.34} & 0.193 & \second{1.97} \\
Deblur + I2V & 25.72 & 0.130 & 2.14 & 26.78 & \best{0.093} & 2.24 & 26.57 & 0.164 & 4.40 & 24.47 & \second{0.192} & 3.20 \\
\oursrow\ours{\paper (blind)} & \best{26.41} & \best{0.121} & \best{0.57} & \second{26.80} & 0.115 & \best{0.69} & \best{29.93} & \best{0.116} & \best{1.58} & \best{26.54} & \best{0.164} & \best{1.09} \\
\midrule
\multicolumn{13}{l}{\textit{With motion guidance, scored in the commanded order}} \\%
Animation from Blur + true-motion guide & 20.64 & 0.338 & 1.53 & 20.90 & 0.318 & 1.89 & 23.81 & 0.326 & 3.56 & 22.55 & 0.287 & 3.33 \\
\oursrow\ours{\paper + arrow cues} & \best{26.71} & \best{0.117} & \best{0.46} & \best{27.14} & \best{0.110} & \best{0.56} & \best{30.26} & \best{0.112} & \best{1.43} & \best{26.63} & \best{0.162} & \best{1.03} \\
\midrule
\multicolumn{13}{l}{\textit{Not ranked: control, oracle diagnostic and ablation}} \\%
Blur copy & 24.46 & 0.291 & 2.42 & 25.70 & 0.243 & 2.47 & 27.00 & 0.257 & 4.55 & 24.00 & 0.265 & 3.12 \\%
\paper + oracle features & 27.07 & 0.104 & 0.34 & 27.79 & 0.097 & 0.42 & 31.00 & 0.092 & 1.15 & 27.10 & 0.136 & 0.83 \\%
\paper, frozen encoder & 24.38 & 0.277 & 2.29 & 25.25 & 0.238 & 2.37 & 26.98 & 0.257 & 4.48 & 24.14 & 0.268 & 3.09 \\%
\paper, features zeroed & 25.17 & 0.222 & 1.75 & 25.84 & 0.183 & 1.58 & 27.78 & 0.205 & 3.34 & 25.17 & 0.232 & 2.22 \\%
Wan2.2 (finetuned) & 25.68 & 0.138 & 0.88 & 26.48 & 0.124 & 0.69 & 29.19 & 0.134 & 1.74 & 26.09 & 0.174 & 1.09 \\%
\bottomrule
\end{tabular}}
\end{table*}%

\subsection{Control with arrow cues}
\label{sec:res_control}

\begin{table}[t]
\centering
\caption{Local control on real held-out clips: two movers with opposite arrow cues (30 clips, 60 commands) and one region reversed against
the rest (66 trials, 19 recordings). Follow: every commanded part moves as commanded. Error: median track error (px) to the
commanded motion.}
\label{tab:control}
\footnotesize
\renewcommand{\arraystretch}{0.9}
\setlength{\tabcolsep}{4pt}
\begin{tabular}{l cc cc}
\toprule
& \multicolumn{2}{c}{Two movers (60)} & \multicolumn{2}{c}{Region reversed (66)} \\
\cmidrule(lr){2-3}\cmidrule(lr){4-5}
Method & Follow$\uparrow$ & Error$\downarrow$ & Follow$\uparrow$ & Error$\downarrow$ \\
\midrule
Wan-Move 14B & 58.3\% & 4.58 & 33.3\% & 5.96 \\
Tora & 35.0\% & 4.63 & 22.7\% & 8.99 \\
Go-with-the-Track & \second{93.3\%} & \second{1.75} & \second{65.2\%} & \second{3.33} \\
Animation from Blur & 83.3\% & 1.85 & 27.3\% & 4.44 \\
\oursrow\ours{\paper} & \best{96.7\%} & \best{1.27} & \best{93.9\%} & \best{1.64} \\
\bottomrule
\end{tabular}
\end{table}

In practice, a user should be able to choose which way each object moves. The blur leaves this open, so a sparse
arrow cue sets it without prescribing a trajectory or speed (Fig.~\ref{fig:more_qualitative}), even when objects
move in different directions.

\noindent\textbf{Local control on real clips.}\enspace
Cues control each region separately. With opposite cues on two moving regions, \paper follows both more often
and more accurately than trajectory-controlled generators given the cues as point tracks (Table~\ref{tab:control}).
The gap widens when one region is played backwards against the rest of the clip
(93.9\% of trials, against at most 65.2\%). Cues need not be precise: rotated or roughly drawn
ones still work (Fig.~\ref{fig:control}b).

Cues drawn by people work too. Nine participants freely drew arrows on 16 held-out images, and the cued region
follows its cue in 93\% of trials. When one cue is reversed, our method reverses just that region in 75\% of
trials, against 51\% for Tora given the same arrows as point tracks (supplement).

\noindent\textbf{Across motion types.}\enspace
On synthetic scenes, where the true motion is known exactly, we vary the camera motion, the number of objects and
where the cues go (Fig.~\ref{fig:control}a). When 2--3 objects are told to move in opposite directions, all of them
follow in 95.8\% of commands. Without cues, each object's direction is ambiguous in the blur and cannot be resolved,
so the output matches only 11.1\% of commands.

\subsection{Role of dynamics features}
\label{sec:res_probes}
\label{sec:res_ablations}

\begin{figure}[t]
\centering
\includegraphics[width=\columnwidth]{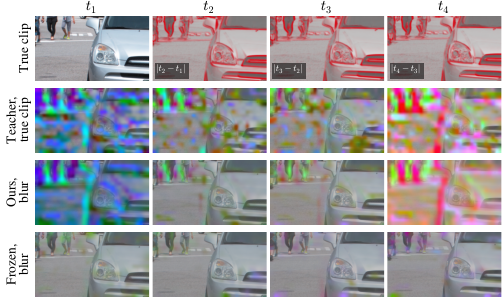}
\caption{Feature change over time on a real held-out clip: top-3 PCA colours of each encoder's change across its four
time steps. Top row: first frame, then frame-to-frame change (red). Ours on the blur changes where the people walk, like
the teacher on the true clip; frozen V-JEPA barely changes.}
\label{fig:pca}
\end{figure}

Our approach needs the renderer to respond to the features and the encoder to infer these useful features from a
blur. We test this from the oracle and ablation rows of Table~\ref{tab:main}.

\noindent\textbf{Feature quality.}\enspace
Features of the true video give the best results everywhere, while frozen V-JEPA features of the blur, or no
features at all, push the output toward a blur copy (Table~\ref{tab:main}). The renderer therefore relies on the
features. The adaptation of V-JEPA is what makes features of a blur useful. The adapted features also carry motion that
frozen ones lack. This is illustrated in Fig.~\ref{fig:pca}, where we can see that our features change where the
people walk, like the sharp video's features, while frozen V-JEPA's barely change. A linear probe also reads the blur's motion
axis better from our features than from frozen V-JEPA's features of the same blur (Fig.~\ref{fig:motiv_probes}).

Without features, Wan2.2 fine-tuned the same way comes close in blind motion, but loses fidelity and cannot take
cues. But our method with cues has a lower EPE than the fine-tuned Wan2.2 on every set, about half on GoPro. The features thus bring both fidelity and control.

\begin{figure}[t]
\centering
\includegraphics[width=\columnwidth]{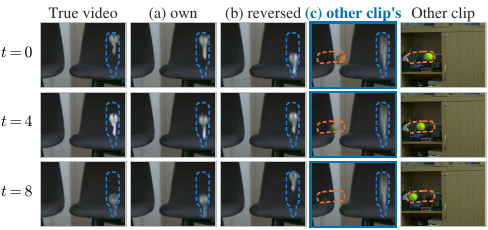}
\caption{Feature interventions on a blurred falling key (held-out scene). Given teacher features of (a) its own video,
the key falls; (b) the reversed video, it rises; (c) a clip of a ball flying left, it stays a still streak, with faint
motion where the ball moved (orange). Blue: the key's region.}
\label{fig:interventions}
\end{figure}

\noindent\textbf{Changing the features.}\enspace
Swapping a video's teacher features for those of its reversal reverses the generated motion in all 14 clips. With
another clip's features, the key stays still and faint motion appears where that clip's object moved, and the
renderer even makes the key rise instead of fall (Fig.~\ref{fig:interventions}). The features thus decide where and
which way things move; since the renderer accepts implausible dynamics, plausibility has to come from feature inference.

\begin{figure}[t]
\centering
\includegraphics[width=\columnwidth]{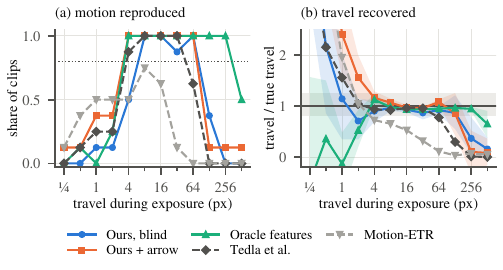}
\caption{Recoverable motion range for a textured 128-px object, 8 clips per travel. (a) Share of clips whose motion is
reproduced: the object is tracked in every frame, travels at least half the true distance and stays within $30^\circ$ of the
true direction (dotted: the 80\% pass level); (b) generated travel over true travel (1: correct; 0.8--1.25: close; bands: 95\%
bootstrap CIs, solid lines only). Oracle: features of the true video. Dashed: baselines, blind.}
\label{fig:range}
\end{figure}

\phantomsection\label{sec:res_range}\noindent\textbf{Recoverable motion range.}\enspace
As a textured object travels farther during the exposure (Fig.~\ref{fig:range}), our method recovers its motion from 8
to 64\,px, whereas Tedla et al.\ falls off beyond 32\,px. With features of the true video, the same renderer reaches
256\,px, so for severe blur the bottleneck is inferring the features, not rendering them.

\noindent\textbf{Temporal controllability.}\enspace
The number of output frames can be changed without retraining. When the encoder reads the blur 4 or 16 times instead
of 8, the renderer produces 5 or 17 frames. Cues still control the time order at both lengths in
almost every window, and with 17 frames the motion between frames shrinks from 3.4 to 2.1\,px, giving a smoother
video of the same exposure (supplement).

\section{Conclusion}

Motion blur is not only a degradation to remove: it records how the scene moved during the exposure. \paper reads
this record with a V-JEPA encoder adapted to predict, from a single blurred image, the features its frozen copy
extracts from the sharp video, and a video generator renders the sharp frames from the blur and these features.
Without cues, our method recovers motion more accurately than prior blur-to-video methods at competitive image
quality. When the blur cannot tell which way an object moved, a simple arrow lets the user choose, while the
blur still decides how far and along which path it moves. Because the generated video follows the dynamics features, users can control it by
changing them; the central task is therefore to infer the correct dynamics features from the blurred image.

\clearpage
\section*{LLM Usage Statement}
LLMs assisted in writing code, running and documenting experiments, and editing the text and figures; the authors
checked all results and claims.

{\small
\bibliographystyle{ieeenat_fullname}
\bibliography{references}
}

\clearpage
\flushbottom
\renewcommand{\encodingdefault}{T1}\fontencoding{T1}\selectfont
\IfFileExists{colortbl.sty}{\renewcommand{\oursrow}{\rowcolor{oursbg}}}{}
\renewcommand{\dbltopfraction}{0.9}\renewcommand{\topfraction}{0.9}\renewcommand{\textfraction}{0.07}
\renewcommand{\dblfloatpagefraction}{0.8}\renewcommand{\floatpagefraction}{0.8}
\setcounter{dbltopnumber}{4}\setcounter{topnumber}{4}
\setlength{\textfloatsep}{8pt plus 2pt minus 2pt}\setlength{\dbltextfloatsep}{8pt plus 2pt minus 2pt}
\setlength{\floatsep}{6pt plus 2pt minus 2pt}\setlength{\dblfloatsep}{6pt plus 2pt minus 2pt}
\setlength{\intextsep}{6pt plus 2pt minus 2pt}\setcounter{bottomnumber}{3}\renewcommand{\bottomfraction}{0.6}
\setcounter{section}{0}\setcounter{figure}{0}\setcounter{table}{0}\setcounter{equation}{0}\setcounter{footnote}{0}
\renewcommand{\theequation}{S\arabic{equation}}\renewcommand{\theHequation}{S\arabic{equation}}
\renewcommand{\thetable}{S\arabic{table}}\renewcommand{\theHtable}{S\arabic{table}}
\renewcommand{\thefigure}{S\arabic{figure}}\renewcommand{\theHfigure}{S\arabic{figure}}
\renewcommand{\thesection}{S\arabic{section}}\renewcommand{\theHsection}{S\arabic{section}}
\addtocontents{toc}{\protect\setcounter{tocdepth}{2}}
\maketitlesupplementary
\begingroup
\makeatletter
\renewcommand*\l@section{\@dottedtocline{1}{0em}{2.1em}}%
\renewcommand*\l@subsection{\@dottedtocline{2}{0.6em}{2.9em}}%
\makeatother
\setcounter{tocdepth}{2}
\renewcommand{\contentsname}{Contents}
\small
\setlength{\parskip}{0pt}
\tableofcontents
\endgroup
\medskip
\noindent\textbf{Table conventions.} $\uparrow$/$\downarrow$: higher/lower is better. PSNR (dB),
LPIPS~\cite{zhang2018lpips}, EPE (optical-flow end-point error, px), Track
(point-track error, px); details in \cref{sec:supp_metrics}. Brackets: 95\% cluster-bootstrap interval; a difference is
significant when its interval does not contain 0 (uncorrected). (n): number of windows. N/A: not applicable; --: not scored.
\bigskip

\section{Problem formulation: from image formation to a learned inverse}
\label{sec:supp_pgm}

\begin{figure*}[t]
\centering
\includegraphics[width=\textwidth]{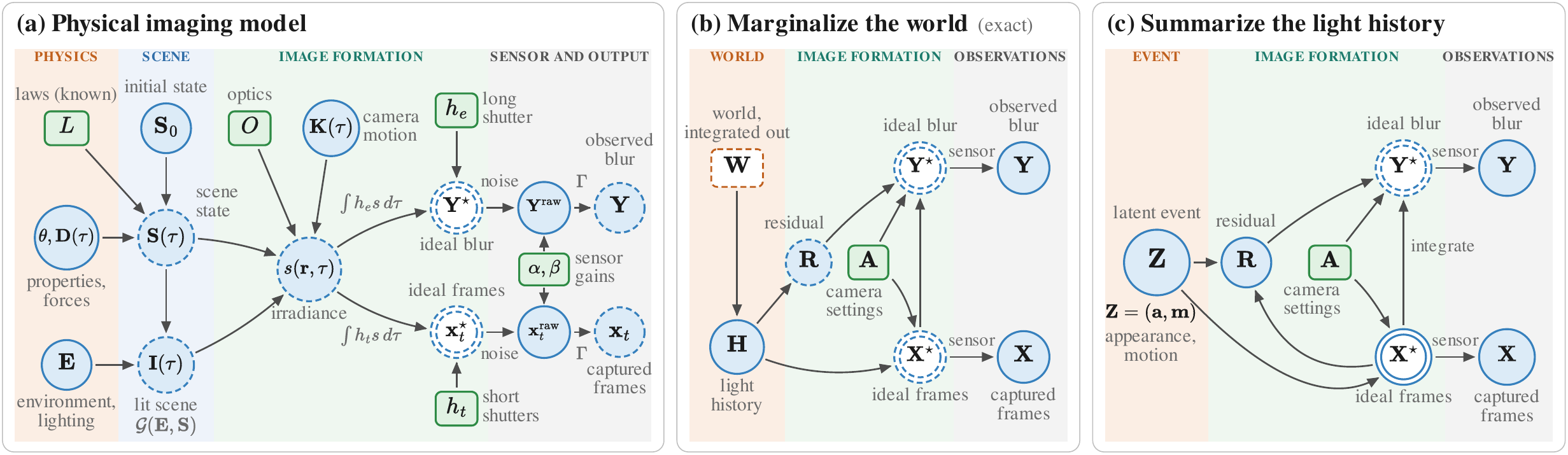}
\caption{\textbf{Integrating out the world is exact: the blur depends on the scene only through its light history and the latent event that summarizes it.} The forward model in three steps. (a) The physical imaging model (clipping and quantization not drawn). (b) The world
$\mathbf W$ integrated out exactly: only the light history $\mathbf H$ reaches the camera. (c) A latent event
$\mathbf Z=\psi(\mathbf H)$ summarizes $\mathbf H$, giving main-paper \cref{fig:background_pgm}(a).
Solid circle: random variable; dashed: deterministic given its parents; double: ideal signal before the sensor;
green box: known quantity.}
\label{fig:supp_pgm_steps}
\end{figure*}

\begin{figure*}[t]
\centering
\includegraphics[width=\textwidth]{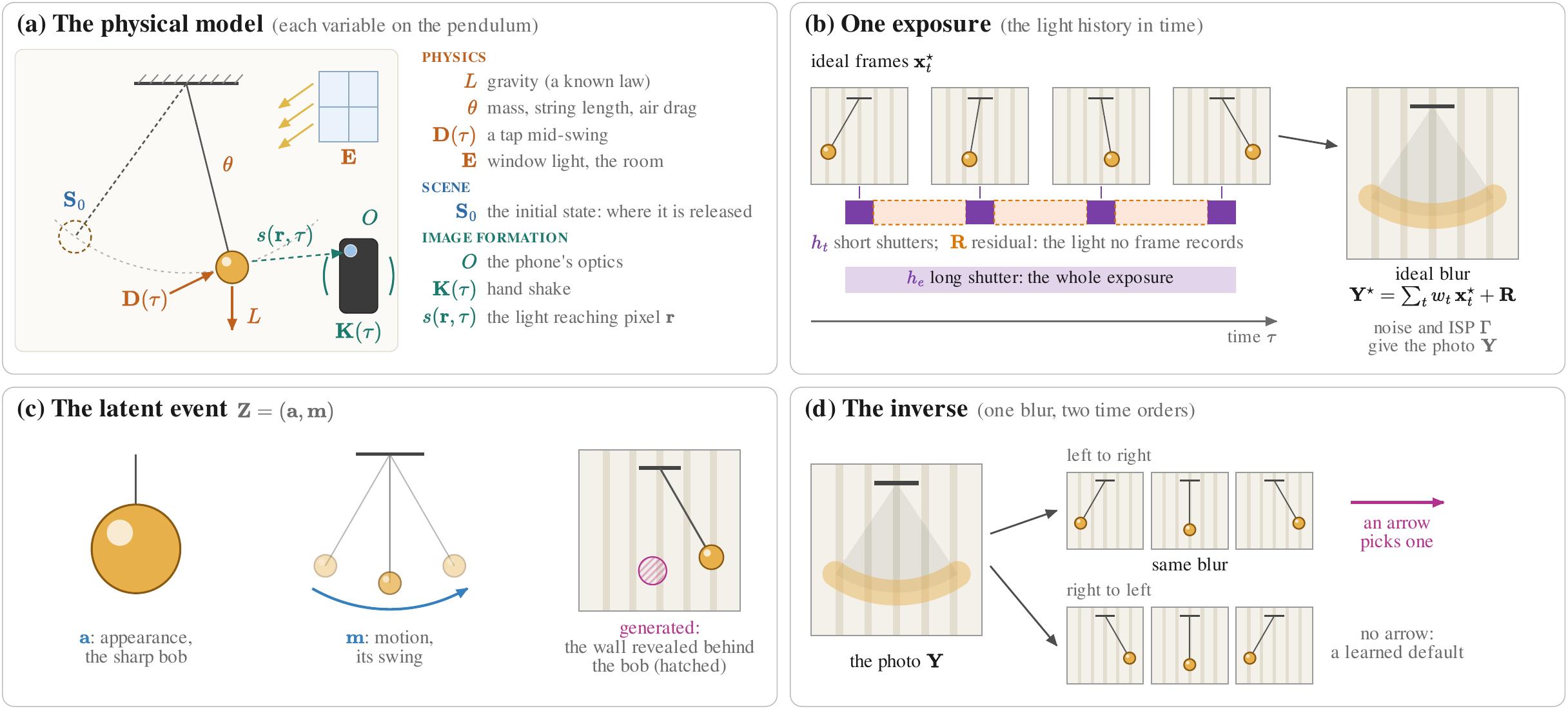}
\caption{\textbf{A swing and its reversal give the same blur, so one blur leaves the time direction open; an arrow or the learned default selects it.} The running example: a pendulum filmed with a handheld phone (\cref{sec:supp_pgm_physical,sec:supp_pgm_world,sec:supp_pgm_latent,sec:supp_pgm_inverse}).
(a) The variables of \cref{fig:supp_pgm_steps}(a), coloured by zone. (b) One exposure: short frames, the residual $\mathbf R$
between them, and their sum. (c) The latent event $\mathbf Z=(\mathbf a,\mathbf m)$: the sharp bob and its swing, not the
wall revealed behind it. (d) The swing and its reversal give the same blur; an arrow or a learned default picks one.}
\label{fig:supp_pendulum}
\end{figure*}

This section gives the full model behind main-paper \cref{sec:model} and the reasoning that
leads from it to our method. We build the forward model in three steps (\cref{fig:supp_pgm_steps}):
(a) every physical variable that shapes a blurred photograph, (b) an exact marginalization of the
world into the light that reaches the sensor, and (c) a summary of that light by a latent event.
Step (c) gives the graph of main-paper \cref{fig:background_pgm}(a). We then derive the reduced
joint distribution and state what the inverse requires; it motivates each design decision of the method. \Cref{tab:supp_approx} lists every step, whether it is exact, where it enters our
system and when it breaks. As a running example, take a pendulum filmed with a handheld phone
(\cref{fig:supp_pendulum}).

\subsection{The physical imaging model}
\label{sec:supp_pgm_physical}
\Cref{fig:supp_pgm_steps}(a) names every variable and \cref{fig:supp_pendulum}(a) shows each on the
pendulum. Known laws $L$ (gravity), unknown properties
$\theta$ (the bob's mass, string length, air drag), external forces $\mathbf D(\tau)$ (a tap
mid-swing) and an initial state $\mathbf S_0$ fix the scene state $\mathbf S(\tau)$ over time.
The environment $\mathbf E$ (window light, the room) lights it, giving
$\mathbf I(\tau)=\mathcal G(\mathbf E,\mathbf S(\tau))$. Optics $O$ and camera motion
$\mathbf K(\tau)$ (hand shake) project the lit scene onto the sensor as the irradiance
$s(\mathbf r,\tau)$, the light reaching pixel $\mathbf r$ at time $\tau$. Everything the camera can
record comes from this one field. A long exposure integrates it through the shutter response $h_e$,
\begin{equation}
    \mathbf{Y}^\star(\mathbf r)=
    \int h_e(\mathbf r,\tau)\,s(\mathbf r,\tau)\,d\tau,
    \label{eq:supp_long_exposure}
\end{equation}
and each short frame $\mathbf x_t^\star$ integrates it through its own response $h_t$. The sensor
adds noise to each ideal signal, giving RAW values. The image signal processor (ISP) $\Gamma$
(demosaicing, tone curve, compression) then turns them into the observed blur $\mathbf Y$ and the
captured frames $\mathbf x_t$. Only $\mathbf Y$ is observed at inference; the frames are training
targets. The camera settings $\mathbf A$ collect the shutter responses and
sensor parameters. Each test benchmark fixes them, and neither network receives them; training mixes sources with
different frame rates and exposures, so the learned networks implicitly average over $\mathbf A$.

\subsection{Marginalizing the world}
\label{sec:supp_pgm_world}
The world $\mathbf W=(\theta,\mathbf D,\mathbf S_0,\mathbf E,\mathbf K)$ determines the complete
light history
\begin{equation}
\begin{split}
    \mathbf H&=\{s(\mathbf r,\tau)\}_{\tau\in[0,T_e]}=\mathcal F_L(\mathbf W),\\
    p(\mathbf H)&=\int\delta\bigl(\mathbf H-\mathcal F_L(\mathbf W)\bigr)\,p(\mathbf W)\,d\mathbf W,
\end{split}
    \label{eq:supp_light_history}
\end{equation}
where $\mathcal F_L$ is the deterministic dynamics, lighting and projection under the laws $L$.
Integrating $\mathbf W$ out is exact for any $p(\mathbf W)$, because both ideal signals depend on $\mathbf W$ only
through $\mathbf H$. It preserves the joint distribution of $\mathbf Y^\star$ and $\mathbf X^\star$, and it
needs no independence between camera and object motion (\cref{fig:supp_pgm_steps}(b)). The graph of
\cref{fig:supp_pgm_steps}(a) assumes only that camera motion changes image formation, not the scene's dynamics (no
edge from $\mathbf K$ to $\mathbf S$). The price is that
$p(\mathbf H)$ integrates over every possible world and has no closed form. This prior is what any
inverse must supply.

\paragraph{Sensor readout.}
For calibrated linear RAW, shot and read noise are commonly approximated as
$p(\mathbf Y^{\mathrm{raw}}\mid\mathbf Y^\star)\approx
\mathcal N\bigl(\mathbf Y^\star,\operatorname{diag}(\alpha\mathbf Y^\star+\beta)\bigr)$, with gain
$\alpha$ and read-noise variance $\beta$, and the short frames with their own parameters. Our data
are in sRGB (the standard RGB colour space), so $\Gamma$ cannot be undone exactly. We stand in for its inverse with the sRGB transfer
function and model no sensor likelihood. Training instead adds shot noise to the sRGB blur and JPEG compression as
augmentation (\cref{tab:supp_training}), not a RAW noise model. Clipping happens on the RAW values before $\Gamma$; the
high-frame-rate source frames are already clipped and tone-mapped, so averaging them underestimates bright streaks.

\paragraph{Light between frames.}
Short frames need not cover the whole exposure. Splitting the integral of
\cref{eq:supp_long_exposure} at the frame windows is exact and gives main-paper
Eq.~\eqref{eq:exposure_gap},
$\mathbf Y^\star=\sum_{t}w_t\mathbf x_t^\star+\mathbf R$. The sum holds the weighted ideal frames.
With box shutters whose frame windows are disjoint and inside the exposure, $h_e=w_t h_t$ on each window, and the
residual $\mathbf R$ is the light that falls between the windows, which no frame records (\cref{fig:supp_pendulum}(b)):
non-negative, and zero when the windows tile the exposure. With other shutters $\mathbf R$ also holds in-window
differences and can be negative. A rolling shutter makes the windows, and so $w_t$, depend on the row; we assume a
global shutter. Training pairs are built from
high-frame-rate clips, so $\mathbf R$ must be filled in (\cref{sec:supp_synth_blur}).

\subsection{Summarizing the light history}
\label{sec:supp_pgm_latent}
$\mathbf H$ is far too large to estimate from one photograph. For any summary
$\mathbf Z=\psi(\mathbf H)$, $p(\mathbf X^\star\mid\mathbf A)=\int p(\mathbf Z)\,
p(\mathbf X^\star\mid\mathbf Z,\mathbf A)\,d\mathbf Z$ is the chain rule: it asserts no independence, so $\mathbf Z$ only
has to be informative, and what $\mathbf Z$ means is an interpretation. We write
$\mathbf Z=(\mathbf a,\mathbf m)$ to separate the roles it plays: $\mathbf a$ anchors appearance
(the sharp bob) and $\mathbf m$ summarizes motion (its swing). Content that neither determines,
such as background revealed behind the bob, must be generated (\cref{fig:supp_pendulum}(c)). This gives the graph of
\cref{fig:supp_pgm_steps}(c). $\mathbf Z$ is a conceptual separation, not a complete physical
state. The forward model is never fitted to data and its exposure likelihood is never evaluated, so consistency with the
blur is learned rather than enforced.

\paragraph{Graph.}
The graph has the latent event $\mathbf{Z}$, the ideal sharp sequence $\mathbf{X}^\star$, the
residual $\mathbf{R}$, the ideal blur $\mathbf{Y}^\star$, the observed blur
$\mathbf{Y}$, the captured frames $\mathbf{X}$, and the known camera settings $\mathbf{A}$. Its
edges are $\mathbf{Z}\!\rightarrow\!\mathbf{X}^\star$, $\mathbf{Z}\!\rightarrow\!\mathbf{R}$,
$\mathbf{X}^\star\!\rightarrow\!\mathbf{R}$, $\mathbf{X}^\star\!\rightarrow\!\mathbf{Y}^\star$,
$\mathbf{R}\!\rightarrow\!\mathbf{Y}^\star$, $\mathbf{A}\!\rightarrow\!\mathbf{X}^\star$ (the
camera sets the frame times), $\mathbf{A}\!\rightarrow\!\mathbf{Y}^\star$,
$\mathbf{Y}^\star\!\rightarrow\!\mathbf{Y}$ and $\mathbf{X}^\star\!\rightarrow\!\mathbf{X}$.
The shutter responses in $\mathbf A$ also shape $\mathbf R$, and its noise parameters enter both readouts; the figure
and the factors $p(\mathbf R\mid\mathbf Z,\mathbf X^\star)$, $p(\mathbf Y\mid\mathbf Y^\star)$ and
$p(\mathbf X\mid\mathbf X^\star)$ leave this implicit, which is exact for the fixed $\mathbf A$ of a benchmark. We also
assume the event does not depend on the settings, $p(\mathbf Z\mid\mathbf A)=p(\mathbf Z)$; auto-exposure that adapts
to the scene breaks this.

\paragraph{Full joint.}
$\mathbf{A}$ is known, so every distribution is conditioned on it. The chain rule in topological
order, with each node conditioned only on its parents, gives
\begin{equation}
\begin{split}
    &p(\mathbf{Z},\mathbf{X}^\star,\mathbf{R},\mathbf{Y}^\star,\mathbf{Y},\mathbf{X}\mid\mathbf{A})\\
    &\quad=\underbrace{p(\mathbf{Z})}_{\text{event}}\,
         \underbrace{p(\mathbf{X}^\star\mid\mathbf{Z},\mathbf{A})}_{\text{sharp frames}}\,
         \underbrace{p(\mathbf{R}\mid\mathbf{Z},\mathbf{X}^\star)}_{\text{light between frames}}\\[2pt]
    &\qquad\;\underbrace{p(\mathbf{Y}^\star\mid\mathbf{X}^\star,\mathbf{R},\mathbf{A})}_{\text{ideal blur}}\,
         \underbrace{p(\mathbf{Y}\mid\mathbf{Y}^\star)}_{\text{sensor and ISP}}\,
         \underbrace{p(\mathbf{X}\mid\mathbf{X}^\star)}_{\text{frame readout}}.
\end{split}
\label{eq:supp_full_joint}
\end{equation}
The last factor treats the captured frames as a readout of $\mathbf{X}^\star$ separate from the
blur, with its own noise, as in physical paired capture through a beam splitter; our synthetic pairs instead compute $\mathbf{Y}$ from
$\mathbf{X}$ (\cref{sec:supp_synth_blur}). We need the forward model to train: large registered datasets of real blurred
images with the sharp video behind them are not available, so the forward model forms each training blur from
high-frame-rate video, estimating the residual $\mathbf R$ between the captured frames, and these pairs train both the
encoder and the renderer.

\paragraph{Ideal blur and residual.}
The ideal blur is deterministic given its parents. Writing the exposure weights' dependence on
the camera settings as $w_t(\mathbf{A})$,
\begin{equation}
    p(\mathbf{Y}^\star\mid\mathbf{X}^\star,\mathbf{R},\mathbf{A})
    =\delta\Bigl(\mathbf{Y}^\star-\textstyle\sum_{t=1}^{T}
      w_t(\mathbf{A})\,\mathbf{x}_t^\star-\mathbf{R}\Bigr).
    \label{eq:supp_blur_delta}
\end{equation}
$\mathbf{R}$ is set by the continuous light history between the frame windows. $\mathbf{Z}$
compresses that history and $\mathbf{X}^\star$ samples it, so neither need determine
$\mathbf{R}$. Hence $p(\mathbf{R}\mid\mathbf{Z},\mathbf{X}^\star)$ stays a general conditional.

\paragraph{Marginalizing.}
Integrating \cref{eq:supp_full_joint} over $\mathbf{Y}^\star$ collapses the point mass, and
integrating over the unobserved $\mathbf{R}$ absorbs it into the exposure likelihood:
\begin{equation}
\begin{split}
    p(\mathbf{Z},\mathbf{X}^\star,\mathbf{Y},\mathbf{X}\mid\mathbf{A})
    ={}& p(\mathbf{Z})\,p(\mathbf{X}^\star\mid\mathbf{Z},\mathbf{A})\\
       & p(\mathbf{Y}\mid\mathbf{X}^\star,\mathbf{Z},\mathbf{A})\,
         p(\mathbf{X}\mid\mathbf{X}^\star),
\end{split}
\label{eq:supp_reduced}
\end{equation}
\begin{equation}
\begin{split}
    p(\mathbf{Y}\mid\mathbf{X}^\star,\mathbf{Z},\mathbf{A})
    ={}&\int p\Bigl(\mathbf{Y}\,\Big|\,\mathbf{Y}^\star
      =\textstyle\sum_{t=1}^{T}w_t(\mathbf{A})\,\mathbf{x}_t^\star
      +\mathbf{R}\Bigr)\\
     &\;p(\mathbf{R}\mid\mathbf{Z},\mathbf{X}^\star)\,d\mathbf{R},
\end{split}
\label{eq:supp_blur_likelihood}
\end{equation}
the sensor model evaluated at the ideal blur and averaged over the residual. Through
$p(\mathbf{R}\mid\mathbf{Z},\mathbf{X}^\star)$ the blur depends on the event as well as on the
sampled frames, which alone do not fix the light between them. When the frame windows cover the
whole exposure, $\mathbf{R}=\mathbf{0}$ and the factor reduces to
$p(\mathbf{Y}\mid\mathbf{X}^\star,\mathbf{A})$.

\paragraph{Assumptions.} Each step on the way to \cref{eq:supp_reduced} is exact or rests on a stated assumption
(\cref{tab:supp_approx} gives when each breaks):
\begin{itemize}\setlength\itemsep{1pt}
\item \textbf{World $\mathbf W$:} integrated out exactly, since the ideal signals depend on it only through $\mathbf H$.
We assume camera motion changes image formation, not the scene's dynamics; camera and object motion may still be
correlated.
\item \textbf{Light history $\mathbf H$ $\to$ $\mathbf Z$:} the chain rule for any summary $\psi$; no independence is
asserted.
\item \textbf{Camera settings $\mathbf A$:} fixed per test benchmark, independent of the event, and averaged over in
training; global box shutters with disjoint frame windows.
\item \textbf{Ideal blur $\mathbf Y^\star$:} exact, since it is a deterministic function of $\mathbf X^\star$,
$\mathbf R$ and $\mathbf A$ (\cref{eq:supp_blur_delta}).
\item \textbf{Residual $\mathbf R$:} the integral of \cref{eq:supp_blur_likelihood} is exact. To fill it we assume short
gaps and smooth motion, so that $\mathbf R$ given the event and the frames has a small spread around its mean, and we
estimate it by moving neighbouring frames along their motion (interpolation in linear light). Long gaps, occlusion,
lighting change and fast or irregular motion break this.
\item \textbf{Sensor and ISP:} no likelihood is modelled; Poisson--Gaussian noise on calibrated RAW is the reference,
and for sRGB data we invert $\Gamma$ with the sRGB transfer function and augment with noise and JPEG. Clipping, low light
and an unknown ISP break this.
\item \textbf{Captured frames $\mathbf X$:} read out from $\mathbf X^\star$ apart from the blur, with independent noise;
our synthetic pairs compute $\mathbf Y$ from $\mathbf X$ instead, treating each frame as an instantaneous sample.
\end{itemize}

\begin{table}[!htbp]
\centering
\caption{Each step from the physical model to our system: exact (E) or approximate (A), how our system
implements it, and when it breaks.}
\label{tab:supp_approx}
\footnotesize
\setlength{\tabcolsep}{3pt}
\begin{tabular}{@{}L{0.24\columnwidth}cL{0.32\columnwidth}L{0.30\columnwidth}@{}}
\toprule
Step & E/A & In our system & Breaks when \\
\midrule
World $\to$ light history $\mathbf H$ & E & nothing to implement & exact for any $p(\mathbf W)$ \\
Camera does not move the scene & A & no edge $\mathbf K\to\mathbf S$ & camera motion changes the scene \\
Settings $\mathbf A$ & A & fixed per test set; mixed in training & settings outside the training mix; auto-exposure \\
Shutter & A & global, box, disjoint windows & rolling shutter, overlapping or non-box windows \\
Sensor noise, ISP & A & sRGB curve for $\Gamma^{-1}$; noise, JPEG augmentation & clipping, low light, unknown ISP \\
Residual $\mathbf R$ & A & frames interpolated by RIFE~\cite{huang2022rife}, linear-light mean (\cref{sec:supp_synth_blur}) & long gaps, occlusion, lighting change, fast or irregular motion \\
$\mathbf Z=\psi(\mathbf H)$ & E & nothing to implement & exact (chain rule) \\
$\mathbf Z\to$ features $\mathbf Z'$ & A & frozen V-JEPA on the sharp clip & small or weakly encoded objects \\
Posterior $\to$ one estimate & A & adapted encoder $E_\phi$; $\mathcal L_{\mathrm{inv}}$, arrows, $\mathcal L_{\mathrm{prior}}$ & several valid videos remain \\
Renderer ignores arrows $\mathbf C$ & A & arrows act only on $\widehat{\mathbf Z}'$ & the features do not carry the direction the arrow chose \\
\bottomrule
\end{tabular}
\end{table}

\subsection{The inverse and how we approach it}
\label{sec:supp_pgm_inverse}
Recovering the video means inverting this model. Dropping the fixed $\mathbf{A}$, the posterior
under the assumed model is
\begin{equation}
    p(\mathbf{X}^\star\mid \mathbf{Y})\propto
    \int \underbrace{p(\mathbf{Y}\mid \mathbf{X}^\star,\mathbf{Z})}_{\text{consistency with the blur}}\,
    \underbrace{p(\mathbf{X}^\star\mid \mathbf{Z})\,p(\mathbf{Z})}_{\text{plausible dynamics}}\,d\mathbf{Z},
    \label{eq:supp_z_posterior}
\end{equation}
and the captured frames follow as $p(\mathbf X\mid\mathbf Y)=\int p(\mathbf X\mid\mathbf X^\star)\,
p(\mathbf X^\star\mid\mathbf Y)\,d\mathbf X^\star$. Even under the assumed model, the posterior marginalizes an unobserved event. An answer must
therefore agree with the exposure and be a plausible evolution of the scene.

A blurred photograph leaves three things open: the motion
(which regions move, where, and in which direction), the sharp appearance, and content that no
frame shows, such as background uncovered by a moving object. Integration also makes the
direction ambiguous. Under a time-symmetric exposure, a clip and its time reversal give the same
blur (main-paper \cref{sec:model}; for the pendulum, \cref{fig:supp_pendulum}(d)). Spatially separate
motions can each be reversed (a global shutter keeps this symmetry; a rolling shutter breaks it). Each decision of the method answers one of these points.

\paragraph{Infer motion; let the blur carry appearance.}
The blur already shows most of the appearance, so we infer only a representation of the event
and give the renderer the blur itself; a sharp keyframe is not needed (\cref{sec:supp_slot}).

\paragraph{Dynamics features in place of $\mathbf Z$.}
$\mathbf Z$ is never observed, but the dynamics features of the captured clip can be computed:
$\mathbf Z'=E_{\bar\phi}(\mathbf X)$ from a frozen video encoder. They exist for every
training clip and give a dense target. We do not assume that $\mathbf Z'$ equals $\mathbf Z$ or
any part of it. The chain rule over $\mathbf Z'$ is exact, and the learned factors approximate it
(main-paper Eqs.~\eqref{eq:feature_mediated_model}
and~\eqref{eq:deterministic_feature_posterior}).

\paragraph{One estimate under an ambiguous posterior.}
The posterior over $\mathbf Z'$ can have two modes of equal likelihood, the captured order and its reversal,
weighted by a prior that need not be time-symmetric (gravity).
A single estimate trained with squared error toward one of them would average the two. The
order-invariant loss $\mathcal L_{\mathrm{inv}}$ therefore matches whichever order is closer (main-paper
\cref{sec:ambiguity_training}). The remaining choice is left to the user or to the no-cue
default.

\paragraph{Arrows as an intervention.}
The arrows $\mathbf C$ are not part of the forward model. They select one temporal orientation
inside feature inference and never reach the renderer (main-paper \cref{sec:sparse_cues};
\cref{sec:supp_enc_arch}; simulation in \cref{sec:supp_cues}).

\paragraph{The no-cue default.}
When the blur leaves the order open and no arrow is given, physically valid motion is the
better guess. The ranking loss $\mathcal L_{\mathrm{prior}}$ widens the margin toward the valid
order on falling-object scenes (main-paper \cref{sec:narrow_default}; \cref{sec:supp_prior}).

\paragraph{A generative renderer for what is unseen.}
Content that the blur does not determine has to be generated, so the renderer is a video
diffusion model conditioned on the blur and the features (main-paper
\cref{sec:feature_renderer}).

\section{Architecture in detail}
\label{sec:supp_arch}

The encoder implements $q_\phi$ and the renderer $p_\omega$ of main-paper \cref{eq:feature_mediated_model}
(main-paper \cref{sec:cross_observation_adaptation,sec:sparse_cues,sec:feature_renderer}). This section
gives the architecture of each, the training settings of both and the cost per video. The analyses of what each part
does are in \cref{sec:supp_encoder,sec:supp_renderer,sec:supp_cues}.

\subsection{Encoder}
\label{sec:supp_enc_arch}
Main-paper \cref{fig:method_training}(b, d) shows the encoder.
The encoder has the architecture of V-JEPA~2.1 ViT-L (the large vision transformer) at 384\,px (24 blocks, 1024-dimensional final tokens). A frozen
copy is the teacher that encodes the sharp clip. The student starts from the same weights and is adapted with LoRA (low-rank adaptation) on
the attention and MLP (multilayer perceptron) projections (settings in \cref{tab:supp_training}). At test it receives $N=8$ identical copies of the blur;
it is trained to match the teacher's final-layer tokens. As a regularizer, it also matches the layer-normalized features
at two intermediate blocks (\cref{sec:supp_layers}).

\par\smallskip\noindent\colorbox{gray!12}{\parbox{\dimexpr\linewidth-2\fboxsep\relax}{\textbf{Key design choice.}
Arrow cues act on the encoder only: a cue changes the predicted features but has no direct path to the renderer.}}
\par\smallskip

\textbf{Arrow path.} Each arrow is rasterized as three channels: the horizontal and vertical unit direction, weighted by local
coverage, and the coverage itself. A small convolutional network (CNN; five $3\times3$ convolutions, total stride 16) maps this image to one
feature per token cell. Two MLP heads, one per injection point (after the patch embedding and after block 7, counted from
0), turn each cell's feature $f$ into an additive bias for every token, from $[f,\ \tau_k f]$, where $\tau_k$ runs linearly
from $-1$ to $1$ over the clip's tubelets $k$. This signed ramp lets one static arrow push the token sequence forward or
backward in time. The last layer of each head is zero-initialized, clips without an arrow receive no bias, and the arrow
encoder is trained together with the LoRA weights.

\subsection{Renderer}
\label{sec:supp_ren_arch}
Main-paper \cref{fig:method_training}(e) shows the renderer.
We implement $p_\omega$ with Wan~2.2 TI2V-5B~\cite{wan2025wan}, trained with its own flow-matching
objective. Its weights are adapted with LoRA on each block's attention and FFN (feed-forward network)
projections (settings in \cref{tab:supp_training}).

$\mathbf{Y}$ reaches the renderer in two ways. The first is Wan's own image-conditioning path. It fixes
$\mathbf{Y}$'s VAE (variational autoencoder) latent in the first latent slot at timestep 0 (noise-free), while every other
position denoises from one shared sampled timestep. This path is pretrained, not zero-initialized, so it ties
that slot to the blur from the first training step. At initialization, subsequent frames also tend to
reproduce the blur; training teaches them to denoise toward the sharp sequence instead.
The conditioning slot is never denoised, so it is excluded from the loss and cropped before
evaluation. The second path is a zero-initialized embedding that tiles the blur latent across frames.
The renderer generates 17 frames: five before the exposure, the first of which is the blur's image slot, then the nine
exposure frames, then three after. In training, the frames around the exposure are the true neighbouring frames, weighted
0.5 in the loss, and give Wan temporal context; only the nine exposure frames are output.

\par\smallskip\noindent\colorbox{gray!12}{\parbox{\dimexpr\linewidth-2\fboxsep\relax}{\textbf{Key design choice.}
The renderer is trained on teacher features and never on predicted features.
This is by design: predicted features carry the arrows, so a renderer trained on them could learn the cue itself.}}
\par\smallskip

The features enter through cross-attention. A connector projects the feature field into a separate, narrow
key/value space, injected into every one of Wan's 30 DiT (diffusion transformer) blocks via zero-initialized cross-attention. Because this is
cross-attention, not concatenation, the key/value set need not match Wan's token count. The key/value set is the full grid
of final-layer V-JEPA tokens, projected pointwise (not pooled or resampled). A learned distance bias favors nearby
keys: each attention logit is reduced by $\alpha_{\mathrm{block,head}}\ge0$ (one weight per block and head, via a
softplus) times the squared spatial distance between the query and key positions.

\subsection{Training details}
\label{sec:supp_training}
\Cref{tab:supp_training} lists the training settings of the encoder and the renderer of the
main comparison. The blur recipe is in \cref{sec:supp_synth_blur}.

\textbf{Order-invariant loss.} We use $\mathcal L_{\mathrm{inv}}$ (main-paper \cref{eq:inverse-loss}) on the real
videos, whose time order is not labelled by physics: on their steps without an arrow, the prediction is matched to whichever of
the captured order and its reversal is closer. Steps with an arrow match the arrow's order, and the physics-labelled
falling-object scenes match their valid order directly, with $\mathcal L_{\mathrm{prior}}$ (\cref{tab:supp_training}).
$\mathcal L_{\mathrm{inv}}$ is a mild regularizer: with plain captured-order matching in its place, the order-invariant feature error rises by 1.4\% (78 validation windows), less
than the 2.2\% spread between two training seeds.

\begin{table*}[!tp]
\centering
\caption{Training settings of the main comparison's encoder and renderer. Steps are optimizer steps; data sources are
sampling probabilities, not clip counts.}
\label{tab:supp_training}
\footnotesize
\begin{tabular}{@{}p{0.13\textwidth}p{0.41\textwidth}p{0.41\textwidth}@{}}
\toprule
& Encoder $q_\phi$ & Renderer $p_\omega$ \\
\midrule
Backbone & V-JEPA~2.1 ViT-L at 384 px (frozen copy as teacher) & Wan~2.2 TI2V-5B \\
Trained weights & LoRA, rank 16, $\alpha{=}32$, on the attention (qkv, output) and MLP projections; the arrow encoder
 & LoRA, rank 64, $\alpha{=}128$, on the attention (q, k, v, output) and FFN projections; the feature cross-attention (width 64, 4 heads, one per block); patch embedding, output projection, condition embedder and one norm layer per block in full, at $0.1\times$ the learning rate \\
Optimizer & AdamW, $\beta{=}(0.9,0.999)$, weight decay 0.01, gradient clipping 1.0, 500 warm-up steps then constant & as the encoder \\
Learning rate & $10^{-4}$ for the first 40k steps, then $5\times10^{-5}$ & $10^{-4}$ \\
Batch & effective 2 (2, or 1 with 2 accumulation steps), one GPU, bfloat16 & 2, one GPU, bfloat16 \\
Steps & about 95k & 115k \\
Objective & MSE (mean squared error) to the teacher's final-layer tokens (real videos without an arrow: $\mathcal L_{\mathrm{inv}}$; with an arrow: its order; falling-object scenes: their valid order) plus $0.5\times$ the mean MSE at blocks 8 and 12; pair margin (\cref{sec:supp_cues}): weight 0.5, margin 0.5, $\epsilon{=}0.05$; no-cue ranking loss on 25\% of steps: weight 0.5, $\tau{=}0.5$, $\epsilon{=}0.05$
 & flow-matching velocity MSE, weight 0.5 on latents that hold only frames outside the exposure; timesteps logit-normal $(0,1)$ with shift 5 \\
Feature input & arrows on about 72\% of the steps (85\% of single real clips; \cref{tab:supp_cue_mix}) & teacher tokens with Gaussian noise ($0.3\times$ their root mean square) and token dropout 0.2 \\
Feature tokens & at most 576, 800 or 1024 per sample & 576 / 800 / 1024 with probability 0.4 / 0.3 / 0.3 \\
Training data & steps by cue source: see \cref{tab:supp_cue_mix}; real clips drawn from GoPro 0.225, Adobe240 0.225, X4K 0.1875, clips of Tedla et al.~\cite{tedla2025generating} 0.25, FMO 0.05625, TbD 0.05625
 & synthetic multi-object scenes 0.30, Adobe240 0.175, clips of Tedla et al. 0.175, GoPro 0.14, X4K 0.14, FMO 0.035, TbD-3D 0.035 \\
Augmentation & shot noise (probability 0.5) and JPEG compression (0.3) on the blur & the same, with crops of 384 to 704 px at aspect ratios 0.5 to 2, horizontal flip and time reversal (0.5 each) \\
\bottomrule
\end{tabular}
\end{table*}

\paragraph{Training compute.} Every training run uses one A100 80GB in bfloat16. A renderer run of 30{,}000 steps takes
about 18\,h (2.2\,s per step), measured on Wan2.2 (finetuned), the feature-free renderer of \cref{tab:supp_nofeat}.

\subsection{Cost per video}
\paper needs about half the time and a third of the compute of Tedla et al.\ per video.
\Cref{tab:supp_cost} gives the inference cost of every method. For \paper, 87\% of the time is the 50 sampling steps of
the renderer.

\begin{table*}[!tp]
\centering
\caption{\textbf{\paper takes about half the time per video of Tedla et al.\ (34 vs.\ 69\,s), and arrows add no cost.} Inference cost on one A100 80GB: seconds per video (median of videos 2--5, excluding model loading and warm-up), output frames (resolution in parentheses; \paper generates 17 and keeps the 9 of the exposure), peak allocated GPU memory, parameters loaded / trained (millions) and FLOPs (floating-point operations) per video. $\geq$: lower bound, deformable convolutions not counted.}
\label{tab:supp_cost}
\footnotesize
\setlength{\tabcolsep}{3.5pt}
\begin{adjustbox}{max width=\textwidth}
\begin{tabular}{lccccc}
\toprule
Method & s / video $\downarrow$ & Output frames & Peak memory (GiB) $\downarrow$ & Parameters loaded / trained (M) & TFLOPs / video $\downarrow$ \\
\midrule
Tedla et al., GoPro model & 69.4 & 7 & 17.1 & 6,672 / 1,694 & 9,058 \\
Tedla et al., in-the-wild model & 69.3 & 16 (720 $\times$ 1280) & 19.1 & 6,672 / 1,694 & 8,825 \\
Motion-ETR~\cite{zhang2022exposure} & 0.56 & 9 & 2.5 & 7.7 / 7.7 & $\geq$\,1.4 \\
Jin et al.~\cite{jin2018learning} & 0.38 & 7 & 1.3 & 18.2 / 18.2 & 6.2 \\
Animation from Blur~\cite{zhong2022animation} & 0.36 & 7 & 4.3 & 191.8 / 191.8 & 19.6 \\
\oursrow\textbf{\paper (blind)} & 33.5 & 9 (736 $\times$ 1280) & 34.7 & 11,980 / 290 & 2,879 \\
\oursrow\paper + arrows & 33.6 & 9 & 34.7 & 11,980 / 290 & 2,879 \\
\bottomrule
\end{tabular}
\end{adjustbox}
\end{table*}

\section{Encoder analysis}
\label{sec:supp_encoder}

This section analyses the adapted encoder: where in V-JEPA the supervision sits, what its features encode, and how it
chooses the time order without arrows (the ranking loss and the no-cue default).

\subsection{Choice of supervised layers}
\label{sec:supp_layers}

\begin{figure*}[!tp]
\centering
\includegraphics[width=\textwidth]{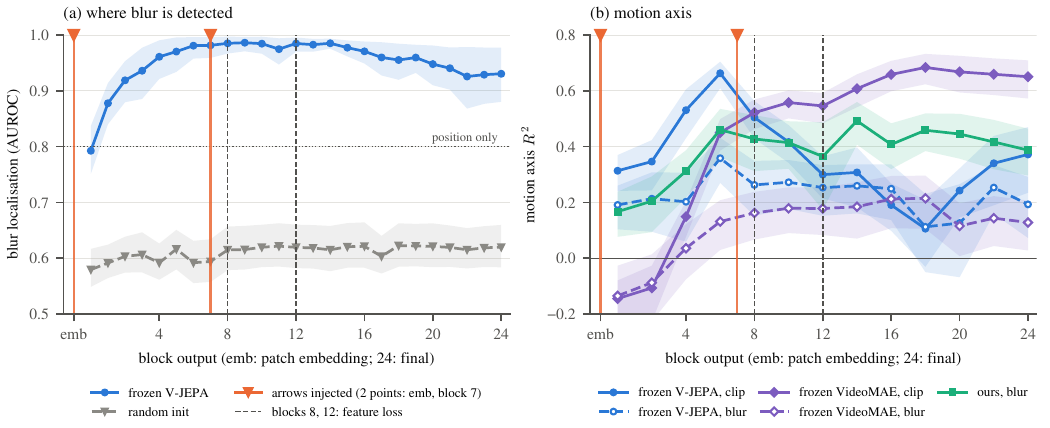}
\caption{\textbf{Blur is linearly readable early in frozen V-JEPA, and from block 4 on the adapted encoder's curve lies above both frozen encoders on the blur.} Layer analysis. (a) Blur localization per block (position-matched AUROC). (b) Motion-axis $R^2$ of linear probes
per block (176 windows): frozen encoders on the true clip (solid) and on the blur (dashed); ours: the adapted encoder on the
blur. Orange: arrow injection points (patch embedding, block 7); grey dashed: supervised blocks 8 and 12. Plain
cross-validation per block; main-paper \cref{fig:motiv_probes} shows the blur values with the block chosen inside the
training folds. Shaded: 95\% interval, bootstrap over the 11 test recordings (a) and over the source clips (b).}
\label{fig:supp_layers}
\end{figure*}

The supervised blocks and the arrow injection depth come from prior probing of V-JEPA~2~\cite{joseph2026interpreting},
where motion direction becomes linearly decodable at about one third of the depth. Besides the
final tokens, the adaptation matches layer-normalized features at blocks 8 and 12 (counted from 0, of 24). Arrows enter
after the patch embedding and after block 7 (main paper, \cref{sec:cross_observation_adaptation,sec:sparse_cues}).
\Cref{fig:supp_layers} gives two analyses of our own.

\begin{figure*}[!tp]
\centering
\includegraphics[width=\textwidth]{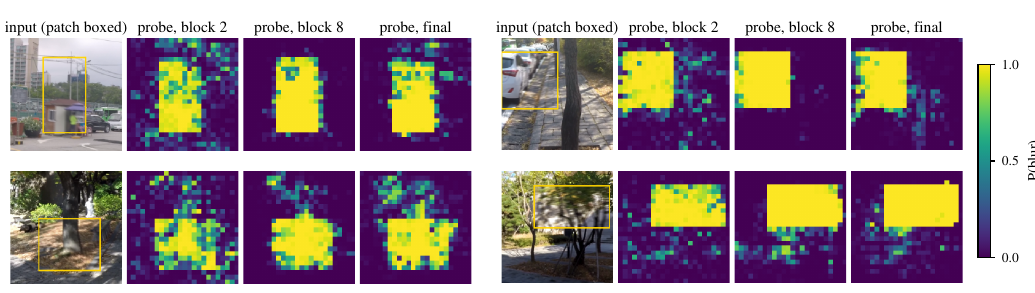}
\caption{\textbf{The probe localizes a spliced blurred patch, noisily at block 2 and cleanly from block 8 on.} Blur probe setup of \cref{fig:supp_layers}(a), frozen V-JEPA. Each group: a held-out GoPro test crop with a blurred
patch spliced in (boxed), then the linear probe's per-token probability of blur at blocks 2, 8 and the final block. The four
crops span the range of blur strength.}
\label{fig:supp_blur_patch}
\end{figure*}

\textbf{Where blur is detected (a).} In frozen V-JEPA, blur is decodable early and best in the middle blocks. A blurred
patch spliced into sharp GoPro crops gives exact per-token labels. A linear probe per block then separates blurred from
sharp tokens. We score it by AUROC (area under the receiver operating characteristic curve), the chance that a blurred token scores above a sharp one at the same position (0.5:
chance, 1: perfect), on tokens at least three cells from the patch border and a scene-disjoint split. Blur becomes
decodable from block 2, where the probe clearly beats a probe on cell position alone; it peaks over blocks 6--14 and
declines toward the output. In the probe's per-token map, the patch is noisy at block 2 and clean from block 8 on
(\cref{fig:supp_blur_patch}). As controls, panel (a) also shows the random network and a probe on cell position alone. Activation patching in frozen
V-JEPA places most of the blur evidence in the patch embedding and the first blocks (0--7). This analysis measures blur, not motion.

\textbf{Where motion is decodable (b).} In frozen V-JEPA, the motion axis is more readable in blocks 4--10 than at
the output. These are the per-block probes behind main-paper \cref{fig:motiv_probes}, with every encoder on one
$14\times14$ grid. On the true clip, frozen V-JEPA reads the motion axis best at block 6 and worst at
block 18. VideoMAE peaks late, at block 18, about as high as frozen V-JEPA's peak (\cref{fig:supp_layers}b). From a blur, the
input the method sees, VideoMAE reads less of the axis than frozen V-JEPA: with the block chosen inside the training
folds, axis $R^2$ 0.208 against 0.359, a significant difference. The adapted encoder's curve lies above frozen V-JEPA's on the
blur at every block from 4 on; the per-block intervals overlap at most blocks, and the difference is significant with the
block chosen inside the training folds ($+0.093$ [$+0.013$, $+0.164$]). Neither encoder recovers the signed direction from a blur: a linear probe of the signed direction does
worse than predicting the mean at every block.

Block 8 lies in this band, and the second arrow input enters just before it; block 12 follows the prior probing above. We did not vary these blocks. The renderer of the main comparison reads the final V-JEPA layer only.

\subsection{What the inferred features encode}
\Cref{tab:supp_probe} gives linear probes of per-region motion on V-JEPA's own $18\times32$ grid; main-paper \cref{fig:motiv_probes} uses
one $14\times14$ grid for every encoder. Besides the blur rows, the table has a sharp-frame control and the
true clip. The true clip's final-layer value is a reference, not an upper bound: the adapted encoder's blur features
read the axis better than frozen V-JEPA's final layer reads it from the true clip. Two facts are consistent with this.
The axis of a streak is visible in a single image, so one blurred photograph already carries it, and frozen V-JEPA's final
layer reads the axis worse than blocks 4--10 do (\cref{fig:supp_layers}b).

\begin{table}[!tp]
\centering
\caption{\textbf{The adapted encoder reads more motion from the blur than frozen V-JEPA, and its motion axis does not come from image content alone.} Linear probes of per-region motion from final-layer, tubelet-averaged features (176 windows: 56 real, 120 synthetic; ridge regression, 5 scene-cluster folds; $R^2<0$: worse than predicting the mean). MAG: log motion magnitude; AXIS: motion axis on the 713 moving cells. The 120 synthetic windows carry most of the AXIS gain; on the 56 real
windows the adapted encoder's gain over frozen V-JEPA is not significant.}
\label{tab:supp_probe}
\footnotesize
\setlength{\tabcolsep}{3pt}
\begin{adjustbox}{max width=\columnwidth}
\begin{tabular}{L{0.42\columnwidth}cc}
\toprule
Features & MAG $R^2$ $\uparrow$ & AXIS $R^2$ $\uparrow$ \\
\midrule
Adapted encoder, sharp center frame (content only) & 0.621 [0.555, 0.676] & $-$0.357 [$-$0.515, $-$0.242] \\
Frozen V-JEPA on the sharp center frame & 0.629 [0.564, 0.682] & $-$0.338 [$-$0.475, $-$0.239] \\
Frozen V-JEPA on the same blur & 0.542 [0.452, 0.609] & 0.265 [0.159, 0.358] \\
Frozen V-JEPA, true clip (reference) & 0.797 [0.748, 0.836] & 0.400 [0.305, 0.475] \\
\oursrow\textbf{Adapted encoder, blur} & \textbf{0.747 [0.691, 0.789]} & \textbf{0.562 [0.488, 0.622]} \\
\bottomrule
\end{tabular}
\end{adjustbox}
\end{table}

\paragraph{PCA (principal component analysis) view of the features (main-paper \cref{fig:pca}).}\label{sec:supp_pca}
The figure shows each encoder's final-layer tokens on V-JEPA's $18\times32$ grid for the four tubelets. The teacher reads
the true clip; ours and frozen V-JEPA read the blur repeated eight times. From each encoder's tokens we subtract its
response to a featureless input and each cell's mean over the tubelets. The result is projected onto a 3-component PCA
fitted on ours, smoothed over $2\times2$ cells and overlaid on the frame. The clip shown is the real held-out window where
the energy of our token change over the tubelets most exceeds frozen V-JEPA's; ours exceeds it on 27\% of the 56 real
windows. The figure shows a selected case.

\subsection{Ranking loss}
\label{sec:supp_default}
\label{sec:supp_prior}
This is the full construction of $\mathcal L_{\mathrm{prior}}$ (main paper,
\cref{sec:narrow_default}). Each synthetic falling-object group pairs a valid
settling sequence $\mathbf{X}^+$ with blur-matched invalid orders $\mathbf{X}_j^-$
($j$ indexes the negatives tested for that example). The primary tested negative is an
energy-gaining reversal of an otherwise passive object. Reversals consistent with an
external force, e.g.\ a throw, are not labeled invalid, so active dynamics are excluded.
Let $\mathbf{T}^+{:=}\mathbf{T}(\mathbf{X}^+)$,
$\mathbf{T}_j^-{:=}\mathbf{T}(\mathbf{X}_j^-)$ be their frozen-teacher features. Let
$\operatorname{MSE}_c(\mathbf U,\mathbf V)$ denote the mean squared
distance between final-layer tokens at spatial cell $c$ (averaged over
temporal tubelets and channels). Define
$g_{j,c}=\operatorname{MSE}_c(\mathbf{T}^+,\mathbf{T}_j^-)$
and the normalized preference of the predicted features $\widehat{\mathbf{Z}}'$
\begin{equation}
    r_{j,c}=
    \frac{\operatorname{MSE}_c(\widehat{\mathbf{Z}}',\mathbf{T}_j^-)-\operatorname{MSE}_c(\widehat{\mathbf{Z}}',\mathbf{T}^+)}
         {g_{j,c}},
\end{equation}
positive when $\widehat{\mathbf{Z}}'$ is nearer the valid order at cell $c$. The loss uses only cells that are separated enough
relative to this example's own most-separated cell,
$\mathcal V_j=\{c:g_{j,c}>\epsilon\max_{c'}g_{j,c'}\}$ for a fixed fraction
$\epsilon$. A negative $j$ with $\mathcal V_j=\varnothing$ is dropped from
both the sum and $J$ below, not counted as zero. Over these cells we apply a weighted one-sided ranking loss:
\begin{equation}
    \mathcal L_{\mathrm{prior}}=
    \frac{1}{J}\sum_j\sum_{c\in\mathcal V_j}
    \underbrace{\frac{g_{j,c}}{\sum_{k\in\mathcal V_j}g_{j,k}}}_{\text{cell weight}}
    \underbrace{[\tau-r_{j,c}]_+}_{\text{margin hinge}},
    \label{eq:supp_prior_loss}
\end{equation}
where $\tau$ is a margin hyperparameter and $J$ counts only the negatives
with $\mathcal V_j\neq\varnothing$. Each included negative is therefore weighted
equally, regardless of how many that example offered. If none has
eligible cells ($J=0$), $\mathcal L_{\mathrm{prior}}$ is zero for that
example. The hinge zeroes once $r_{j,c}\ge\tau$. The loss thus ranks the valid order
ahead of each negative by a margin rather than regressing to
$\mathbf{T}^+$.

\subsection{Learned no-cue default}
Many training steps have no arrow (\cref{tab:supp_cue_mix}): real clips drawn without one, uncued objects in the
synthetic scenes, and the falling-object (drop) scenes. On the drop scenes the valid order is known, and the ranking loss
shapes the preference between a clip and its reversal; on real clips without an arrow, $\mathcal L_{\mathrm{inv}}$ accepts
either order (\cref{sec:supp_training}). We test the default on the same kind of scene: a dropped object settling, against
its energy-gaining reversal, which gives the same blur. \Cref{tab:supp_default_tokens} scores it in feature space and in the generated video
on two synthetic sets: the drop scenes of the validation set and fresh drop scenes whose objects and backgrounds are absent
from training (the video rates of main-paper \cref{fig:motiv_default}). A static prediction cannot pass the feature
test, ties count as failures in the video test, and the published methods are out of domain here.

In feature space, the default is learned, not inherited from V-JEPA: frozen V-JEPA's features prefer the settling order in
fewer than half of the pairs, and our encoder trained without the drop scenes is at chance; adding them puts the preference
there. In the generated video, the final model's rate on fresh scenes is significantly below that of the MSE-only ablation and
not significantly above that of the encoder trained without drop scenes, although its margin over the latter is significant.
Compared with MSE alone on the same clips and for the same steps, the ranking loss raises the feature rate
on validation scenes only; on fresh scenes it gives a wider rendered margin at a similar rate (ablation rows). Arrow cues
are exempt from the ranking loss, so an arrow overrides the default (lower block): a valid arrow is followed in almost every trial, and so is an impossible one that
commands the reversal. ``All cues'' includes the commands for which the arrow simulator drew no arrow; these fall back to
the default.

A renderer trained without features has no order preference of its own: its rate is at chance (Wan2.2 (finetuned)).
The video rates are nevertheless at least as high as the feature rates in every encoder row, also for the encoder trained
without the drop scenes. The feature rate only asks which clip's final-layer features are nearer, so it does not
count all of the order information the renderer reads from the features. We do not separate this from an order preference
that the renderer may acquire when it is trained with features.

\begin{table*}[!tp]
\centering
\caption{\textbf{Without arrows, \paper prefers the settling order over its energy-gaining reversal more often than every compared method; an arrow still overrides this default.} Learned no-cue default on synthetic drop scenes. Each pair is a dropped object settling (the valid clip) and its
energy-gaining time reversal; both give the same blur. Validation: held-out scenes with backgrounds unseen in training
(some objects seen); fresh: objects and backgrounds never seen in training. Encoder features: share of pairs whose predicted features are nearer
(final-layer MSE) the settling clip's than the reversal's (no renderer). Video: share of pairs whose generated blind video is
nearer the settling clip; margin: masked PSNR to the settling clip minus to the reversal. The L2-only and ranking-loss
ablations are step-matched to each other.}
\label{tab:supp_default_tokens}\label{tab:supp_default_video}\label{tab:supp_default_arrows}
\scriptsize
\setlength{\tabcolsep}{3pt}
\begin{tabular}{L{0.34\textwidth}cccc}
\toprule
 & \multicolumn{2}{c}{Validation scenes (41 pairs)} & \multicolumn{2}{c}{Fresh scenes (58 pairs)} \\
\cmidrule(lr){2-3}\cmidrule(lr){4-5}
Video only (no encoder features) & Video $\uparrow$ & Margin (dB) $\uparrow$ & Video $\uparrow$ & Margin (dB) $\uparrow$ \\
\midrule
Blur copy (a static video ties every pair) & 0.000 & 0.00 & 0.000 & 0.00 \\
Jin et al. & 0.341 [0.146, 0.550] & +0.15 [$-$0.34, +0.71] & 0.138 [0.034, 0.259] & $-$0.27 [$-$0.49, $-$0.07] \\
Animation from Blur & 0.488 [0.325, 0.659] & $-$0.03 [$-$0.21, +0.15] & 0.345 [0.207, 0.483] & $-$0.24 [$-$0.42, $-$0.08] \\
Motion-ETR & 0.415 [0.214, 0.619] & $-$0.25 [$-$1.07, +0.65] & 0.431 [0.259, 0.603] & $-$0.14 [$-$0.59, +0.33] \\
Wan2.2 (finetuned), renderer without features & 0.488 [0.333, 0.643] & +0.09 [$-$0.50, +0.76] & 0.466 [0.345, 0.586] & $-$0.29 [$-$0.65, +0.03] \\
Tedla et al. & 0.488 [0.293, 0.690] & +0.26 [$-$0.41, +0.99] & 0.603 [0.431, 0.759] & +0.37 [$-$0.30, +1.02] \\
\bottomrule
\end{tabular}

\vspace{4pt}
\begin{adjustbox}{max width=\textwidth}
\begin{tabular}{L{0.25\textwidth}cccccc}
\toprule
 & \multicolumn{3}{c}{Validation scenes (41 pairs)} & \multicolumn{3}{c}{Fresh scenes (58 pairs)} \\
\cmidrule(lr){2-4}\cmidrule(lr){5-7}
Encoder & Encoder features $\uparrow$ & Video $\uparrow$ & Margin (dB) $\uparrow$ & Encoder features $\uparrow$ & Video $\uparrow$ & Margin (dB) $\uparrow$ \\
\midrule
Ablation: without drop scenes & 0.512 [0.317, 0.707] & 0.756 [0.575, 0.923] & +1.32 [+0.26, +2.50] & 0.500 [0.345, 0.672] & 0.776 [0.638, 0.897] & +1.16 [+0.61, +1.74] \\
Ablation: + drop scenes, MSE only & 0.512 [0.333, 0.690] & 0.805 [0.643, 0.951] & +2.05 [+0.92, +3.18] & 0.690 [0.552, 0.828] & 0.914 [0.828, 0.983] & +1.99 [+1.38, +2.62] \\
Ablation: + ranking loss & 0.756 [0.571, 0.921] & 0.902 [0.786, 1.000] & +3.63 [+2.47, +4.87] & 0.724 [0.569, 0.862] & 0.897 [0.793, 0.983] & +2.69 [+1.94, +3.44] \\
\oursrow\textbf{\paper} & 0.927 [0.810, 1.000] & 0.927 [0.850, 1.000] & +3.56 [+2.28, +4.97] & 0.690 [0.517, 0.828] & 0.828 [0.690, 0.948] & +2.79 [+2.02, +3.56] \\
\bottomrule
\end{tabular}
\end{adjustbox}

\vspace{4pt}
\begin{tabular}{L{0.42\textwidth}cc}
\toprule
\textit{\paper with arrows, validation scenes; (n): pairs or scenes} & Encoder features & Video \\
\midrule
Valid arrow: follows it $\uparrow$ & 0.961 [0.917, 0.994] (180) & 0.952 [0.857, 1.000] (21) \\
Impossible arrow: keeps the settling order anyway $\downarrow$, all cues & 0.146 [0.024, 0.293] (41) & 0.238 [0.095, 0.429] (21) \\
Impossible arrow: keeps the settling order anyway $\downarrow$, cues with an arrow & 0.029 [0.000, 0.086] (35) & 0.111 [0.000, 0.278] (18) \\
\bottomrule
\end{tabular}
\end{table*}

\section{Renderer analysis}
\label{sec:supp_renderer}

This section gives the evidence on the renderer: that it follows the features, what the image slot does, and what the
features add over a renderer without them. The range test (\cref{sec:supp_range}) shows that at large blur the remaining
error comes from inferring the features, not from rendering them.

\subsection{The renderer follows the features}
\noindent\colorbox{gray!12}{\parbox{\dimexpr\linewidth-2\fboxsep\relax}{\textbf{Key finding.}
The renderer carries out the motion its features command, whether or not that motion is physically possible.
In \cref{tab:supp_renderer_follows}, we command reversals and unforced turns by giving the renderer the teacher features of the commanded clip. The renderer follows them in
every decided case, including all energy-gaining reversals, and does not repair them.}}
\par\smallskip
\noindent A sudden jump in position (teleport), in contrast, is mostly not followed (\cref{tab:supp_renderer_follows}). This is the desired
behavior: a jump within the exposure is not continuous motion, and the output stays continuous instead of reproducing it. The
jump falls inside one V-JEPA tubelet, so the features may not encode it; the test cannot tell whether the continuity comes
from the features or from the renderer.

\begin{table}[!htbp]
\centering
\caption{The renderer of \paper given the teacher features (frozen V-JEPA) of each candidate clip (synthetic validation set, all scene kinds, 154 blurs). Decided: the nearest clip (masked PSNR) beats the next by at least 0.5\,dB. Follow: share of decided cases nearest the commanded clip; repaired: nearest a valid order instead (N/A: the command is already valid).}
\label{tab:supp_renderer_follows}
\footnotesize
\setlength{\tabcolsep}{3.5pt}
\begin{adjustbox}{max width=\columnwidth}
\begin{tabular}{L{0.40\columnwidth}ccc}
\toprule
Commanded motion & Follow $\uparrow$ & Decided & Repaired \\
\midrule
The true clip & 1.000 & 154 / 154 & N/A \\
Admissible reversal & 1.000 & 87 / 87 & N/A \\
Infeasible (energy-gaining) reversal & 1.000 & 48 / 48 & 0.000 \\
Unforced turn & 1.000 & 108 / 110 & 0.000 \\
Teleport & 0.267 [0.186, 0.358] & 101 / 151 & 0.079 \\
\bottomrule
\end{tabular}
\end{adjustbox}
\end{table}

\subsection{The image slot}
\label{sec:supp_slot}
The renderer reads appearance from the blur itself, and a sharp image swapped into its place at inference does not help. The blur fills Wan's
image-conditioning slot, and the renderer is trained with it there. \Cref{tab:supp_clean_frame} replaces it at inference
on the Adobe240 windows of main-paper \cref{tab:main}. With the frame deblurred by NAFNet~\cite{chen2022nafnet} used by the Deblur + I2V (image-to-video) baseline of the main paper,
PSNR falls and EPE rises, and the true middle frame, an oracle, gives a similar loss. Only LPIPS improves, in both
cases. Each of these differences from the blur is significant (paired). A sharp first image pulls the video towards that one instant, so the motion over the exposure is
lost.

\begin{table}[!htbp]
\centering
\caption{The image slot (Adobe240, blind): only the image in the renderer's slot changes. Blur: \paper (blind) of main-paper
\cref{tab:main}.}
\label{tab:supp_clean_frame}
\footnotesize
\begin{tabular}{lccc}
\toprule
Image in the slot & PSNR $\uparrow$ & LPIPS $\downarrow$ & EPE $\downarrow$ \\
\midrule
True middle frame (oracle) & 25.16 & 0.100 & 2.08 \\
NAFNet-deblurred frame & 25.17 & 0.104 & 1.98 \\
\oursrow\textbf{Blur (\paper)} & 26.80 & 0.115 & 0.69 \\
\bottomrule
\end{tabular}
\end{table}

\subsection{What the features add}
\label{sec:supp_features_add}
\Cref{tab:supp_nofeat} compares \paper with Wan2.2 (finetuned), a renderer trained without features:
our recipe, data and LoRA from the Wan~2.2 base, with its feature input zeroed in training and at test and trained until its
loss stops improving, so it takes no cues. Without features, PSNR is lower and LPIPS higher on all five sets, all significant; the motion error (EPE) is level with that of
\paper blind, with no significant difference on any set.

\begin{table}[!htbp]
\centering
\caption{Wan2.2 (finetuned), the renderer trained without features, minus \paper (blind), paired. Its own scores are in
main-paper \cref{tab:main} and \cref{tab:supp_x4k}.}
\label{tab:supp_nofeat}\label{sec:supp_no_feature}
\footnotesize
\setlength{\tabcolsep}{3pt}
\begin{adjustbox}{max width=\columnwidth}
\begin{tabular}{lccc}
\toprule
Set (windows) & PSNR & LPIPS & EPE \\
\midrule
GoPro (22) & $-0.74$ [$-1.23$, $-0.29$] & $+0.017$ [$+0.008$, $+0.030$] & $+0.32$ [$-0.00$, $+0.84$] \\
Adobe240 (154) & $-0.32$ [$-0.57$, $-0.09$] & $+0.010$ [$+0.007$, $+0.013$] & $-0.00$ [$-0.06$, $+0.07$] \\
I2-2000FPS (31) & $-0.74$ [$-1.16$, $-0.25$] & $+0.019$ [$+0.013$, $+0.025$] & $+0.16$ [$-0.11$, $+0.43$] \\
Real camera blur, RBI (115) & $-0.45$ [$-0.67$, $-0.28$] & $+0.009$ [$+0.005$, $+0.014$] & $-0.00$ [$-0.09$, $+0.06$] \\
X4K (15) & $-0.63$ [$-0.99$, $-0.35$] & $+0.025$ [$+0.012$, $+0.038$] & $+0.02$ [$-0.17$, $+0.28$] \\
\bottomrule
\end{tabular}
\end{adjustbox}
\end{table}

\paragraph{Rows that reuse the trained renderer.}\label{sec:supp_not_retrained} The rows ``\paper, frozen encoder'' and
``\paper, features zeroed'' of main-paper \cref{tab:main} are inference-time interventions: the renderer of \paper is
not retrained for either input. They show that this renderer relies on the adapted encoder's features. They do not show
what a renderer trained on frozen-encoder features would reach; the trained comparison is the one above.

\section{Arrow cues and control}
\label{sec:supp_cues}

Arrows select one temporal orientation while the encoder infers the features (main-paper \cref{sec:sparse_cues}). This section
describes how the training arrows are simulated, the synthetic control test set we render in Blender with its six
scene-motion classes, and the control results with simulated arrows and with arrows drawn by people.

\subsection{Training arrows}
\label{sec:supp_cues_real}
\label{sec:supp_cues_synth}
\label{sec:supp_cues_mix}

Training has no user, so every training arrow is simulated from known motion. For a real clip, RAFT~\cite{teed2020raft}
point tracks of the sharp frames are clustered into coherent motion groups, and one to three groups each get an arrow along
their mean path. The arrow is then randomized in shape, length, rotation, placement and thickness
(\cref{tab:supp_arrow_random}). It keeps the direction of the motion but not its speed, length or exact path; \cref{fig:supp_arrow_perturb}
(top) shows training arrows from every source. The encoder is never given optical flow, tracks or speed, so how far
and how fast each part moves must come from the blur. \Cref{sec:supp_enc_arch} gives how arrows enter the encoder.

Two synthetic generators add scenes with exact motion: textured objects moving over paintings, and Flickr2K
photographs~\cite{lim2017enhanced} with a segmented foreground under camera and object motion. Their scene kinds teach when to follow an arrow and when not to (\cref{tab:supp_scene_kinds}). An arrow on a static object or in a static scene must not create motion. An
arrow at an angle to the motion is trained to set its sign without pulling the video off the blur axis.

Two rules make the cues local and consistent. In a region-reversal pair, one motion group is played backwards inside its
region; the blur is unchanged, and a pair margin loss asks each prediction to be closer to its own target. Flips and time
reversals are applied jointly to the clip, the blur and the arrows.

\begin{table}[!htbp]
  \centering
  \small
  \caption{Randomization of each simulated arrow on real clips and layered natural images. Lengths are fractions of the
image's shorter side. Objects over paintings use the same length, rotation, placement and thickness, but a straight arrow
with a slight bow (probability 0.5) and no wobble.}
  \label{tab:supp_arrow_random}
  \label{tab:supp_track_params}
  \begin{tabular}{@{}p{0.27\linewidth}p{0.65\linewidth}@{}}
    \toprule
    Property & Simulation \\
    \midrule
    Shape & straight chord with probability 0.5 (always if the true path
      is shorter than 4 px); otherwise a quadratic fit of the path in time \\
    Length & uniform in $[0.12, 0.45]$, independent of the true
      displacement \\
    Rotation & uniform in $[-10^\circ, 10^\circ]$ \\
    Hand wobble & sinusoidal offset along the normal, amplitude 3\% of the
      drawn length, random frequency and phase \\
    Placement & at the group's mid-exposure centroid plus a Gaussian shift
      with standard deviation 0.03 \\
    Thickness & uniform in $[0.015, 0.04]$ \\
    \bottomrule
  \end{tabular}
\end{table}

\begin{table}[!htbp]
  \centering
  \small
  \caption{Training cue sources: the probability that a training step draws each source (sums to 1).}
  \label{tab:supp_cue_mix}
  \begin{adjustbox}{max width=\columnwidth}
  \begin{tabular}{@{}L{0.50\linewidth}L{0.25\linewidth}c@{}}
    \toprule
    Source & Arrows & Prob. \\
    \midrule
    Falling objects (main-paper \cref{sec:narrow_default}) & none & 0.25 \\
    Real clip, single & RAFT; no arrow in 15\% & 0.1875 \\
    Real region-reversal pair & RAFT, up to 3 & 0.1875 \\
    Objects over paintings & exact motion & 0.1875 \\
    Layered natural images & exact motion & 0.1875 \\
    \bottomrule
  \end{tabular}
  \end{adjustbox}
\end{table}

\begin{table}[!htbp]
  \centering
  \small
  \caption{Scene kinds of the two synthetic generators of \cref{tab:supp_cue_mix}: the share of each kind within its
    generator (each block sums to 1).}
  \label{tab:supp_scene_kinds}
  \begin{adjustbox}{max width=\columnwidth}
  \begin{tabular}{@{}L{0.78\linewidth}c@{}}
    \toprule
    Scene kind & Share \\
    \midrule
    \multicolumn{2}{@{}l}{\textit{Objects over paintings}} \\
    A global cue on co-moving objects & 0.14 \\
    Every object cued, with mixed directions across the two variants & 0.22 \\
    A single arrow on one object or on empty background, which nearby objects tend to follow & 0.20 \\
    A random arrow on a static object, which must stay static & 0.14 \\
    A fully static scene with random arrows, which must not invent motion & 0.10 \\
    An arrow at a random angle to the true motion & 0.20 \\
    \midrule
    \multicolumn{2}{@{}l}{\textit{Layered natural images}} \\
    Camera motion only & 0.25 \\
    Two independent objects under a static camera, some uncued, sometimes with an extra arrow on the static background & 0.35 \\
    Camera plus one object, both cued & 0.25 \\
    A static scene with random arrows & 0.15 \\
    \bottomrule
  \end{tabular}
  \end{adjustbox}
\end{table}

\subsection{Synthetic control test set}
\label{sec:supp_synth_control}
Real footage does not give the true motion and mask of each object, and its scenes cannot be chosen by type. We
therefore render a categorized test set in Blender, used only for testing, in which every object's motion and mask
are exact.

\paragraph{Rendering.} Each window is a path-traced render (Cycles) of 19 frames at 240\,fps and $640\times640$ pixels.
A real 360$^\circ$ high-dynamic-range (HDR) panorama from Poly Haven provides the background and the lighting, and an invisible
shadow-catcher ground lets objects cast shadows onto the photographed ground. The moving objects are photogrammetry
scans from Google Scanned Objects and Poly Haven props at real size; the first object covers 2--30\% of the frame, the
others less. Their
motion comes from Bullet rigid-body physics: ballistic flight, bounces, slides and rolls. Every scored object stays in frame
and moves at most 9.5\,px per frame. In these panorama scenes a moving camera only rotates. Full-3D scenes, used in the
classes with a moving camera, add a textured ground with props from 1.5 to 60\,m and a translating camera, so near objects show
parallax. Poisson--Gaussian noise in linear light and one H.264 encode are added to every frame, and a per-pixel
object mask is kept for scoring only. The blur is the linear-light mean of an 11-frame exposure, and its nine truth
frames are chosen as in the main comparison (\cref{sec:supp_protocol}).

\paragraph{Scene-motion classes.} The windows fall into six classes, a static or moving camera crossed with zero, one
or two to three moving objects: L0, L1 and L3 have a static camera, and L2, L4 and L5 a moving one, with zero, one and two
to three objects in that order (\cref{tab:supp_control_levels}). Of 116 rendered windows, 113 are scored: in two a moving
object leaves the frame, and one has no background motion to place an arrow on. Each moving object gets one arrow along its true path in the image (in
L2, the arrow goes on the background), and each class is tested under two opposite commands. L0 has nothing to
command and is used only for reconstruction. The objects are rigid: there are no people or animals, and nothing in
the panorama moves.

\begin{table}[!htbp]
\centering
\caption{\textbf{With arrows, every commanded part follows in 0.958 of the mixed-direction windows, against 0.111 without arrows.} Control per scene-motion class of the Blender test set (main-paper \cref{fig:control}(a)): share of windows
in which every commanded part follows its arrow, for the class's two commands (first / second), with and without arrows.
$^\dagger$: in the 18 windows where the camera motion dominates, reversing the object's arrow also reverses the camera
motion, so these windows do not test locality.}
\label{tab:supp_control_levels}
\footnotesize
\setlength{\tabcolsep}{4pt}
\begin{adjustbox}{max width=\columnwidth}
\begin{tabular}{L{0.40\columnwidth}ccO}
\toprule
Class (windows) & Commands & No arrow & \paper $\uparrow$ \\
\midrule
L0 static scene (10) & none & \multicolumn{2}{c}{nothing to command} \\
L1 static camera, 1 object (10) & $+$ / $-$ & 0.600 / 0.400 & 1.000 / 1.000 \\
L2 moving camera, arrow on background (29) & $+$ / $-$ & 0.414 / 0.586 & 0.966 / 1.000 \\
L3 static camera, 2--3 objects (10) & $+-$ / $-+$ & 0.000 / 0.300 & 1.000 / 0.900 \\
L4 moving camera, 1 object (28) & $+$ / $-$ & 0.429 / 0.571 & 1.000 / 1.000$^\dagger$ \\
L5 moving camera, 2--3 objects (26) & $+-$ / $-+$ & 0.077 / 0.115 & 0.962 / 0.962 \\
\midrule
L3 + L5, mixed (36; 72 commands) & $+-$ / $-+$ & 0.111 [0.056, 0.181] & 0.958 [0.889, 1.000] \\
\bottomrule
\end{tabular}
\end{adjustbox}
\end{table}

\subsection{Control results}
\label{sec:supp_control_results}
These cues are simulated from estimated or ground-truth motion and should not be read as a user study.

\paragraph{Synthetic scenes.} \paper follows each object's own command. With two or three objects given opposite arrows
(L3 and L5), every object follows in nearly all commands, against few without arrows (\cref{tab:supp_control_levels}).
In L4 the single object follows its arrow in every window, but where the camera motion dominates (18 windows), reversing
the arrow also reverses the camera motion in all of them, so L4 does not test locality.

\begin{figure*}[!tp]
\centering
\includegraphics[width=\textwidth]{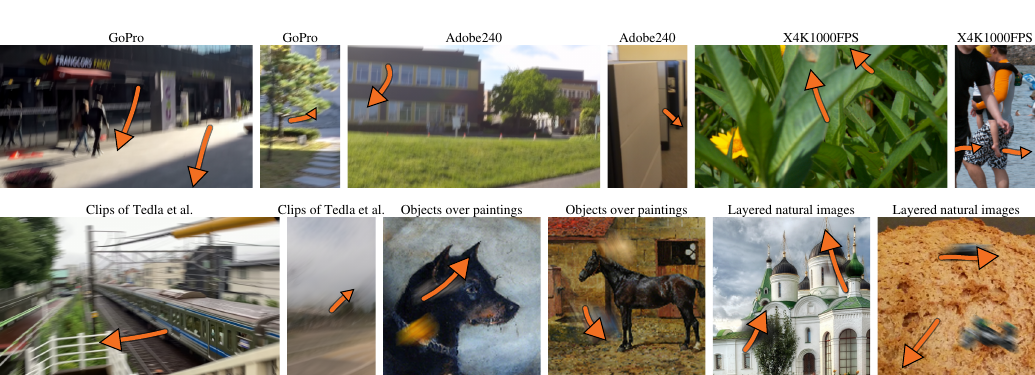}\\[4pt]
\includegraphics[width=\textwidth]{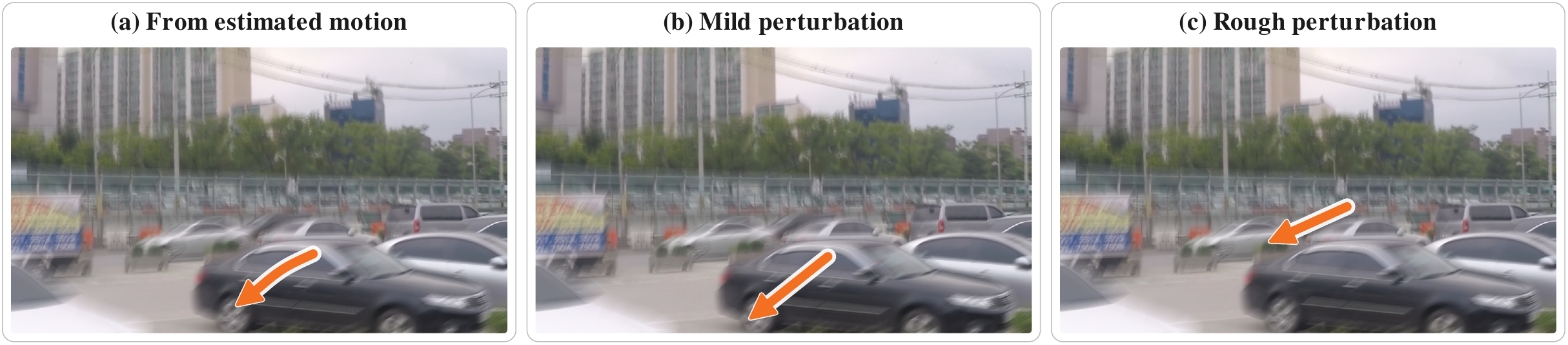}
\caption{\textbf{Arrows are sparse, imprecise strokes; the perturbed test arrows keep only the temporal direction of their source arrow.} Arrows. Top: training arrows as the encoder receives them, on its training blurs, two samples from each training
source (the first two with an arrow in a fixed random order): real clips (GoPro, Adobe240, X4K1000FPS, clips of Tedla et al.;
from RAFT tracks, \cref{tab:supp_arrow_random}) and the two synthetic generators (exact motion, \cref{tab:supp_scene_kinds}).
Bottom: the arrow perturbations of main-paper \cref{fig:control}(b). From the same RAFT-derived source arrow, the mild
(rotation about 7$^\circ$, length $\times$0.8--1.25) and rough (about 25$^\circ$, $\times$0.4--2.5) variants straighten it
and perturb its angle, length, placement and endpoints, keeping its temporal direction.}
\label{fig:supp_arrow_perturb}
\end{figure*}

\paragraph{Real clips.} On real held-out clips, \paper follows in nearly all two-mover commands and region-reversal trials (\cref{tab:supp_control}). Only the region test separates it from Go-with-the-Track~\cite{gowiththetrack2026}, which is
given tracks scaled to the true travel: there \paper follows significantly more often (exact McNemar test); on two movers the difference is not significant. Perturbed arrows and arrows rotated by 35$^\circ$ still work (\cref{tab:supp_control,fig:supp_arrow_perturb}).
Given the true sharp first frame instead of the blur, which a user of a blurred photo does not have, Wan-Move~\cite{wanmove2025} follows
both arrows as often as \paper does from the blur. On the 30 two-mover clips with tracks as long as the drawn arrows, for the two
commands we ran (both arrows forward; one forward, one reversed), it follows in 1.000 and 0.967 of the clips, with no
significant difference from \paper; Tora~\cite{tora2024} from the sharp frame
follows in 0.767 and 0.867. Our advantage on this test therefore comes from working from the blur, not from a better
motion interface.

\paragraph{What the arrow adds.} Two diagnostics on the two-mover clips ask how much of an arrow is used. Arrows whose
axis is read from \paper's own uncued output (median 9.1$^\circ$ from the true axis), with only the sign taken from the
truth, work almost as well as the true ones (\cref{tab:supp_control}). With both arrows turned by 90$^\circ$, 52 of the
60 moving groups still move along the blur's axis, and the 8 that move across it all follow the turned arrow. We
infer that an arrow mostly selects a direction on the axis the blur already fixes; in 8 of 60 groups, against 3 without arrows, it moved the part off
that axis.

\paragraph{More arrows.} Training draws one to three arrows per clip. On the 154 Adobe240 windows we give one to six
arrows simulated from the true motion, nested so that each level adds arrows to the level below, with the sampler noise
fixed (\cref{tab:supp_more_arrows}). The first arrow fixes the direction and removes most of the forward error. Each
further arrow lowers it again, also beyond the three arrows seen in training: six arrows give a significantly lower forward
EPE than one and than three (paired, one sampler seed).

\begin{table}[!htbp]
\centering
\caption{\textbf{Each added arrow lowers the motion error further, also beyond the arrow counts seen in training.} \paper with more arrows from the true motion (Adobe240, 154 windows). Given: mean arrows per window (at most
one per motion group). Forward EPE: flow error to the true clip in its true order. Correct order: share of windows whose
forward EPE is below the reversed-order EPE. First row: \paper (blind), scored in the true order. Bold: best in the column.}
\label{tab:supp_more_arrows}
\footnotesize
\setlength{\tabcolsep}{3.5pt}
\begin{adjustbox}{max width=\columnwidth}
\begin{tabular}{lccc}
\toprule
Arrows requested & Given & Forward EPE $\downarrow$ & Correct order $\uparrow$ \\
\midrule
0 (blind) & 0.00 & 2.160 [1.553, 2.840] & 0.506 \\
1 & 0.93 & 0.638 [0.486, 0.832] & 0.916 \\
2 & 1.81 & 0.581 [0.435, 0.778] & 0.955 \\
3 & 2.68 & 0.563 [0.419, 0.755] & 0.955 \\
4 (beyond training) & 3.54 & 0.542 [0.402, 0.737] & \textbf{0.961} \\
6 (beyond training) & 5.11 & \textbf{0.529 [0.393, 0.721]} & \textbf{0.961} \\
\bottomrule
\end{tabular}
\end{adjustbox}
\end{table}

\paragraph{Arrows from a sentence.} The direction can also be typed. For each commanded part of the two-mover clips we
write one sentence, ``The \emph{name} \emph{place} moves \emph{direction}.'' A vision-language model (Qwen3-VL-8B) names
the part from the blurred image with a box drawn around it, the place is its cell in a $3\times3$ grid, and the direction
is the command rounded to eight compass words. A text-only parser then reads the direction, the same model locates the
named part in the blurred image, and one straight arrow is placed at the center of the returned box. \paper follows both
sentences in 0.967 of commands, the rate of the true arrows (paired difference 0.000 [$-0.048$, $+0.052$];
\cref{tab:supp_control}), although only 19 of the 120 arrows are centered on the true stroke. The parser reads 51 of the
120 sentences, those about a background surface such as grass or sky, as motion of the whole scene. Our first run gave
these sentences no arrow and followed in 0.350 of commands (``objects only''); locating the named surface as for an
object was added after that result. The sentences are written from the true motion, so this tests the interface, not
whether a person can describe the motion from the blur.

\begin{table*}[!tp]
\centering
\caption{\textbf{On real clips \paper follows sparse arrows more often and with lower track error than every trajectory-controlled generator, and stays above 0.86 with perturbed or rotated arrows.} Local control on real held-out clips. Every method gets the same blur and arrows (generators: as point tracks;
Animation from Blur: as its motion guide). Two movers: 30 clips, two moving parts, two commands each; one region reversed: one
region commanded backwards. Follow: share of commands or trials in which every commanded part moves as commanded. Error:
point-track error to the command, median (mean), on the 58 two-mover commands and 64 region trials with a scorable track.
Tora's shortest track is one unit of its 256-pixel canvas. Rotated by 35$^\circ$, two movers: only the both-forward command was
run (30 commands); its Error is taken on both arrowed parts against the true clip played forward (29 commands with a
scorable track). Oracle features: region test only.}
\label{tab:supp_control}
\label{tab:supp_cue_perturb}
\label{tab:supp_axis_sign}
\label{tab:supp_arrow_rot90}
\label{tab:supp_gwtt_paired}
\footnotesize
\setlength{\tabcolsep}{3.5pt}
\begin{adjustbox}{max width=\textwidth}
\begin{tabular}{L{0.30\textwidth}cccc}
\toprule
 & \multicolumn{2}{c}{Two movers (60 commands)} & \multicolumn{2}{c}{One region reversed (66 trials)} \\
\cmidrule(lr){2-3}\cmidrule(lr){4-5}
Method & Follow $\uparrow$ & Error $\downarrow$ & Follow $\uparrow$ & Error $\downarrow$ \\
\midrule
\multicolumn{5}{l}{\textit{Reference errors}} \\
True clip played forward & N/A & 4.67 (8.29) & N/A & 9.55 (18.53) \\
Blur copy (no motion) & N/A & 2.31 (4.03) & N/A & 4.70 (9.23) \\
\midrule
\multicolumn{5}{l}{\textit{Tracks as long as the drawn arrow}} \\
Tora & 0.783 [0.672, 0.887] & 49.32 (51.25) & 0.530 [0.385, 0.672] & 47.23 (52.47) \\
Wan-Move 14B & 0.867 [0.766, 0.964] & 58.05 (61.83) & 0.530 [0.422, 0.645] & 50.61 (58.48) \\
\midrule
\multicolumn{5}{l}{\textit{\paper with other cues}} \\
No arrows & 0.283 [0.197, 0.375] & 2.88 (4.66) & 0.045 [0.000, 0.117] & 5.17 (9.16) \\
Mildly perturbed arrows & 0.950 [0.867, 1.000] & 1.26 (3.25) & 0.939 [0.862, 1.000] & 1.53 (3.67) \\
Roughly perturbed arrows & 0.967 [0.889, 1.000] & 1.25 (3.12) & 0.864 [0.773, 0.942] & 1.88 (5.63) \\
Arrows rotated by 35$^\circ$ & 0.967 [0.897, 1.000] (30) & 0.91 (2.03) & 0.924 [0.864, 0.972] & 1.56 (4.41) \\
Axis from the uncued output, true sign & 0.933 [0.833, 1.000] & 1.27 (3.14) & 0.924 [0.857, 0.973] & 1.67 (3.49) \\
Arrows from sentences & 0.967 [0.917, 1.000] & -- & -- & -- \\
Arrows from sentences, objects only & 0.350 [0.172, 0.533] & -- & -- & -- \\
Oracle features (diagnostic) & -- & -- & 1.000 [1.000, 1.000] & 1.23 (1.81) \\
\midrule
\multicolumn{5}{l}{\textit{Tracks scaled to the true travel (main-paper Table~\ref{tab:control})}} \\
Tora & 0.350 [0.210, 0.500] & 4.63 (10.18) & 0.227 [0.132, 0.328] & 8.99 (17.86) \\
Wan-Move 14B & 0.583 [0.467, 0.696] & 4.58 (7.99) & 0.333 [0.188, 0.478] & 5.96 (11.71) \\
Animation from Blur & 0.833 [0.766, 0.903] & 1.85 (3.50) & 0.273 [0.145, 0.422] & 4.44 (9.29) \\
Go-with-the-Track & 0.933 [0.839, 1.000] & 1.75 (3.70) & 0.652 [0.538, 0.754] & 3.33 (9.16) \\
\oursrow\textbf{\paper (arrows)} & \textbf{0.967 [0.889, 1.000]} & \textbf{1.27 (3.18)} & \textbf{0.939 [0.879, 0.986]} & \textbf{1.64 (3.19)} \\
\bottomrule
\end{tabular}
\end{adjustbox}
\end{table*}

\subsection{Arrows drawn by people}
\label{sec:supp_human_arrows}
\begin{figure*}[!tp]
\centering
\includegraphics[width=\textwidth]{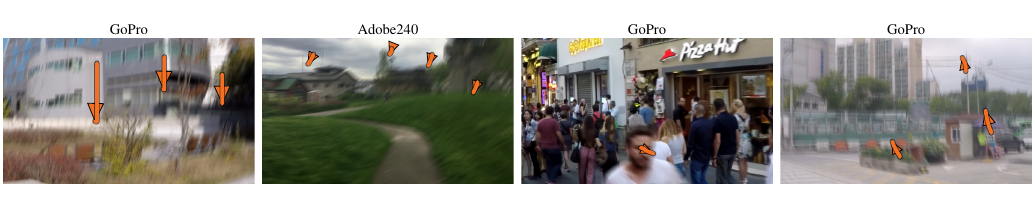}
\caption{\textbf{People draw few, short and imprecise arrows, the cue \paper is built for.} Arrows drawn by one participant, as drawn, on four of the 16 test photos: the photos most participants drew on, and
the participant with the median number of arrows over the 16 photos.}
\label{fig:supp_human_arrows}
\end{figure*}

All cues above are simulated. Here nine people drew arrows on 16 held-out photos of the region-control test (11 GoPro,
5 Adobe240), on a public web page that showed each blurred photo beside a looping clip of its true motion. They drew
one arrow on each moving part, up to five per photo, without tracing the motion exactly. Each trial renders a photo with its arrows as drawn and
again with one arrow reversed. The readout is the direction of the output's mean optical flow (RAFT) inside that
arrow's band (its stroke widened by its thickness). The result is descriptive. \Cref{fig:supp_human_arrows} shows how one
participant drew.

\paper follows arrows drawn by people in most trials (\cref{tab:supp_human_arrows}). Over the participants, the band moves along the arrow in almost every trial and reverses with it in most. Given the same arrows as point tracks, Tora and
Wan-Move less often reverse the band while keeping the others (paired block); Wan-Move ran on a subset of 46 trials, for
compute. The page showed the true clip on purpose: the test measures whether the model follows
arrows drawn by people who know the motion, not whether a person can read the direction from the blur alone.

\paragraph{Participants and consent.} Participation was voluntary. Before drawing, the page told each participant what it
stores: their arrows, the time they drew them, an anonymous account identifier and optional initials. It also said
that the drawings are used only to test arrow-controlled video generation, and that they can stop at any point. No
image, voice or other personal data of participants was collected, and no initials appear in this paper.

\begin{table*}[!tp]
\centering
\caption{\textbf{\paper follows arrows drawn by people in 0.93 of trials and, when one arrow is reversed, reverses only its object more often than Tora and Wan-Move.} \paper on arrows drawn by people (16 held-out photos; both sets pooled). Follows: the band moves along the
arrow as drawn; reversed: against it when that arrow alone is reversed; against as drawn: it already moves against the drawn
arrow (a failure); others kept: every other arrow still follows, counted only on trials with another moving arrow (273 of the 315; 33 of the 46);
rev.\ and kept: both. Paired rows: same trials.}
\label{tab:supp_human_arrows}
\label{tab:supp_human_paired}
\footnotesize
\setlength{\tabcolsep}{3pt}
\begin{adjustbox}{max width=\textwidth}
\begin{tabular}{lcccccc}
\toprule
Photos & Trials & Follows $\uparrow$ & Reversed $\uparrow$ & Against as drawn $\downarrow$ & Others kept $\uparrow$ & Rev.\ and kept $\uparrow$ \\
\midrule
Tora, both sets & 315 & 0.80 [0.71, 0.88] & 0.75 [0.69, 0.81] & 0.18 [0.10, 0.27] & 0.66 [0.57, 0.75] & 0.51 [0.44, 0.59] \\
\midrule
\oursrow Both sets & 315 & 0.93 [0.86, 0.97] & 0.86 [0.74, 0.94] & 0.03 [0.01, 0.04] & 0.80 [0.69, 0.88] & 0.75 [0.63, 0.84] \\
\oursrow GoPro & 223 & 0.93 [0.87, 0.98] & 0.84 [0.68, 0.95] & 0.03 [0.01, 0.05] & 0.77 [0.64, 0.88] & 0.73 [0.57, 0.85] \\
\oursrow Adobe240 & 92 & 0.91 [0.70, 1.00] & 0.90 [0.73, 0.95] & 0.02 [0.00, 0.04] & 0.86 [0.72, 0.97] & 0.80 [0.63, 0.92] \\
\midrule
\multicolumn{7}{l}{\textit{Same 46 trials (Wan-Move's subset)}} \\
Tora & 46 & 0.85 [0.74, 0.93] & 0.70 [0.59, 0.81] & 0.15 [0.07, 0.26] & 0.73 [0.53, 0.89] & 0.54 [0.38, 0.72] \\
Wan-Move & 46 & 0.93 [0.87, 1.00] & 0.78 [0.64, 0.90] & 0.07 [0.00, 0.13] & 0.76 [0.59, 0.90] & 0.63 [0.51, 0.75] \\
\oursrow\paper & 46 & 0.98 [0.93, 1.00] & 0.89 [0.74, 1.00] & 0.00 [0.00, 0.00] & 0.82 [0.61, 0.97] & 0.85 [0.70, 0.97] \\
\midrule
\multicolumn{7}{l}{\textit{Paired differences}} \\
\paper $-$ Tora & 315 & $+0.13$ [$+0.04$, $+0.22$] & $+0.11$ [$-0.02$, $+0.22$] & $-0.16$ [$-0.25$, $-0.07$] & $+0.14$ [$+0.02$, $+0.26$] & $+0.23$ [$+0.10$, $+0.35$] \\
\paper $-$ Wan-Move & 46 & $+0.04$ [$-0.04$, $+0.12$] & $+0.11$ [$-0.03$, $+0.27$] & $-0.07$ [$-0.13$, $+0.00$] & $+0.06$ [$-0.22$, $+0.30$] & $+0.22$ [$+0.06$, $+0.38$] \\
\bottomrule
\end{tabular}
\end{adjustbox}
\end{table*}

\section{Additional experiments and results}
\label{sec:supp_results}

This section gives further evidence for the results of the main paper. It covers reliability across three sampler
seeds; Animation from Blur with a guide from the true motion; other test sets (real camera blur,
I2-2000FPS, X4K1000FPS and B-AIST++); time order and rendered travel as the number of frames changes; the range of recoverable motion; downstream point tracks and camera motion; and the comparison with
PickMoment. The systems, data and protocol are in \cref{sec:supp_details}.

\subsection{Reliability across sampler seeds}
\label{sec:supp_reliability}
The generated videos depend on the sampler seed, so we rerun the Adobe240 and real camera blur (RBI) cells of main-paper
\cref{tab:main} with two more seeds. Only the seed changes, and every video differs from its seed-1 video. The
scores barely move (\cref{tab:supp_seed}): across the three seeds, no cell moves by more than 0.08\,dB PSNR, 0.002
LPIPS or 0.05\,px EPE from the value printed in the main paper. Every gap between the rows keeps its sign and
significance on all three seeds. On held-out GoPro, X4K and Adobe240 windows, three seeds of \paper blind and with
arrows likewise change no cell by more than 0.13\,dB, 0.004 LPIPS or 0.09\,px Track, and 3 of 150 comparisons with the
baselines change significance, all of them borderline.

\begin{table*}[!tp]
\centering
\caption{\textbf{The main-table gaps do not depend on the sampler seed: cells move by at most 0.05\,dB across three seeds, and no gap changes its significance.} Three sampler seeds for main-paper \cref{tab:main}'s Adobe240 and RBI cells (only the seed changed). Cells and
gaps: mean $\pm$ standard deviation over seeds; gaps: paired, first-named row minus second. Every gap is significant on all three seeds,
except (level): not significant on any.}
\label{tab:supp_seed}\label{tab:supp_main_ci}
\footnotesize
\setlength{\tabcolsep}{3.5pt}
\begin{adjustbox}{max width=\textwidth}
\begin{tabular}{lcccccc}
\toprule
 & \multicolumn{3}{c}{Adobe240 (154 windows, 39 recordings)} & \multicolumn{3}{c}{RBI (115 windows, 5 videos)} \\
\cmidrule(lr){2-4}\cmidrule(lr){5-7}
Method & PSNR$\uparrow$ & LPIPS$\downarrow$ & EPE$\downarrow$ & PSNR$\uparrow$ & LPIPS$\downarrow$ & EPE$\downarrow$ \\
\midrule
Wan2.2 (finetuned) & 26.46 $\pm$ 0.05 & 0.124 $\pm$ 0.000 & 0.68 $\pm$ 0.02 & 26.10 $\pm$ 0.04 & 0.173 $\pm$ 0.001 & 1.07 $\pm$ 0.02 \\
\oursrow\paper (blind) & 26.80 $\pm$ 0.01 & 0.115 $\pm$ 0.000 & 0.69 $\pm$ 0.00 & 26.54 $\pm$ 0.01 & 0.164 $\pm$ 0.000 & 1.09 $\pm$ 0.00 \\
\oursrow\paper + arrows & 27.13 $\pm$ 0.01 & 0.110 $\pm$ 0.000 & 0.57 $\pm$ 0.01 & 26.63 $\pm$ 0.01 & 0.162 $\pm$ 0.000 & 1.03 $\pm$ 0.00 \\
\midrule
\multicolumn{7}{l}{\textit{Gaps between rows}} \\
\paper (blind) $-$ Wan2.2 & $+0.34 \pm 0.04$ & $-0.009 \pm 0.000$ & $+0.01 \pm 0.02$ (level) & $+0.44 \pm 0.04$ & $-0.008 \pm 0.001$ & $+0.03 \pm 0.03$ (level) \\
\paper + arrows $-$ \paper (blind) & $+0.33 \pm 0.01$ & $-0.005 \pm 0.000$ & $-0.13 \pm 0.00$ & $+0.09 \pm 0.01$ (level) & $-0.003 \pm 0.000$ & $-0.06 \pm 0.01$ (level) \\
\paper + arrows $-$ Wan2.2 & $+0.67 \pm 0.04$ & $-0.014 \pm 0.000$ & $-0.12 \pm 0.02$ & $+0.53 \pm 0.03$ & $-0.011 \pm 0.001$ & $-0.03 \pm 0.02$ (level) \\
\bottomrule
\end{tabular}
\end{adjustbox}
\end{table*}

\begin{figure*}[p]
\centering
\includegraphics[width=\textwidth]{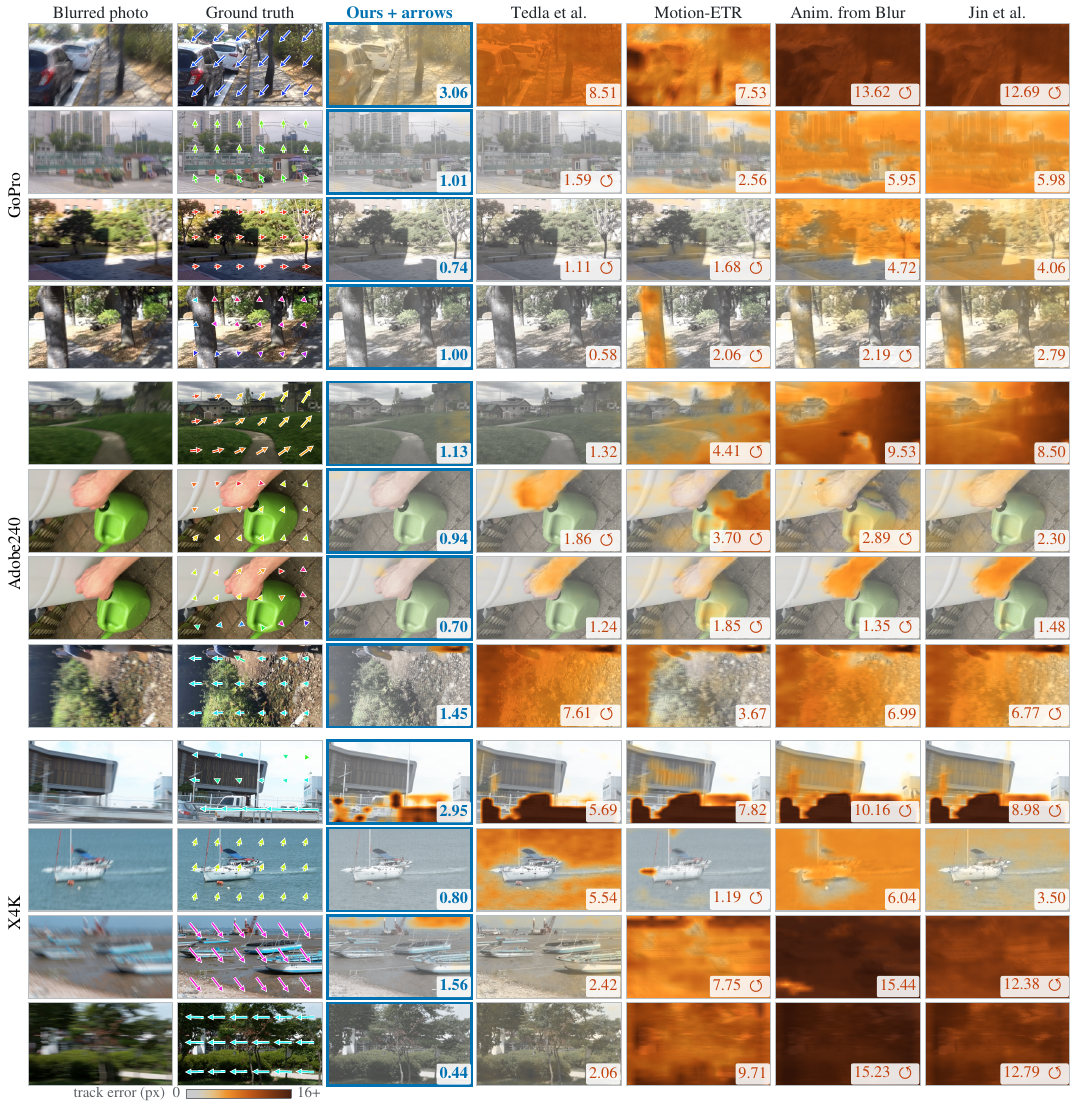}
\caption{\textbf{On held-out clips the motion of \paper stays closest to the truth; Tedla et al.\ is closer in one of the twelve clips.} More held-out results (four clips each from GoPro, Adobe240 and X4K), drawn as in \cref{fig:qualitative}:
middle frames tinted by per-point Track error (0--16+\,px), true motion as arrows ($\times4$) on the ground truth, mean Track
inset. \paper + arrows in the commanded order; blind baselines in the order their Track uses ($\circlearrowleft$: reversed).
Tedla et al.\ has the lower Track in one clip (GoPro, fourth row).}
\label{fig:supp_examples}
\end{figure*}

\subsection{Animation from Blur with a guide from the true motion}
\label{sec:supp_afb}
Animation from Blur~\cite{zhong2022animation} resolves the direction ambiguity with a dense regional motion
guide. \Cref{tab:supp_afb_guide} gives it the guide its paper defines, computed from the true frames, beside \paper with
arrows on GoPro, Adobe240, X4K1000FPS and B-AIST++. The guide sums the RAFT flows between consecutive true sharp frames and quantizes the sum
per pixel into four quadrant directions and a static class below 0.5\,px. It is dense true motion at every pixel, far
more than our one to three arrows, so the comparison favors the baseline. Its released weights are trained on B-AIST++
only, so the other three sets are zero-shot for it.

Guided, its motion error in the forward order falls on every set but stays above that of \paper with arrows (\cref{tab:supp_afb_guide}). Its
PSNR also stays below that of the blur copy (the blurred input repeated as every frame; main-paper \cref{tab:main}) on GoPro, Adobe240 and X4K. On B-AIST++, its home set, \paper with arrows has significantly higher PSNRp in the commanded order and lower LPIPSp (PSNR and LPIPS per image patch, Tedla et al.'s protocol); their motion error (Track, not shown) differs not significantly (paired). Scoring each patch in its better order raises our arrow row far more than the guided baseline
(\cref{tab:supp_afb_guide}), so that scoring overstates the gap.

Two checks show the guide is used as intended. First, the guided output is closer to the truth than to the
time-reversed truth on every set. Second, on B-AIST++ blurred with the authors' own recipe, it reaches a higher forward
PSNRp than its paper reports with a ground-truth guide (27.68 against 25.45\,dB). Given a guide, Animation from Blur does not sample, so
there is one run; we did not use the best-of-$N$ protocol of its paper.

\begin{table*}[!tp]
\centering
\caption{\textbf{Even with a guide from the true motion, Animation from Blur trails \paper with arrows on every set and metric.} Animation from Blur (AfB) with a guide from the true motion, beside \paper. Cells: PSNR / LPIPS / EPE; B-AIST++:
per-patch PSNRp / LPIPSp (Tedla et al.'s protocol) / EPE; X4K: Track (on 14 of the 15 clips) instead of EPE. (n): clips (B-AIST++: videos).
Forward: commanded order (guided) or true order (blind); better: the better order per clip and metric (B-AIST++: per patch).}
\label{tab:supp_afb_guide}
\footnotesize
\setlength{\tabcolsep}{4pt}
\begin{adjustbox}{max width=\textwidth}
\begin{tabular}{llcccc}
\toprule
Order & Method & GoPro (22) & Adobe240 (154) & X4K (15) & B-AIST++ (32) \\
\midrule
Forward & AfB, sampled guide & 22.32 / 0.306 / 2.54 & 23.18 / 0.266 / 2.67 & 21.75 / 0.388 / 7.28 & 19.79 / 0.065 / 0.79 \\
 & AfB, true-motion guide & 20.64 / 0.338 / 1.53 & 20.90 / 0.318 / 1.89 & 20.59 / 0.407 / 4.91 & 22.46 / 0.056 / 0.68 \\
\oursrow\cellcolor{white} & \paper (blind) & 24.98 / 0.143 / 2.04 & 24.85 / 0.135 / 2.16 & 23.06 / 0.220 / 8.22 & 21.77 / 0.057 / 0.75 \\
\oursrow\cellcolor{white} & \paper + arrows & 26.71 / 0.117 / 0.46 & 27.14 / 0.110 / 0.56 & 24.70 / 0.189 / 1.66 & 24.46 / 0.041 / 0.52 \\
\midrule
Better & AfB, sampled guide & 22.67 / 0.302 / 2.23 & 23.58 / 0.263 / 2.35 & 21.99 / 0.386 / 6.93 & 22.92 / 0.047 / 0.60 \\
 & AfB, true-motion guide & 20.64 / 0.338 / 1.53 & 20.90 / 0.318 / 1.89 & 20.59 / 0.407 / 4.91 & 22.73 / 0.056 / 0.68 \\
\oursrow\cellcolor{white} & \paper (blind) & 26.41 / 0.121 / 0.57 & 26.80 / 0.115 / 0.69 & 24.85 / 0.191 / 1.69 & 26.79 / 0.037 / 0.57 \\
\oursrow\cellcolor{white} & \paper + arrows & 26.71 / 0.117 / 0.46 & 27.20 / 0.110 / 0.56 & 24.72 / 0.189 / 1.66 & 26.76 / 0.035 / 0.50 \\
\bottomrule
\end{tabular}
\end{adjustbox}
\end{table*}
\subsection{Other test sets: real camera blur, I2-2000FPS, X4K1000FPS and B-AIST++}
\label{sec:supp_other_sets}
Every table here is scored as in main-paper \cref{tab:main} (protocol in \cref{sec:supp_protocol}), with the extra metric each
set needs (Track; Tedla et al.'s per-patch PSNRp and LPIPSp on B-AIST++).

\begin{figure*}[t]
\centering
\includegraphics[width=\textwidth]{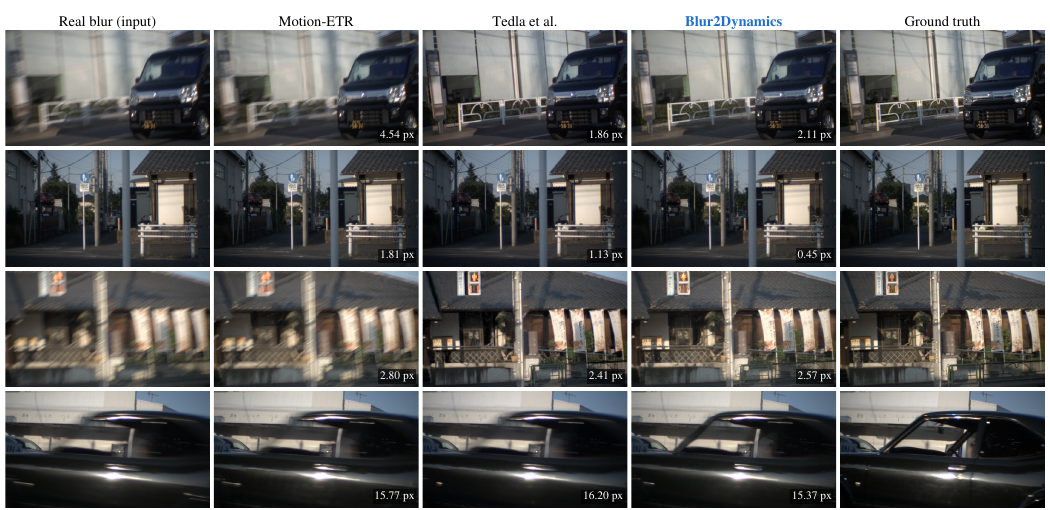}
\caption{\textbf{In typical windows of real camera blur, \paper renders the motion more accurately than Tedla et al.} Real camera blur (RBI): middle output frame of a typical window in each of the four test videos that main-paper
\cref{fig:rbi} does not show, the window whose Track difference (\paper minus Tedla et al.) is closest to that video's
mean difference, among windows with a valid Track. Blind methods in their better time order. Numbers: Track (px; lower is
better).}
\label{fig:supp_rbi}
\end{figure*}

\paragraph{Real camera blur.} On real camera blur, \paper is level with Tedla et al. in PSNR and has lower LPIPS
and motion error (Track, the CoTracker3~\cite{karaev2025cotracker3} point-track error) than every compared method (\cref{tab:supp_real_blur}).
\Cref{fig:supp_rbi} shows a typical window from each of the four test videos that main-paper \cref{fig:rbi} does
not show: the window whose Track difference to Tedla et al.\ is closest to that video's mean difference. Motion-ETR keeps
the input blur; \paper and Tedla et al.\ are close on the first three windows. In the last, the Track difference is small
(15.37 against 16.20\,px), but Tedla et al.'s output is visibly much worse: it leaves the moving car blurred, while \paper
recovers its shape and roofline.

\begin{table}[!htbp]
\centering
\caption{\textbf{On real camera blur \paper has the highest PSNR and the lowest LPIPS and Track error of the tested methods; Tedla et al.\ has the higher per-patch PSNR.} \paper (blind) on real camera blur, scored at $1280\times736$. R: the real 18\,ms blur; S: the plain average of the 9 sharp frames captured during it. Track in px; PSNRp: PSNR per patch (Tedla et al.'s protocol). Bold: lowest error.}
\label{tab:supp_real_blur}
\footnotesize
\setlength{\tabcolsep}{3.5pt}
\begin{adjustbox}{max width=\columnwidth}
\begin{tabular}{L{0.40\columnwidth}cccc}
\toprule
Input / method & PSNR $\uparrow$ & LPIPS $\downarrow$ & Track $\downarrow$ & PSNRp $\uparrow$ \\
\midrule
\multicolumn{5}{l}{\textit{The same 270 beam-splitter windows, real blur (R) or plain average (S)}} \\
Blur copy (R) & 24.36 & 0.260 & 6.86 & 29.09 \\
Blur copy (S) & 28.42 & 0.220 & 6.85 & 36.85 \\
\oursrow\paper (R) & 26.85 & 0.163 & 2.01 & 31.08 \\
\oursrow\paper (S) & 29.76 & 0.108 & 1.78 & 35.50 \\
\midrule
\multicolumn{5}{l}{\textit{The 5 official test videos (115 windows), real blur}} \\
Blur copy (R) & 24.00 & 0.265 & 6.66 & 29.13 \\
Motion-ETR (R) & 24.06 & 0.250 & 4.52 & 29.59 \\
Tedla et al., GoPro (R) & 25.80 & 0.212 & 3.40 & 30.99 \\
Tedla et al., in-the-wild (R) & 26.34 & 0.193 & 2.83 & 31.63 \\
\oursrow\textbf{\paper (R)} & 26.54 & \textbf{0.164} & \textbf{2.18} & 31.18 \\
\bottomrule
\end{tabular}
\end{adjustbox}
\end{table}

\paragraph{I2-2000FPS: protocol and rows.}\label{sec:supp_i2}
The I2-2000FPS column of main-paper \cref{tab:main} uses 31 windows of the 2000\,fps videos~\cite{chennuri2024quiver},
motion-matched to GoPro: each clip's exposure is the shortest centred window of 11 to 150 frames whose true motion reaches
2.79\,px, the median of the held-out GoPro windows (\cref{tab:supp_i2}).

Blind, \paper has higher PSNR and lower LPIPS and EPE than every compared method (\cref{tab:supp_i2}). With arrows, its forward-order EPE is also below Tedla et al.'s. The last two
rows of \cref{tab:supp_i2} give the paired differences with Tedla et al.

\begin{table*}[!tp]
\centering
\caption{\textbf{On high-speed I2-2000FPS clips \paper has the best fidelity and motion error of the tested methods, blind and with arrows.} I2-2000FPS (31 windows). Blind rows: better time order per window and metric; ``+ arrows'', oracle and true-motion-guide rows:
commanded order; EPE, forward: true order only. Last two rows: paired differences, each column under its own order rule.
$^\ddagger$Scored on its own three output frames.}
\label{tab:supp_i2}
\footnotesize
\setlength{\tabcolsep}{3.5pt}
\begin{adjustbox}{max width=\textwidth}
\begin{tabular}{lcccc}
\toprule
Method & PSNR $\uparrow$ & LPIPS $\downarrow$ & EPE $\downarrow$ & EPE, forward $\downarrow$ \\
\midrule
Blur copy & 27.00 [25.56, 28.52] & 0.257 [0.224, 0.295] & 4.55 [3.54, 5.58] & 4.59 [3.59, 5.62] \\
\paper, frozen encoder & 26.98 [25.55, 28.48] & 0.257 [0.225, 0.294] & 4.48 [3.46, 5.51] & 4.59 [3.58, 5.62] \\
\paper, features zeroed & 27.78 [26.33, 29.24] & 0.205 [0.181, 0.231] & 3.34 [2.38, 4.35] & 4.31 [3.07, 5.64] \\
Wan2.2 (finetuned) & 29.19 [27.68, 30.76] & 0.134 [0.117, 0.153] & 1.74 [1.23, 2.27] & 5.24 [3.60, 7.21] \\
\paper + oracle features & 31.00 [29.59, 32.46] & 0.092 [0.082, 0.104] & 1.15 [0.79, 1.57] & 1.15 [0.79, 1.57] \\
\midrule
Animation from Blur & 25.28 [23.98, 26.67] & 0.294 [0.260, 0.334] & 4.41 [3.47, 5.38] & 4.94 [3.98, 5.95] \\
Deblur + I2V$^\ddagger$ & 26.57 [24.95, 28.27] & 0.164 [0.145, 0.183] & 4.40 [3.48, 5.36] & 4.50 [3.59, 5.45] \\
Jin et al. & 27.35 [25.94, 28.78] & 0.222 [0.197, 0.249] & 3.83 [2.93, 4.75] & 4.48 [3.50, 5.49] \\
Motion-ETR & 27.26 [25.86, 28.64] & 0.230 [0.202, 0.258] & 3.58 [2.81, 4.37] & 5.58 [4.40, 6.88] \\
Animation from Blur + true-motion guide & 23.81 [22.44, 25.20] & 0.326 [0.289, 0.367] & 3.56 [2.78, 4.37] & 3.56 [2.78, 4.37] \\
Tedla et al. & 27.60 [26.23, 29.03] & 0.147 [0.131, 0.163] & 2.22 [1.66, 2.87] & 5.61 [3.99, 7.50] \\
\midrule
\oursrow\paper (blind) & 29.93 [28.67, 31.21] & 0.116 [0.102, 0.131] & 1.58 [1.22, 1.96] & 3.24 [2.06, 4.67] \\
\oursrow\paper + arrows & 30.26 [28.92, 31.69] & 0.112 [0.098, 0.125] & 1.43 [1.07, 1.81] & 1.43 [1.07, 1.81] \\
\midrule
\paper (blind) $-$ Tedla et al. & $+2.32$ [$+1.61$, $+3.07$] & $-0.031$ [$-0.042$, $-0.020$] & $-0.64$ [$-1.34$, $-0.08$] & $-2.37$ [$-4.94$, $+0.13$] \\
\paper + arrows $-$ Tedla et al. & $+2.65$ [$+1.97$, $+3.41$] & $-0.035$ [$-0.046$, $-0.025$] & $-0.79$ [$-1.49$, $-0.24$] & $-4.18$ [$-6.13$, $-2.44$] \\
\bottomrule
\end{tabular}
\end{adjustbox}
\end{table*}

\begin{table}[!htbp]
\centering
\caption{\textbf{On X4K \paper has the best fidelity of the tested methods and a lower Track error than every published baseline (1.69 vs.\ 2.38\,px for Tedla et al.); the renderer without features is level with it in motion.} X4K1000FPS (15 test clips; Track on the 14 with valid ground-truth flow). Blind rows in the better time order,
``+ arrows'' and the true-motion guide in the commanded order.}
\label{tab:supp_x4k}
\footnotesize
\setlength{\tabcolsep}{3.5pt}
\begin{adjustbox}{max width=\columnwidth}
\begin{tabular}{lcccc}
\toprule
Method & PSNR $\uparrow$ & LPIPS $\downarrow$ & Track $\downarrow$ & EPE $\downarrow$ \\
\midrule
Blur copy & 24.26 & 0.367 & 7.22 & 4.22 \\
\paper, frozen encoder & 23.24 & 0.374 & 7.11 & 4.14 \\
\paper, features zeroed & 23.30 & 0.338 & 6.58 & 3.81 \\
Wan2.2 (finetuned) & 24.22 & 0.216 & 1.65 & 1.00 \\
\paper + oracle features & 25.69 & 0.150 & 0.92 & 0.61 \\
\midrule
Animation from Blur & 21.99 & 0.386 & 6.93 & 4.11 \\
Jin et al. & 24.41 & 0.325 & 5.88 & 3.34 \\
Animation from Blur + true-motion guide & 20.59 & 0.407 & 4.91 & 3.04 \\
Motion-ETR & 22.83 & 0.347 & 4.46 & 2.72 \\
Deblur + I2V & 24.01 & 0.255 & 4.22 & 4.14 \\
Tedla et al. & 23.83 & 0.233 & 2.38 & 2.47 \\
\oursrow\paper (blind) & 24.85 & 0.191 & 1.69 & 0.99 \\
\oursrow\paper + arrows & 24.70 & 0.189 & 1.66 & 1.10 \\
\bottomrule
\end{tabular}
\end{adjustbox}
\end{table}

\begin{table}[!htbp]
\centering
\caption{\textbf{Under Tedla et al.'s own B-AIST++ protocol, \paper (blind) has the best fidelity and, with arrows, the lowest EPE; blind, Tedla et al.\ has a slightly lower EPE.} B-AIST++ (32 videos), Tedla et al.'s patch-wise protocol. Blind rows: PSNRp / LPIPSp in the better order per patch;
guided rows: commanded order. Bold / underlined: best / second best within each ranked block.}
\label{tab:supp_baist}
\footnotesize
\setlength{\tabcolsep}{3.5pt}
\begin{adjustbox}{max width=\columnwidth}
\begin{tabular}{lccc}
\toprule
Method & PSNRp $\uparrow$ & LPIPSp $\downarrow$ & EPE $\downarrow$ \\
\midrule
\multicolumn{4}{l}{\textit{Not ranked}} \\%
Blur copy & 21.52 & 0.104 & 0.73 \\%
\paper, frozen encoder & 21.87 & 0.103 & 0.72 \\%
\paper, features zeroed & 22.81 & 0.086 & 0.67 \\%
Wan2.2 (finetuned) & 25.71 & 0.040 & 0.56 \\%
\paper + oracle features & 27.55 & 0.026 & 0.41 \\%
\midrule
Deblur + I2V & 23.99 & 0.062 & 1.04 \\
Motion-ETR & 22.86 & 0.069 & 0.71 \\
Jin et al. & 22.97 & 0.076 & 0.70 \\
Animation from Blur & 22.92 & \second{0.047} & 0.60 \\
Tedla et al. & \second{26.46} & \best{0.037} & \best{0.53} \\
\oursrow\paper (blind) & \best{26.79} & \best{0.037} & \second{0.57} \\
\midrule
\multicolumn{4}{l}{\textit{With motion guidance, commanded order}} \\%
Animation from Blur + true-motion guide & 22.46 & 0.056 & 0.68 \\
\oursrow\paper + arrows & \best{24.46} & \best{0.041} & \best{0.52} \\
\bottomrule
\end{tabular}
\end{adjustbox}
\end{table}

\paragraph{X4K1000FPS.}\label{sec:supp_x4k}
\Cref{tab:supp_x4k} gives the rows of main-paper \cref{tab:main} on the 15 X4K1000FPS test clips, which were
held out from training. \paper has the highest PSNR and the lowest LPIPS, Track and EPE of the compared methods.
Oracle features lower its motion error further.

\paragraph{B-AIST++.} \Cref{tab:supp_baist} gives main-paper \cref{tab:main}'s rows on B-AIST++ under Tedla et al.'s patch-wise protocol.
Blind, \paper has the best PSNRp and ties Tedla et al.\ for the best LPIPSp, while Tedla et al.\ has the lowest EPE; with
guidance, \paper with arrows is better than Animation from Blur with the true-motion guide on all three.

\subsection{Time order and travel by the number of frames}
\label{sec:supp_trained_length}
Blind scores such as Track use the better of the two time orders, so they do not show how often the output plays in the true order.
\Cref{tab:supp_time_size} measures this directly on 39 validation windows, together with how much of the true motion
the output renders. At the trained length ($K=8$: the encoder reads 8 frames, the renderer outputs nine), blind \paper plays in the true order in only part of the windows and renders about half of the true travel (the distance moved during the exposure). With one arrow per moving part the order is almost always right. With oracle features (the true video's features), the renderer renders more of the travel.

\noindent\colorbox{gray!12}{\parbox{\dimexpr\linewidth-2\fboxsep\relax}{\textbf{Key finding.}
Without retraining, the number of output frames can change: with arrows the output keeps the true time order at 4 and 16
frames as well as at the trained 8, and more frames spread the motion over more steps. At 32 frames the order is lost, even
with oracle features, so that length is outside the renderer's range.}}
\par\smallskip
\noindent Asking for 16 frames instead of 8 lowers the motion per frame, though by less than in the truth; at 4 frames the
output with arrows covers about a third of the true travel. The encoder's feature error stays close to its value at the trained length
at 4 and 16 frames and rises at 32 (\cref{tab:supp_time_size}).

\begin{table}[!htbp]
\centering
\caption{\textbf{Without retraining, \paper reads the blur at other frame counts: with arrows the time order stays right from 4 to 16 frames and fails only at 32.} Time order, travel (distance moved during the exposure) and motion per frame versus $K$, the number of frames the
encoder reads (the renderer generates $K+1$; no retraining); 39 validation windows from 29 scenes, one 33-frame exposure each;
means over windows.
Feature error: direction-invariant feature error relative to the repeated-blur floor (below 1: better).}
\label{tab:supp_time_size}
\footnotesize
\setlength{\tabcolsep}{4pt}
\begin{adjustbox}{max width=\columnwidth}
\begin{tabular}{lcccc}
\toprule
$K$ & 4 & 8 (trained) & 16 & 32 \\
\midrule
\multicolumn{5}{l}{\textit{True order: share of outputs in the true time order} $\uparrow$} \\
Oracle features & 1.00 & 1.00 & 1.00 & 0.41 \\
\paper (blind) & 0.54 & 0.67 & 0.64 & 0.59 \\
\paper + arrows & 0.95 & 0.97 & 0.97 & 0.33 \\
\midrule
\multicolumn{5}{l}{\textit{Travel: rendered / true travel (1 = all of the true travel)}} \\
Oracle features & 0.41 & 0.75 & 0.84 & 0.44 \\
\paper (blind) & 0.24 & 0.52 & 0.63 & 0.23 \\
\paper + arrows & 0.34 & 0.55 & 0.63 & 0.33 \\
\midrule
\multicolumn{5}{l}{\textit{Motion per frame (px; compare with the ground truth row)}} \\
Ground truth & 12.01 & 6.22 & 3.14 & 1.59 \\
Oracle features & 4.92 & 4.64 & 2.74 & 2.24 \\
\paper (blind) & 2.85 & 3.24 & 2.09 & 1.62 \\
\paper + arrows & 4.04 & 3.41 & 2.14 & 1.80 \\
\midrule
\multicolumn{5}{l}{\textit{Feature error} $\downarrow$} \\
Encoder & 0.563 & 0.551 & 0.565 & 0.613 \\
\bottomrule
\end{tabular}
\end{adjustbox}
\end{table}

\subsection{Range of recoverable motion}
\label{sec:supp_range}
This subsection gives the test behind main-paper \cref{fig:range}.

\paragraph{Test.} The test isolates the amount of motion. A textured disc 128\,px wide moves in a straight line at
constant speed over a static natural-image background. It is rendered at $1280\times736$ with $4\times$ supersampling,
so its position is known exactly at every time. Its travel during the exposure is swept from 0 to 512\,px in doubling
steps, along four directions (0, 45, 90 and 135$^\circ$) and over two textures with their own backgrounds: 8 clips per
travel, 104 in all. In each output frame the disc is found by matching its known texture.

\paragraph{When motion counts as reproduced.} At a given travel, at least 80\% of the clips must move the disc at least
half the true distance, within 30$^\circ$ of the true direction, and by more than it moves in outputs for a static scene. A blind
output only has to get the axis right, since it has no cue for the time order; with an arrow, the sign must be right
too.

\begin{figure}[!htbp]
\centering
\includegraphics[width=\columnwidth]{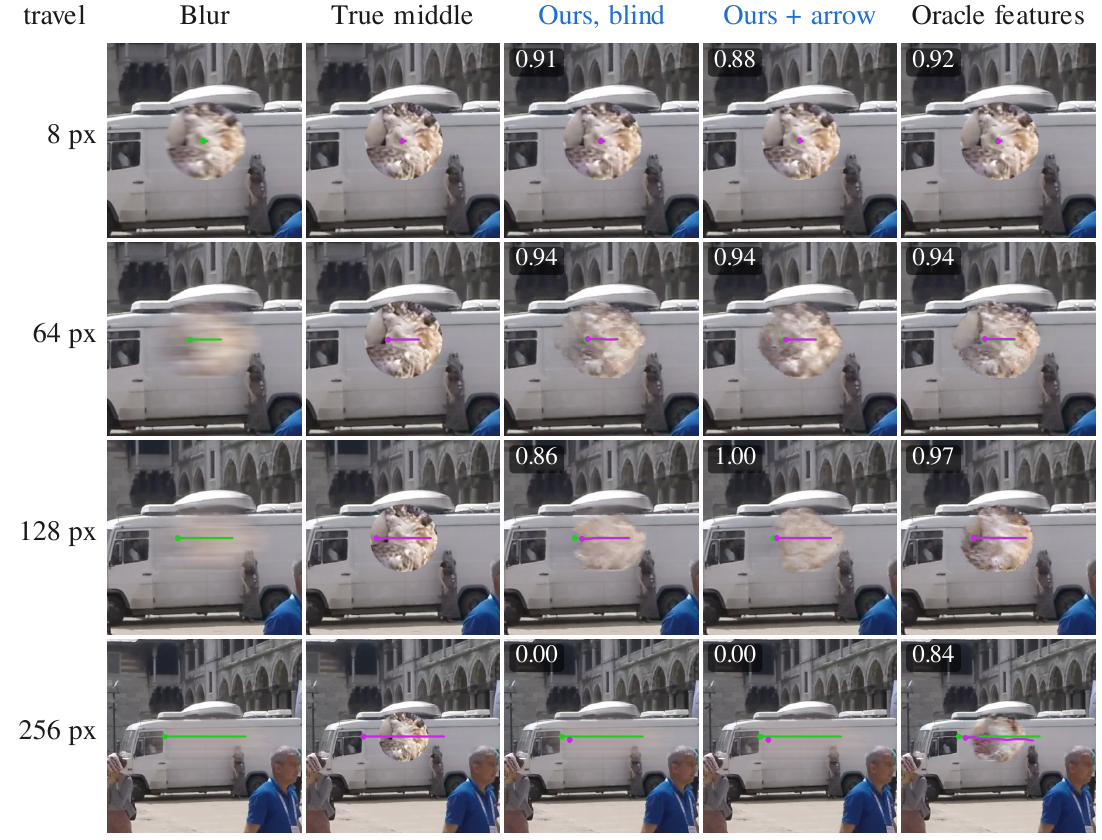}
\caption{\textbf{The recoverable motion range is limited: from about 128\,px of travel the predicted features lose the motion, while oracle features still render it at 256\,px.} Motion-range examples at 8, 64, 128 and 256\,px of travel (rows). Columns: blur, true middle frame, and
middle frames of \paper (blind), with an arrow and with oracle features. Green: true path; magenta: measured positions;
numbers: rendered / true travel (1 = all of it), as in main-paper \cref{fig:range}(b).}
\label{fig:supp_range_panels}
\end{figure}

\paragraph{Results.} Blind, motion is reproduced from 8\,px; with an arrow or with oracle features, from 4\,px. All
three keep about the right amount through 64\,px. From 128\,px the blind and arrow outputs lose the disc or render the
wrong amount in most clips, while oracle features hold up to 256\,px and fail at 512\,px. Tedla et al. reproduce motion
from 4\,px but break at 64\,px, and Motion-ETR never reaches the 80\% level. \Cref{fig:supp_range_panels} shows one clip
per travel.

\paragraph{Small travel.} When nothing moves, outputs jitter by about 1.4\,px on average. Below about 4\,px, an arrow
can set the direction of this jitter-sized motion but not its true amount.

\paragraph{Limits.} The test uses one object size and one synthetic object type, and every true direction points right
or down, so it does not separate a learned preference for those directions.

\subsection{Downstream use: point tracks and camera motion}
\label{sec:supp_downstream}
Tedla et al.~\cite{tedla2025generating} pass their output video to MegaSaM~\cite{li2025megasam}. We run two
downstream tools on the stored outputs of every method for the GoPro and X4K windows and compare each result with the
same tool's result on the true clip (\cref{tab:supp_downstream}). Methods without a direction input take the better
time order.

\paragraph{Point tracks.} CoTracker3~\cite{karaev2025cotracker3} tracks a 16-pixel lattice of points from the middle
frame. The error is the mean distance to the true clip's tracks at matched times, over the points that move at least
2\,px in the true clip. \paper has the lowest error, and every paired difference to another method excludes zero
(\paper + arrows minus Tedla et al.'s general model: $-1.20$ [$-2.17$, $-0.43$]\,px).

\paragraph{Camera motion.} MegaSaM estimates depth and camera poses from each clip, with every frame a keyframe. We take
the image shift between the first and last matched frames that its poses and depth imply, and report its distance to the
true clip's shift on the 28 windows where the true shift exceeds 2\,px (8.53\,px on average). \paper has the lowest mean
error but is not separated from Tedla et al.'s two models (\paper minus the general model: $-0.56$ [$-1.35$,
$+0.24$]\,px). From a blur copy MegaSaM recovers almost no camera motion.

\paragraph{Limits.} With 7 to 16 frames MegaSaM marks the camera poses as not observable in 466 of 470 runs, so we use
only the image-space shift, not the 3D path. The point cloud of a single frame comes from MegaSaM's monocular depth
networks and looks alike for every input, the blur copy included, so it is not evidence about the video. Deblur + I2V
starts at the center of the exposure and is compared over the frames it shares with the true clip.

\begin{table}[!htbp]
\centering
\caption{\textbf{Downstream tools agree: point tracks of the \paper videos are closest to the true clip's, and its camera-shift error is the lowest, though not separated from Tedla et al.'s.} Downstream readouts against the true clip's. Track: CoTracker3 point-track error (px) on 22 GoPro and 15 X4K
windows. Camera: error (px) of the camera-induced image shift recovered by MegaSaM, on the 28 windows whose true shift
exceeds 2\,px. Mean [95\% interval over recordings]. Bold / underlined: best / second best. N/A: not applicable.}
\label{tab:supp_downstream}
\footnotesize
\setlength{\tabcolsep}{3pt}
\begin{adjustbox}{max width=\columnwidth}
\begin{tabular}{lcccc}
\toprule
 & \multicolumn{3}{c}{Track $\downarrow$} & Camera $\downarrow$ \\
\cmidrule(lr){2-4}\cmidrule(lr){5-5}
Method & All (37) & GoPro & X4K & 28 windows \\
\midrule
Animation from Blur & 6.26 [4.70, 8.23] & 4.21 & 9.27 & 7.99 [5.60, 11.02] \\
Blur copy (no motion) & 5.80 [4.43, 7.45] & 4.06 & 8.35 & 8.46 [6.15, 11.37] \\
Deblur + I2V & 5.23 [3.96, 6.82] & 3.46 & 7.83 & 4.13 [2.95, 5.63] \\
Jin et al. & 4.50 [3.25, 6.07] & 2.72 & 7.09 & 5.22 [3.29, 7.83] \\
Animation from Blur + true-motion guide & 4.20 [2.99, 5.74] & 2.52 & 6.66 & 3.96 [2.22, 6.32] \\
Motion-ETR & 2.93 [2.08, 4.01] & 1.64 & 4.82 & 2.80 [1.46, 4.59] \\
Tedla et al., GoPro model & N/A & 1.87 & N/A & 1.55 [0.34, 3.40] (19) \\
Tedla et al., general model & 2.47 [1.57, 3.61] & 1.64 & 3.69 & 2.00 [1.13, 3.13] \\
\midrule
\oursrow\textbf{\paper (blind)} & \second{1.39 [1.00, 1.91]} & \second{0.99} & \second{1.98} & \best{1.44 [0.83, 2.28]} \\
\oursrow\textbf{\paper + arrows} & \best{1.26 [0.88, 1.78]} & \best{0.82} & \best{1.91} & \second{1.52 [0.68, 2.84]} \\
\bottomrule
\end{tabular}
\end{adjustbox}
\end{table}

\subsection{Comparison with PickMoment}
\label{sec:supp_pickmoment}
PickMoment~\cite{shin2026pickmoment} fine-tunes an image generator (FLUX.2-klein) on GoPro to output, from one blurred
image and a requested part of the exposure, the image of that moment in a single deterministic pass; a video is a stack of
independent queries. It differs from \paper in two ways. It has no representation of the motion and no temporal model:
each frame is generated on its own, whereas \paper infers dynamics features of the whole exposure and renders all frames
jointly with a video model. And it takes no input besides the blur and the time, so the direction of the motion cannot be
set: which way a clip plays is fixed by what it learned, and we score it in its better time order per window, like every
blind method (\cref{tab:supp_pickmoment}).

\begin{figure*}[!tp]
\centering
\includegraphics[width=\textwidth]{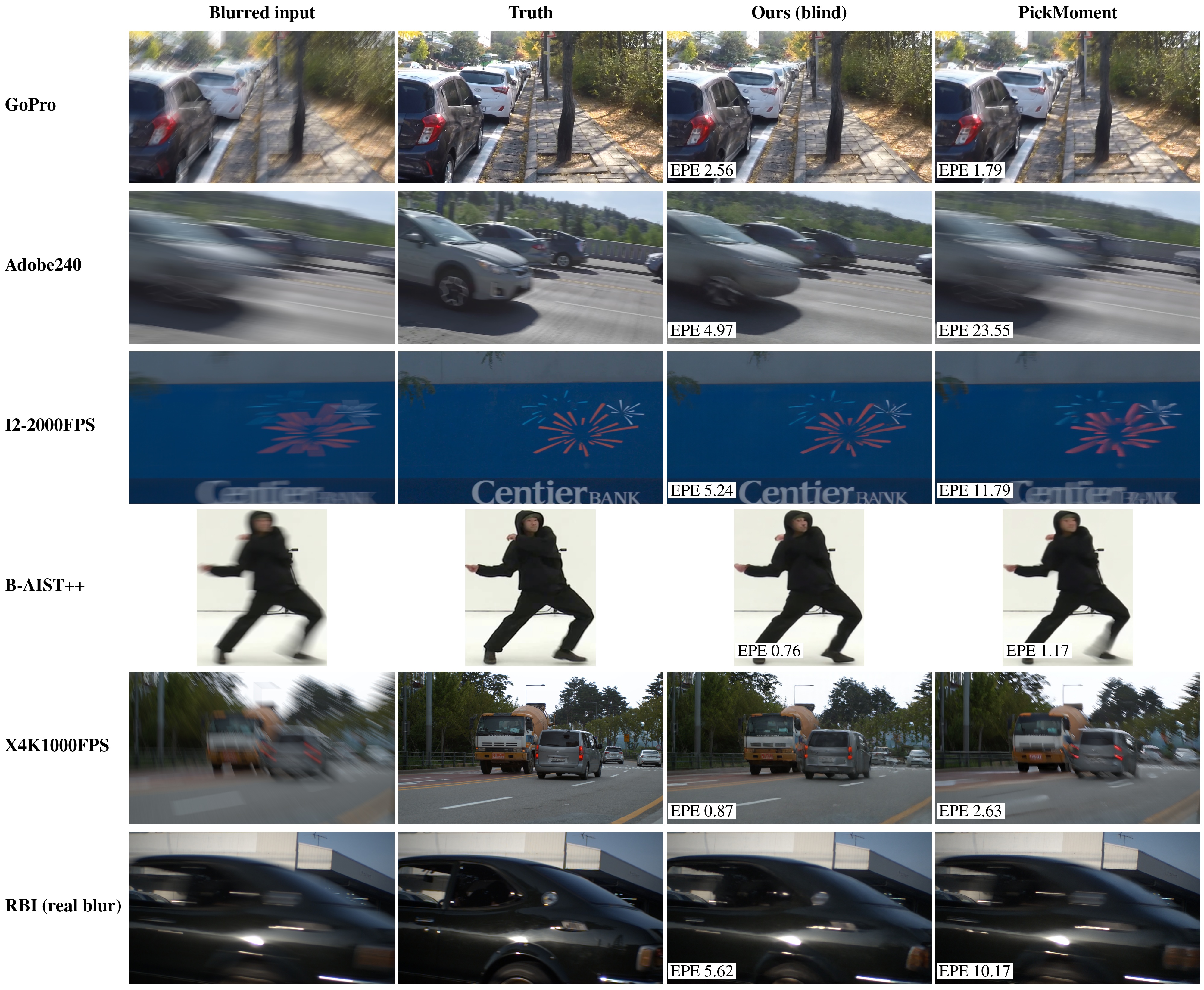}
\caption{\textbf{On the window with the most motion in each held-out set, \paper has the lower EPE than PickMoment in five of six.} PickMoment and \paper (blind) on one window per held-out set, chosen by one rule only: the window with the most
true motion (highest blur-copy EPE). Each cell is the middle frame; the corner gives that window's EPE.}
\label{fig:supp_pickmoment}
\end{figure*}

Its released weights are trained on GoPro only. On GoPro it has better picture quality than \paper with no significant
difference in motion error. Outside GoPro it is worse in motion, picture or both on every set; on real camera blur, for
example, its PSNR is significantly lower and its motion error more than twice ours (\cref{tab:supp_pickmoment}). Without
arrows, on the fresh drop scenes of \cref{tab:supp_default_tokens}, it renders the settling order in 0.672 of the pairs, not
tested against ours; it is not among the methods of main-paper \cref{fig:motiv_default}.
\Cref{fig:supp_pickmoment} shows one window per set, the one with the most true motion: PickMoment leaves fast
objects smeared (Adobe240), distorts the fireworks logo (I2-2000FPS) and ghosts the moving foot (B-AIST++).

\begin{table}[!tp]
\centering
\caption{\textbf{\paper (blind) has a lower motion error than PickMoment on every held-out set, significantly on four; PickMoment has the higher PSNR on GoPro and Adobe240.} PickMoment (trained on GoPro only) and \paper (blind) on every held-out set, both in the better time order per
window; B-AIST++: Tedla et al.'s patch protocol (\cref{tab:supp_baist}). Bold: the better of the two where the paired
difference over the same windows is significant. PickM.: PickMoment.}
\label{tab:supp_pickmoment}
\footnotesize
\setlength{\tabcolsep}{2.5pt}
\begin{adjustbox}{max width=\columnwidth}
\begin{tabular}{lcOcOcO}
\toprule
 & \multicolumn{2}{c}{PSNR $\uparrow$} & \multicolumn{2}{c}{LPIPS $\downarrow$} & \multicolumn{2}{c}{EPE $\downarrow$} \\
\cmidrule(lr){2-3}\cmidrule(lr){4-5}\cmidrule(lr){6-7}
Set (windows) & PickM. & Ours & PickM. & Ours & PickM. & Ours \\
\midrule
GoPro (22) & \textbf{27.64} & 26.41 & \textbf{0.110} & 0.121 & 0.67 & 0.57 \\
Adobe240 (154) & \textbf{27.65} & 26.80 & 0.117 & 0.115 & 1.16 & \textbf{0.69} \\
I2-2000FPS (31) & 28.69 & \textbf{29.93} & 0.146 & \textbf{0.116} & 3.13 & \textbf{1.58} \\
B-AIST++ (32) & 23.02 & \textbf{26.79} & 0.051 & \textbf{0.037} & 0.59 & 0.57 \\
RBI (115) & 25.15 & \textbf{26.54} & 0.223 & \textbf{0.164} & 2.51 & \textbf{1.09} \\
X4K (15) & 24.43 & 24.85 & 0.260 & \textbf{0.191} & 1.87 & \textbf{0.99} \\
\bottomrule
\end{tabular}
\end{adjustbox}
\end{table}

\section{Evaluation protocol and data}
\label{sec:supp_details}

\begin{table*}[!tp]
\centering
\caption{\textbf{The better-time-order convention does not create the ranking: on moving pixels \paper (blind) stays lowest on three of four sets, and with arrows the true order is almost always the better one.} Blind methods scored in the better time order (\cref{sec:supp_protocol}), from main-paper \cref{tab:main}'s
per-window scores. Top: EPE as printed / on moving pixels / in the true order only; Back: share of windows in which the
reversed order has the lower EPE. Bottom: Adobe240 PSNR under each order rule. AfB: Animation from Blur.}
\label{tab:supp_order}
\scriptsize
\setlength{\tabcolsep}{2.2pt}
\begin{adjustbox}{max width=\textwidth}
\begin{tabular}{lcccccccc}
\toprule
 & \multicolumn{2}{c}{GoPro (22)} & \multicolumn{2}{c}{Adobe240 (154)} & \multicolumn{2}{c}{I2-2000FPS (31)} & \multicolumn{2}{c}{RBI (115)} \\
\cmidrule(lr){2-3}\cmidrule(lr){4-5}\cmidrule(lr){6-7}\cmidrule(lr){8-9}
Method & EPE / moving / forward $\downarrow$ & Back & EPE / moving / forward $\downarrow$ & Back & EPE / moving / forward $\downarrow$ & Back & EPE / moving / forward $\downarrow$ & Back \\
\midrule
Blur copy & 2.42 / 2.82 / 2.43 & 0.64 & 2.47 / 3.32 / 2.48 & 0.51 & 4.55 / 6.68 / 4.59 & 0.45 & 3.12 / 4.90 / 3.15 & 0.45 \\
Animation from Blur & 2.23 / 2.56 / 2.54 & 0.55 & 2.35 / 3.06 / 2.67 & 0.52 & 4.41 / 6.39 / 4.94 & 0.58 & 4.14 / 6.27 / 4.42 & 0.52 \\
Deblur + I2V & 2.14 / 2.39 / 2.42 & 0.45 & 2.24 / 3.03 / 2.53 & 0.54 & 4.40 / 8.75 / 4.50 & 0.16 & 3.20 / 4.74 / 3.40 & 0.50 \\
Jin et al. & 1.61 / 1.86 / 2.04 & 0.27 & 1.77 / 2.54 / 2.40 & 0.42 & 3.83 / 5.68 / 4.48 & 0.48 & 3.47 / 5.57 / 4.23 & 0.40 \\
Tedla et al. & 1.10 / 1.29 / 1.97 & 0.36 & 1.29 / 1.87 / 2.60 & 0.40 & 2.22 / 3.03 / 5.61 & 0.52 & 1.97 / 2.83 / 3.57 & 0.46 \\
Motion-ETR & 1.00 / 1.13 / 2.80 & 0.68 & 1.55 / 2.13 / 2.84 & 0.53 & 3.58 / 4.86 / 5.58 & 0.58 & 2.20 / 3.02 / 3.70 & 0.53 \\
Wan2.2 (finetuned) & 0.88 / 1.01 / 2.67 & 0.59 & 0.69 / 1.34 / 3.22 & 0.56 & 1.74 / 2.77 / 5.24 & 0.58 & 1.09 / 1.85 / 2.89 & 0.54 \\
\oursrow\paper (blind) & 0.57 / 0.62 / 2.04 & 0.27 & 0.69 / 1.31 / 2.16 & 0.49 & 1.58 / 2.53 / 3.24 & 0.42 & 1.09 / 1.87 / 3.60 & 0.43 \\
\midrule
AfB, true-motion guide & 1.53 / 1.63 / 1.53 & 0.00 & 1.89 / 2.35 / 1.89 & 0.00 & 3.56 / 5.02 / 3.56 & 0.00 & 3.33 / 5.37 / 3.33 & 0.05 \\
\oursrow\paper + arrows & 0.46 / 0.51 / 0.46 & 0.00 & 0.56 / 1.18 / 0.56 & 0.05 & 1.43 / 2.43 / 1.43 & 0.03 & 1.03 / 1.69 / 1.03 & 0.04 \\
\bottomrule
\end{tabular}
\end{adjustbox}\par\medskip
\begin{adjustbox}{max width=\textwidth}
\begin{tabular}{lcccccccO}
\toprule
Adobe240 PSNR (dB) $\uparrow$ & AfB & Blur copy & Motion-ETR & Wan2.2 (finetuned) & Jin et al. & Deblur + I2V & Tedla et al. & \paper (blind) \\
\midrule
Better order per window and metric (main table) & 23.58 & 25.70 & 25.87 & 26.48 & 26.61 & 26.78 & 26.81 & 26.80 \\
One order per window (by EPE) & 23.57 & 25.70 & 25.80 & 26.48 & 26.59 & 26.76 & 26.75 & 26.76 \\
Forward (true order) only & 23.18 & 25.70 & 24.49 & 24.15 & 25.18 & 26.30 & 24.99 & 24.85 \\
\bottomrule
\end{tabular}
\end{adjustbox}
\end{table*}

\subsection{Systems and data}
\label{sec:supp_conventions}
\paragraph{Which system each result uses.} Every result in this supplement uses the encoder and the renderer of the main
paper (50 sampling steps), unless a table names another model. Features come from the adapted encoder, except in the oracle
and feature-intervention results, which use teacher features (frozen V-JEPA on the true sharp clip); results read from
the features alone (probes, layer choice) use no renderer. Every generated result uses one sampler seed unless
\cref{tab:supp_seed} says otherwise.

\paragraph{Test sets and held-out status.} The test sets and their window counts are those of main-paper
\cref{sec:experiments}. Training uses the sources of \cref{tab:supp_training}, with the evaluation recordings excluded.
GoPro's test windows come from its 11 official test recordings.
I2-2000FPS and RBI are not among our training sources. No compared method trained on I2-2000FPS or on RBI's camera.

\paragraph{Synthetic blur generator.}\label{sec:supp_synth_blur}

\begin{figure}[t]
\centering
\includegraphics[width=\columnwidth]{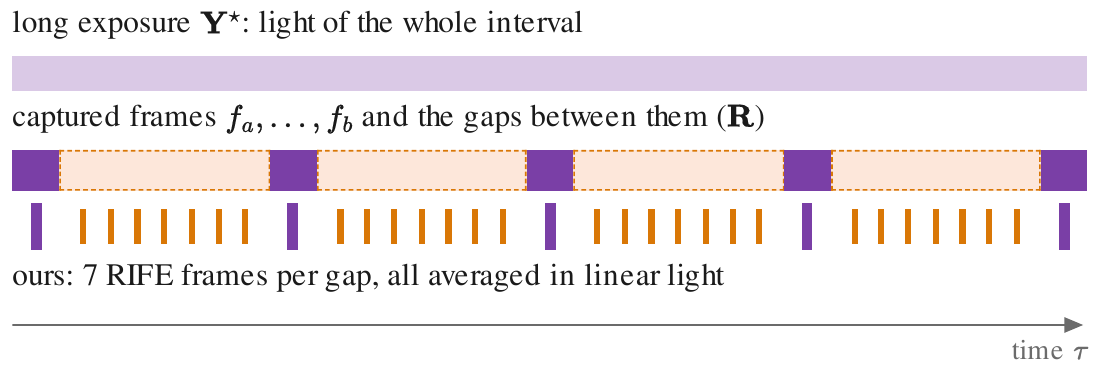}
\caption{One exposure in time. The long exposure collects light over the whole interval; captured
frames cover short windows, and the light in the gaps is the residual $\mathbf R$. We fill each
gap with seven interpolated frames and average all samples in linear light.}
\label{fig:supp_blur_timeline}
\end{figure}

Our synthetic blur approximates main-paper Eq.~\eqref{eq:exposure_gap} with captured
high-frame-rate frames (\cref{fig:supp_blur_timeline}). Given sRGB frames
$f_a,\dots,f_b\in[0,1]$, the blurred observation is
\begin{equation}
    \mathbf{Y}=\mathrm{EOTF}^{-1}\!\left(
    \frac{8\sum_{k=a}^{b-1}J_k+\mathrm{EOTF}(f_b)}{8(b-a)+1}\right),
\label{eq:supp_synth_blur}
\end{equation}
\begin{equation}
\begin{split}
    J_k=\tfrac18\Bigl[&\mathrm{EOTF}(f_k)\\
    &+\textstyle\sum_{j=1}^{7}
    \mathrm{EOTF}\bigl(\mathrm{RIFE}(f_k,f_{k+1},\tfrac{j}{8})\bigr)\Bigr],
\end{split}
\label{eq:supp_interval}
\end{equation}
where $\mathrm{EOTF}$ (electro-optical transfer function) is the IEC 61966-2-1 sRGB transfer function and
$\mathrm{RIFE}(f_k,f_{k+1},\tau)$ is the frame that RIFE~v4.25~\cite{huang2022rife}
interpolates at fraction $\tau$ between $f_k$ and $f_{k+1}$. Each $J_k$ is the linear-light mean
over one inter-frame interval.
\Cref{eq:supp_synth_blur} is therefore the display-encoded mean of $8(b-a)+1$ equally weighted
linear-light samples. Mapped onto main-paper Eq.~\eqref{eq:exposure_gap}, the captured frames
take the place of $\mathbf{x}_t^\star$ with $w_t=1/(8(b-a)+1)$. The interpolated frames stand in for
the residual $\mathbf{R}$, and $\mathrm{EOTF}^{-1}$ stands in for the camera's processing.

Interpolation assumes that the light in a gap is the neighbouring frames' content moved along a
path: nothing appears or disappears, and the lighting stays fixed (\cref{tab:supp_approx} lists
where this breaks). Together with the stand-in for camera processing, it leaves discrepancies from a real long exposure~\cite{rim2022realistic}.
\Cref{tab:supp_blur_recipe} compares synthetic recipes with real 18\,ms exposures.

The interpolated frames are computed once in native time order and reused, so a window
and its time reversal receive identical blur: the time-reversal ambiguity of main-paper
\cref{sec:model} holds exactly in our data. Both learned components are trained on real clips blurred with this recipe, except X4K, which uses the plain linear-light
average (row L of \cref{tab:supp_blur_recipe}); synthetic scenes form their blur from their own sub-frames. The synthetic falling-object test scenes use a plain frame average instead.

\begin{table}[!htbp]
\centering
\caption{\textbf{Averaging in linear light matches real camera blur best, 1.4\,dB above a plain sRGB average.} Synthetic versus real blur (beam-splitter data, 55 scenes): PSNR of each synthetic recipe, built from the 9 sharp 500\,fps frames of a real 18\,ms exposure, to that exposure after a per-scene colour fit. Test: windows $k=2$--25 of the 5 test videos (main-paper \cref{tab:main}'s RBI column: $k=2$--24); Train: the rest.}
\label{tab:supp_blur_recipe}
\footnotesize
\setlength{\tabcolsep}{3.5pt}
\begin{adjustbox}{max width=\columnwidth}
\begin{tabular}{L{0.36\columnwidth}ccc}
\toprule
Recipe & All (270) & Test (120) & Train (150) \\
\midrule
G, every other frame only & 31.224 & 31.152 & 31.281 \\
S, plain sRGB average & 31.454 & 31.387 & 31.507 \\
L, linear-light average & 32.822 & 32.311 & 33.230 \\
\bottomrule
\end{tabular}
\end{adjustbox}
\end{table}

\subsection{Evaluation protocol}
\label{sec:supp_protocol}
\paragraph{Windows, blur and truth.} Adobe240 has four windows per recording at evenly spaced positions, more than 8 frames
apart. GoPro and Adobe240 windows are blurred with our training recipe (\cref{sec:supp_synth_blur}). I2-2000FPS averages its native frames in linear light, over an exposure set as in \cref{sec:supp_i2}; RBI keeps
the real 18\,ms blur, with the 9 sharp frames captured during it as truth.
For GoPro and Adobe240, the truth is the nine middle frames of the window, at exposure fractions $(i{+}1)/10$,
$i=0,\dots,8$. Every output is resized to the truth's $1280\times736$ before scoring.

\paragraph{Frame times.}\label{sec:supp_frame_times} Every method is scored at the same physical times, although methods
output different numbers of frames. For each truth frame we take the method's output frame nearest in time, within half the
truth spacing. \paper and Motion-ETR output 9 frames, Tedla et al.\ 16 (at exposure fractions $(k{+}0.5)/16$), and Jin et
al.\ and Animation from Blur 7 (at $k/6$); all are scored at the truth frames at exposure fractions 0.2, 0.3, 0.5, 0.7 and 0.8. Deblur + I2V places the
deblurred image at its first frame, the exposure centre, so its output covers only the second half of the exposure; it is
scored at the truth frames at 0.5, 0.7 and 0.8.

\paragraph{Metrics.}\label{sec:supp_metrics} Shared frames are the truth frames at which a method is scored.
\begin{itemize}
\item \textbf{PSNR (dB, higher is better), LPIPS (lower is better):} per shared frame, averaged per window; LPIPS with AlexNet.
\item \textbf{EPE} (px, lower is better): end-point error between the optical flow of the output and that of the truth, both by SEA-RAFT~\cite{wang2024searaft}
(Tartan-C-T-TSKH-spring540x960-M, its spring-M settings) between consecutive shared frames of the output and of the
truth; the end-point error is averaged over all pixels, then over frame pairs, then over windows.
\item \textbf{Moving EPE:} as EPE, averaged only over pixels whose true flow is at least 1\,px.
\item \textbf{Track} (px, lower is better): the CoTracker3 point-track error; only in tables that name it.
\item \textbf{Cells, intervals, differences:} cells are means over windows; intervals are 95\% cluster-bootstrap intervals over
recordings (scenes for synthetic sets), 5{,}000 resamples, seed 0; differences between methods are paired over the same windows.
\end{itemize}

\paragraph{Time order.} A blur does not record which way time ran, so each blind row is scored in the better of the
forward and reversed order, chosen per window and per metric; rows given arrows or a guide are scored in the commanded order.
\Cref{tab:supp_order} shows that choosing one order per window for all metrics instead barely changes the scores, and that
in the true order every blind method has a higher EPE than \paper with arrows.

\paragraph{Baselines.} Published methods run with their authors' public weights and code: Tedla et al.'s in-the-wild model,
Motion-ETR, Jin et al.\ and Animation from Blur. None was retrained on our data, and each takes the same blurred input.
Animation from Blur uses the guide its own predictor samples; \cref{sec:supp_afb} gives it the guide from the true
motion. Deblur + I2V is a NAFNet trained on our training split, followed by Wan2.2-TI2V-5B image-to-video with an empty
prompt (50 steps, our resolution).

For local control (main-paper \cref{tab:control}), Wan-Move 14B, Tora and Go-with-the-Track receive the blur as their
first frame. Each arrow becomes one point track from its tail to its head over the generated clip. The track keeps the arrow's shape and
direction; its length is either the drawn arrow's or the region's true travel. From each generated clip, nine evenly spaced
frames are resized to $1280\times736$ (Go-with-the-Track generates 49 frames at $480\times832$; we take frames
$\mathrm{round}(48k/8)$, $k=0,\dots,8$).

\paragraph{Arrows at test time.} The arrows of the ``+ arrows'' rows (``+ arrow cues'' in the main paper) are simulated from the true clip by the
training procedure (\cref{sec:supp_cues_real}), with a fixed random seed per window. Arrows drawn by people are evaluated separately
(\cref{sec:supp_human_arrows}).

\section{Videos, data and release}
\label{sec:supp_videos}

\paragraph{Data.} All test and training data are public research datasets. Some clips in our figures show people in public places, as recorded in those datasets. The people who drew arrows are
described in \cref{sec:supp_human_arrows}.

\paragraph{Release.} We will release the code, the trained weights and result videos.

\section{Scope and future work}
\label{sec:supp_limitations}

\paper returns one plausible video per blur, not the full set of videos consistent with it; sampling several feature
estimates to represent that uncertainty is left for future work. Without an arrow the blur fixes the motion axis but not
its sign, so the output may play in reverse (\cref{tab:supp_order}); one arrow per moving part supplies the sign. The
learned no-cue default covers one physical case, a dropped object settling, and extending it to other physical behaviors
is also left for future work.

Small objects, very large blur and an arrow on one object under dominant camera motion
(\cref{sec:supp_control_results}) remain the hardest cases. With oracle features the renderer reproduces much larger motion
(\cref{sec:supp_range}), so a stronger encoder would extend the range without changing the renderer.

\end{document}